\documentclass[journal]{IEEEtran}
\usepackage{hyperref}
\hypersetup{
    colorlinks=true,
    citecolor=blue
}
\usepackage[T1]{fontenc}
\usepackage[numbers,compress]{natbib}
 
\usepackage{array}
\usepackage{graphicx} 
\usepackage{subfigure}
\usepackage{algpseudocode}  
\usepackage{amsmath}  
\usepackage{bm}
\usepackage{tabularx}
\usepackage{CJKutf8}
\usepackage[cmintegrals]{newtxmath}
\usepackage{url}
\usepackage[usenames,dvipsnames]{color}
\definecolor{Gray}{gray}{0.95}
\definecolor{mygreen}{rgb}{0.1,0.255,0.1}
\usepackage{colortbl}
\usepackage{multirow}
\usepackage[utf8]{inputenc}
\usepackage{cleveref}

\crefname{section}{§}{§§}
\Crefname{section}{§}{§§}
\usepackage{enumitem}
\usepackage{pifont}

\usepackage{ulem}
\usepackage{amsfonts}
\usepackage{algorithm}
\usepackage{algorithmicx}
\usepackage{xspace}

\usepackage{ulem} 

\usepackage{booktabs} 
\usepackage[table]{xcolor} 

\usepackage{tcolorbox}

\usepackage{tikz,xcolor}
\definecolor{lime}{HTML}{A6CE39}
\DeclareRobustCommand{\orcidicon}{
\begin{tikzpicture}
\draw[lime, fill=lime] (0,0)
circle[radius=0.16]
node[white]{{\fontfamily{qag}\selectfont \tiny \.{I}D}}; 

\end{tikzpicture}

\hspace{-2mm}
}
\foreach \x in {A, ..., Z}{%
\expandafter\xdef\csname orcid\x\endcsname{\noexpand\href{https://orcid.org/\csname orcidauthor\x\endcsname}{\noexpand\orcidicon}}
}

\definecolor{Gray}{gray}{0.95}

\ifCLASSINFOpdf
\else 
\fi

\begin{document}
\title{Privacy-Preserving Federated Embedding Learning for Localized Retrieval-Augmented Generation}

\author{\mbox{Qianren~Mao\orcidA{}, 
Qili~Zhang,
Hanwen~Hao,
Zhentao~Han,
Runhua~Xu\orcidB{},~\IEEEmembership{Member, IEEE}} \\
\mbox{
Weifeng~Jiang,
Qi Hu, 
Zhijun~Chen\orcidC{}, 
Tyler~Zhou, 
Bo~Li,~\IEEEmembership{Member, IEEE}} \\
\mbox{Yangqiu Song\orcidD{}, 
Jin~Dong,
Jianxin~Li\orcidE{},~\IEEEmembership{Member, IEEE} and Philip~S. Yu\orcidF{},~\IEEEmembership{Life Fellow, IEEE}}
\IEEEcompsocitemizethanks{\IEEEcompsocthanksitem

This work was supported by the National Natural Science Foundation of China under Grant 6250073746. Recommendedfor acceptance by xxxx (\textit{Corresponding authors}: Bo Li \& Jianxin Li.)
Thanks to Beijing Advanced Innovation Center for Future Blockchain and Privacy Computing for valuable resources and practical insights. 
Qianren. Mao (E-mail: maoqr@zgclab.edu.cn), is with the Zhongguancun Laboratory, Beijing 100095, China. 
Qili Zhang (E-mail: 20373496@buaa.edu.cn), Hanwen Hao (E-mail: 20373190@buaa.edu.cn), Zhentao Han (E-mail: 20373889@buaa.edu.cn), Runhua Xu (runhua@buaa.edu.cn), Zhijun Chen (E-mail: zhijunchen@buaa.edu.cn), Bo Li (E-mail: libo@buaa.edu.cn) and Jianxin Li (E-mail: lijx@buaa.edu.cn) are with the School of Computer Science and Engineering, Beihang University, Beijing 100191, China. 
Bo Li and Jianxin Li are also with the Zhongguancun Laboratory.
Weifeng Jiang (E-mial:weifeng001@e.ntu.edu.sg) is with Nanyang Technological University (NTU), 639798, Singapore.  
Qi Hu (E-mail: qhuaf@connect.ust.hk) and Yangqiu Song (E-mail: yqsong@cse.ust.hk) are with Hong Kong University of Science and Technology (HKUST), Hong Kong 999077, China. 
Tyler Zhou (E-mail: zmmwl@aliyun.com) serves as architect of Beijing Academy of Blockchain and Edge Computing (BAEC), Beijing, China.
Dong Jin (E-mail: dongjin@baec.org.cn) is the General Director of Beijing Academy of Blockchain and Edge Computing. He is also the General Director of Beijing Advanced Innovation Center for Future Blockchain and Privacy Computing, Beijing, China.
Philip~S. Yu (E-mail: psyu@uic.edu) is with the Department of Computer Science, University of Illinois at Chicago, Chicago 60607 USA.
}

\thanks{
The first four authors contributed equally.
\protect\\
Manuscript received July 3, 2020; revised August XX, 2020.}}
\markboth{Journal of \LaTeX\ Class Files,~Vol.~14, No.~8, August~2015}%
{Shell \MakeLowercase{\textit{et al.}}: Bare Demo of IEEEtran.cls for Computer Society Journals}


\IEEEtitleabstractindextext{%
\begin{abstract}
Retrieval-Augmented Generation (RAG) has recently emerged as a promising solution for enhancing the accuracy and credibility of Large Language Models (LLMs), particularly in Question \& Answer tasks. 
This is achieved by incorporating proprietary and private data from integrated databases. However, private RAG systems face significant challenges due to the scarcity of private domain data and critical data privacy issues. These obstacles impede the deployment of private RAG systems, as developing privacy-preserving RAG systems requires a delicate balance between data security and data availability.
To address these challenges, we regard federated learning (FL) as a highly promising technology for privacy-preserving RAG services. We propose a novel framework called Federated Retrieval-Augmented Generation (\textbf{FedE4RAG}). This framework facilitates collaborative training of client-side RAG retrieval models. The parameters of these models are aggregated and distributed on a central-server, ensuring data privacy without direct sharing of raw data. In FedE4RAG, knowledge distillation is employed for communication between the server and client models. This technique improves the generalization of local RAG retrievers during the federated learning process. Additionally, we apply homomorphic encryption within federated learning to safeguard model parameters and mitigate concerns related to data leakage.
Extensive experiments conducted on the real-world dataset have validated the effectiveness of FedE4RAG. The results demonstrate that our proposed framework can markedly enhance the performance of private RAG systems while maintaining robust data privacy protection.
\end{abstract}

}

\maketitle
\IEEEdisplaynontitleabstractindextext
%
\IEEEpeerreviewmaketitle

\vspace{0.45in}
\IEEEraisesectionheading{\section{Introduction}\label{sec:introduction}}

\IEEEPARstart{R}etrieval-augmented generation (RAG) has emerged as a promising solution by incorporating knowledge from external databases (e.g., vector database), thereby enhancing generation accuracy and credibility, particularly for Question \& Answer tasks~\citep{MaoHLSG0C20},\cite{IzacardG21},\cite{0002WL022},\cite{HofstatterC0Z23},\cite{ESJAS24},\cite{abs-2401-15391}. 
LLMs-based RAG's integration of dynamic external information optimizes its adaptation to vertical domain data environments~\cite{LewisPPPKGKLYR020},\cite{JiangXGSLDYCN23},\cite{abs-2002-08909},\cite{BorgeaudMHCRM0L22},\cite{AsaiWWSH24},\cite{singh2025agenticretrievalaugmentedgenerationsurvey},\cite{abs-2501-00309},\cite{abs-2412-15529}. Access to external databases or the internet allows RAG systems to utilize search engines for accurate, real-time information, enhancing reliability with domain-specific expertise~\citep{JiLFYSXIBMF23},\cite{abs-2304-08979} and continuous knowledge updates\cite{abs-2301-00303}.

However, the practical deployment of RAG systems often encounters substantial challenges in sensitive commercial or institutional environments due to strict data governance frameworks (e.g., GDPR) and privacy concerns~\cite{hoffmann2022training, muennighoff2024scaling, bommasani2021opportunities}. 
\textit{\textbf{Open-access RAG}} architectures inherently expose fine-tuning datasets of inference LLMs~\citep{CarliniTWJHLRBS21},\cite{KandpalWR22},\cite{abs240105778},\cite{wuprivacy} and retrieval datasets of context retrievers~\citep{abs-2402-16893},\cite{abs-2402-08416},\cite{abs-2406-19234},\cite{abs-2405-20446} to privacy attacks. Adversaries may exploit vector representations to glean sensitive insights~\cite{0003XS23},\cite{MorrisKSR23},\cite{HuangTHLL24}, or extract confidential information via carefully crafted queries~\cite{abs-2402-16893},\cite{abs-2405-20446},\cite{ZengZHLX000WYT24},\cite{abs-2402-17840},\cite{abs-2411-14110}.

To mitigate these threats, recent deployments shift towards \textit{\textbf{Privatized RAG}}  systems designed exclusively for controlled zones like local area networks (LANs). Such systems maintain sensitive domain-specific retrieval data exclusively within institutional boundaries, minimizing the possibilities of unauthorized exposure~\cite{abs240106800},\cite{abs230314070}. 
Particularly under stringent regulatory compliance, federated learning (FL) serves as an effective privacy-preserving paradigm for localized RAG deployments. 
FL allows distributed clients to collaboratively train models by sharing only aggregated model parameters rather than raw data, thereby limiting privacy risks significantly.

Nevertheless, implementing FL introduces new security challenges. Traditional federated settings assume the presence of an "\textbf{honest-but-curious}" central server, that honestly follows protocols but may try passively to infer sensitive data from client updates~\cite{yang2019federated},\cite{mothukuri2021survey}. 
Ensuring the privacy and robustness of information exchange thus necessitates adopting secure computation methods and privacy-aware FL mechanisms, such as secure aggregation and federated knowledge distillation~\cite{li2020federated},\cite{wu2022communication}. These mechanisms further minimize information leakage risks and keep sensitive local semantics confidential during the federated update processes.


Privatized RAG systems are particularly relevant in scenarios where enterprise-level domain-specific data is dispersed across multiple institutions. 
For instance, within the legal sector, such data may be distributed among various courts, legal consulting firms, and other judicial entities. Similarly, in the financial domain, data is often fragmented across different institutions like banks, insurance companies, and financial regulatory bodies~\cite{ye2024openfedllm}. Such scenarios strongly necessitate privacy-preserving FL-enabled RAG systems to effectively aggregate and exploit distributed yet protected resources.

\begin{figure}[t]
  \begin{center}
  \includegraphics[width=.45\textwidth]{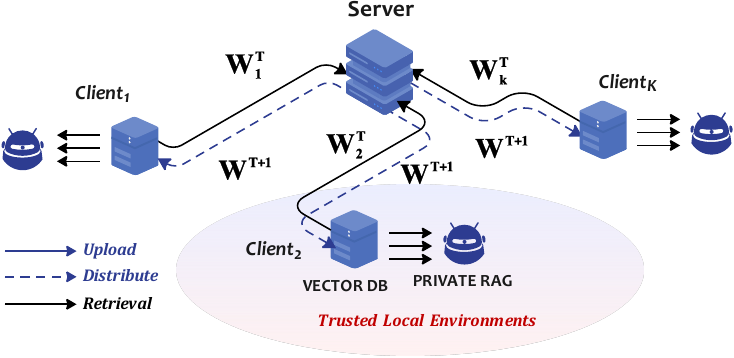}
  \end{center}
  \caption{Schematic overview of our general Federated Embedding Learning for the private \textbf{localized RAG systems}. Different clients provide complementary data, while each client need not expose their data through federated learning.}
  \label{main_fede4rag}
\end{figure}

Motivated by these critical privacy and security considerations, this paper proposes a novel framework, Federated Embedding Learning for Privacy-Preserving Localized RAG (\textbf{FedE4RAG}\footnote{Code available at: \url{https://github.com/DocAILab/FedE4RAG}}), shown in Figure~\ref{main_fede4rag}.  FedE4RAG seeks to boost retrieval effectiveness by harnessing the distinct data features across various clients, such as specialized domain insights, without compromising the confidentiality of their proprietary data. Employing a secure, federated learning approach, FedE4RAG allows multiple clients to collaboratively enhance their generalized embeddings, guaranteeing that their data stays on-site and secure throughout the training process.

Further, we design a secure communication protocol that integrates cryptographic techniques with federated learning and knowledge distillation-based aggregation to ensure the protection of private information from leakage through gradient updates. 
We employ homomorphic encryption~\cite{MarcollaSMBFA22}, to process encrypted gradients directly, eliminating the need for data transformation among clients. We also implement federated knowledge distillation to enable the transfer of specialized local knowledge to a comprehensive global knowledge. 
This method improves the generalizability and applicability of local data representations, ensuring that robust data embeddings, developed iteratively, are transferred back to clients, thereby enhancing the localized RAG system's overall performance.

To demonstrate FedE4RAG's practicality, we specifically validate our system within the financial domain, where privacy standards are exceptionally stringent, by constructing an upstream federated dataset (\textbf{FedE4FIN}\footnote{Available at: \url{https://huggingface.co/datasets/DocAILab/FedE4FIN}}) curated from genuine institutional private data sources. Moreover, we further evaluate downstream generative effectiveness via the newly developed \textbf{RAG4FIN}\footnote{Available at: \url{https://huggingface.co/datasets/DocAILab/RAG4FIN}} dataset. 

Overall, our contributions can be summarized as follows:

\begin{itemize}[leftmargin=*]
\item 
We introduce private-preserving federated learning procedure for localized retrieval-augmented generation (RAG) system. FedE4RAG utilizes federated learning for  fine-tuning those localized retriever in each private RAG system.

\item 
To enhance the global scalability of localized knowledge in federated  learning (FL), we propose federated knowledge distillation for the FL communication. This approach facilitates the transfer of local knowledge to global scale without centralizing client data, thereby upholding stringent privacy measures while enriching the diversity and of private data.

\item 
By implementing FedE4RAG, we confirm the framework's efficacy within the financial domains—where stringent data security is essential and retrieval-augmented generation is notably demanding—utilizing newly developed datasets for upstream retrieval and downstream generation.
\end{itemize}


\section{Related Work}
\label{RW}
\subsection{Knowledge Distillation for Federated Learning}
Federated Averaging (FedAvg) serves as the cornerstone algorithm for Federated Learning (FL)~\cite{konevcny2016federated},\cite{McMahanMRHA17}. It operates through synchronous rounds within a client-server framework. At each round's start, the server broadcasts the global model's parameters to selected clients (i.e., the participants). These clients then train the model locally with their data and send updates back to the server. The server aggregates these updates using a weighted average based on the clients' data volume and updates the global model with these aggregated 'gradients'~\cite{ReddiCZGRKKM21,SunCGYY22}. This process initiates a new round by distributing the updated model, thereby advancing the learning cycle. Parameter-averaging methods like FedAvg face limitations due to potential information leaks and intellectual property issues, as they require sharing model structures. Additionally, varying data across clients can cause models to diverge (i.e., client drift), negatively impacting the global model's effectiveness when parameters are directly aggregated~\cite{LiHYWZ20},\cite{KimKH22},\cite{LiDCH22},\cite{YangLHSL024}. Subsequently, adaptations of Knowledge Distillation (KD) in federated learning have addressed the limitations of parameter-averaging schemes by promoting model heterogeneity and reducing communication costs. These strategies, also motivated by encouraging privacy properties~\cite{PapernotAEGT17}, facilitate the exchange of model outputs or intermediate representations instead of direct parameter or update transfers~\cite{abs-1811-11479},\cite{0001AA20},\cite{ItaharaNKMY23},\cite{GongSKWCDI21}.

Initially, model-agnostic federated learning strategies~\cite{AnilPPODH18} predominantly utilize server-side ensemble distillation during the FedAvg aggregation phase~\cite{AnilPPODH18},\cite{LinKSJ20}, exemplified by distributed co-distillation adaptations. 
These methods facilitate model heterogeneity and enhance communication efficiency by exchanging locally-computed statistics~\cite{abs-1811-11479}, model outputs~\cite{abs-1811-11479},\cite{abs-2110-11027}, and intermediate features~\cite{LiuLZKWBRC20} rather than direct model parameters.
Secondly, to mitigate performance degradation caused by non-IID data distributions, the conventional strategies are categorized as: (a) server-side enhancements to FedAvg aggregation with an additional distillation phase for global model correction, and (b) client-side techniques that locally distill global knowledge to address client drift directly.

In our focus on client-side RAG system, we aggregate and optimize localized RAG retriever. We combine server-side global model correction with client-side model drift control methods, allocating them for model-agnostic learning strategies. During the process of knowledge consolidation, we align only the predicted logits between the teacher and client models, thereby eliminating the need to address structural compatibility between the teacher and student models.

\subsection{Federated Retrieval Augmented Generation}
Federated learning (FL) serves as a technical paradigm for retrieval-augmented generation (RAG), enhancing privacy protection and ensuring regulatory compliance of RAG systems~\cite{abs-2311-15792},\cite{WangZCLMOXZ24}. 
Wang et al.,~\cite{WangKZZ24} propose FeB4RAG, a federated query dataset specifically designed for retrieval-based question answering tasks. This framework is structured such that a query relies on databases from multiple clients, particularly vector database data. C-FedRAG~\cite{abs-2412-13163} utilizes confidential virtual machines with cryptographic attestation, as outlined by the Confidential Computing Consortium, to ensure secure processing by restricting execution to authorized code. By integrating Trusted Execution Enclaves, it protects data privacy in RAG workflows and maintains confidentiality during retrieval and inference (Lee et al., 2024), enhancing security and privacy in federated learning systems. RAFFLE~\cite{muhamedcache} utilizes public healthcare datasets for training and employs information retrieval and augmentation with private patient datasets during inference. By limiting server access to private documents, it mitigates compliance challenges. Each client maintains both confidential and public data. 
Nonetheless, C-FedRAG does not focus on client-driven Localized Retrieval-Augmented Generation or the integration of private data for optimizing the federated framework. Instead, RAFFLE requires prior labeling of private client data and involves a two-step process: federated learning of the server model, followed by client-side inference with private data.

Unlike FeB4RAG, a unified RAG system that integrates results into a server-side framework through federated querying, and unlike C-FedRAG, which employs trusted execution environments to safeguard private data, the proposed FedE4RAG system specifically optimizes client-side private RAG systems via federated learning. FedE4RAG fine-tunes localized retrievers within each client system. In contrast to RAFFLE, which utilizes federated learning on public data and then infers on private data, FedE4RAG focuses on leveraging private data from various clients; this private data is more valuable and confidential across multiple companies. This approach facilitates the transfer of local private knowledge to a global model without centralizing client data, thereby strictly maintaining privacy while simultaneously improving the performance of localized RAG systems.

\section{PRELIMINARIES}\label{PRELIMINARIES}
\subsection{Federated Averaging Learning}
Federated learning (FL) enables client collaboration in model training under central coordination, preserving privacy and leveraging distributed data for AI scalability across sectors. The FedAvg algorithm reduces communication costs by enabling multiple local updates before synchronization and utilizes client computational power, making it ideal for large-scale, decentralized environments. 
To facilitate the training of local models with privatized data shared across multiple clients, naive federated learning (FL) is inherently privacy-preserving, as raw data remains on local devices. However, the risk of privacy breaches persists, as attackers may infer sensitive information from shared model updates or gradients via model inversion attacks. An effective countermeasure involves the implementation of homomorphic encryption for communication between clients and the server~\cite{XuJ022},\cite{abs-2303-10837},\cite{YanL0WH0W24},\cite{XuLLJML24}.

In a FL setting, we consider a set of \( K \) clients, each with its own local dataset \( \mathcal{D}_k \). The global model parameters are denoted by \( \mathbf{W} \). The goal is to minimize a global objective function \( F(\mathbf{w}) \), which is typically defined as the weighted sum of local objective functions:
\begin{equation}
F(\mathbf{w}) = \sum_{k=1}^{K} \frac{n_k}{n} F_k(\mathbf{w}),
\end{equation}
where \( n_k = |\mathcal{D}_k| \) is the number of data points held by client \( k \), and \( n = \sum_{k=1}^{K} n_k \) is the total number of data points across all clients. The local objective function \( F_k(\mathbf{w}) \) is typically the empirical risk over the local dataset \( \mathcal{D}_k \):

\begin{equation}
F_k(\mathbf{w}) = \frac{1}{n_k} \sum_{i \in \mathcal{D}_k} \ell(\mathbf{w}; \mathbf{x}_i, y_i),
\end{equation}
where \( \ell(\mathbf{w}; \mathbf{x}_i, y_i) \) is the loss function for a data point \( (\mathbf{x}_i, y_i) \).
The overall optimization problem in base federated learning can be formally described as:

\begin{equation}
\min_{\mathbf{w}} F(\mathbf{w}) = \min_{\mathbf{w}} \sum_{k=1}^{K} \frac{n_k}{n} \left( \frac{1}{n_k} \sum_{i \in \mathcal{D}_k} \ell(\mathbf{w}; \mathbf{x}_i, y_i) \right).
\end{equation}

\noindent \textbf{Federated Average.} One of the most widely used fusion algorithms in FL is the Federated Averaging (FedAvg) algorithm, introduced by McMahan et al. \cite{McMahanMRHA17}. It iteratively optimizes through local stochastic gradient descent (SGD) on each client, punctuated by periodic model averaging. The FedAvg algorithm proceeds as follows:
\begin{itemize}[leftmargin=*]
 \item \textit{Initialization:} The server initializes the global model parameters \( \mathbf{w}^0 \) (e.g., from BGE model in our settings).
 
 \item \textit{Client Update:} At each communication round \( t \), a subset of clients \( \mathcal{S}_t \subseteq \{1, \ldots, K\} \) is selected. Each client \( k \in \mathcal{S}_t \) performs \( E \) epochs of local SGD on its local dataset \( \mathcal{D}_k \) and updates its local model \( \mathbf{w}_k^t \):
    \begin{equation}
    \mathbf{w}_k^{t+1} = \mathbf{w}^t - \eta \nabla F_k(\mathbf{w}_k^t),
    \end{equation}
    where \( \eta \) is the learning rate.

\item \textit{Model Aggregation:} The server aggregates the updated local models to form a new global model:
    \begin{equation}
    \mathbf{w}^{t+1} = \sum_{k \in \mathcal{S}_t} \frac{n_k}{n_{\mathcal{S}_t}} \mathbf{w}_k^{t+1},
    \end{equation}
    where \( n_{\mathcal{S}_t} = \sum_{k \in \mathcal{S}_t} n_k \) is the total number of data points in the selected clients.

\item \textit{Repeat:} The steps of update and aggregation are repeated for a predefined number of communication rounds.
\end{itemize}

The FedAvg algorithm reduces communication costs by enabling multiple local updates before synchronization and utilizes client computational power, making it ideal for large-scale, decentralized environments. While FL inherently preserves privacy by ensuring that raw data remains on clients' devices, it is vulnerable to privacy risks, such as model inversion attacks, where adversaries infer sensitive information from shared model updates or gradients~\cite{haibo2023ese,HuangYSWLDY24}. Addressing these vulnerabilities necessitates advanced privacy-preserving techniques, such as homomorphic encryption.

\subsection{Homomorphic Encryption-Based FL} 
Homomorphic encryption~\cite{MarcollaSMBFA22} is a cryptographic method that allows computations to be performed directly on encrypted data without requiring decryption. Incorporating HE into FL, termed HE-based Federated Learning (\textbf{HEFL}), ensures robust privacy guarantees by encrypting all communications between clients and the server.  
In homomorphic-encryption-based privacy-preserving federated Learning, each client encrypts its local data and model updates using homomorphic encryption before sending them to the central server. The server aggregates these encrypted updates without decrypting them, protecting clients' data privacy. 
Let  \( \mathcal{E}.\mathtt{E}(\cdot) \) represent the encryption function and  \( \mathcal{E}.\mathtt{D}(\cdot) \), the decryption function. Given a global model \( \mathbf{w} \), each client \( k \) calculates its local gradient \( \nabla F_k(\mathbf{w}_{k}) \) and encrypts it as \( \mathcal{E}.\mathtt{E}(\nabla F_k(\mathbf{w}_{k}))\). 
The server receives these encrypted gradients from all clients, aggregates them homomorphically. The HEFL procedure is formulated as: 
\begin{equation}
  F(\mathbf{w})= \mathcal{E}.\mathtt{D}(\sum_{k=1}^{K} \mathcal{E}.\mathtt{E}(\nabla F_k(\mathbf{w}_{k}))),
\end{equation}
where $\sum_{k=1}^{K} \mathcal{E}.\mathtt{E}(\nabla F_k(\mathbf{w}_{k}))$ represents the encrypted global gradients, and the aggregated gradient can be decrypted using a decryption function \( \mathcal{E}.\mathtt{D}(\cdot) \) in localized RAG system to update the model parameters. 
Despite the computational overhead introduced by encryption, decryption, HEFL provides strong privacy guarantees by ensuring that the server never accesses raw data or unencrypted model updates.

\begin{figure*}[t]
  \center
  \includegraphics[width=1.\textwidth]{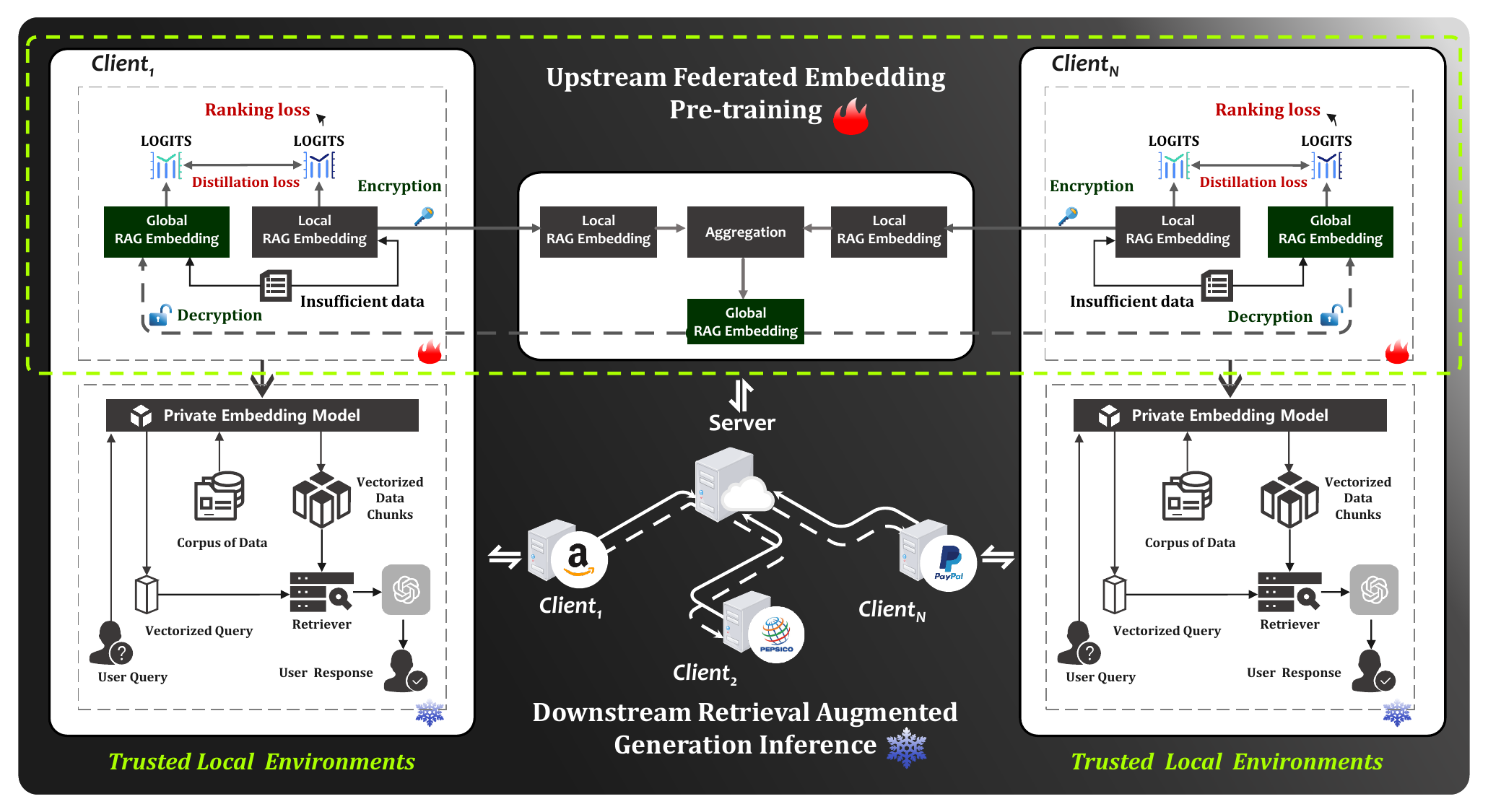}
  \vspace{-0.2in}
  \caption{Schematic representation of the Federated Embedding Learning (FedE4RAG) framework, illustrating the two-phase process: upstream federated embedding pre-training and downstream retrieval-augmented generation inference. The upstream phase focuses on pre-training with federated embedding and knowledge distillation to address data scarcity and enhance local RAG retrievers. The downstream phase depicts private embedding models operating within trusted local environments to perform retrieval and answer generation without data exposure.
  }
 \label{main_framework}
\end{figure*}


\subsection{BGE Embedding in RAG}
The BGE~\cite{xiao2023c} embedding (BAAI General Embedding\footnote{\url{https://bge-model.com}}) series, are most popular embedding models. Utilizing the MAE-style approach from RetroMAE~\cite{abs-2205-12035},\cite{LiuXSC23}, these models demonstrate strong versatility for various downstream NLP applications. 
They are integrated into major RAG and text-embedding frameworks such as Langchain\footnote{\url{https://python.langchain.com/docs/integrations/text_embedding/bge}}, LLamaIndex\footnote{\url{https://docs.llamaindex.ai/en/stable/}}, and HuggingFace\footnote{\url{https://huggingface.co/docs/text-embeddings-inference/index}}. 
Additionally, BGE models provide a solid foundation for fine-tuning, such as general purpose fine-tuning. The pre-trained model is fine-tuned on C-MTP~\cite{xiao2023c} (unlabeled) via contrastive learning, where it is learned to discriminate the paired texts from their mini-batch negative samples~\cite{KarpukhinOMLWEC20}. 
This method allows the model to simultaneously train on multiple positive and negative samples for a given query, which helps in enhancing the model's ability to distinguish between relevant and irrelevant text pairs. 
Empirical evidence indicates that fine-tuned BGE models outperform their original versions~\cite{ChenXZLLL24,GuptaRSP24}. Moreover, BGE offers both ready-to-use embeddings and a framework for developing more advanced embeddings.

In RAG systems, the retriever, which operates as an embedding model, transforms information from external knowledge bases into continuous vector representations. 
By capturing semantic relationships~\cite{abs-1301-3781},\cite{SapBABLRRSC19}, contextual information~\cite{PetersNIGCLZ18},\cite{radford2018improving},\cite{DevlinCLT19},\cite{radford2019language}, linguistic knowledge~\cite{KleinM03}, and other forms of knowledgable representation. By doing so, the retriever model enhances the system's ability to understand user queries and identify relevant context. The selected context and query are subsequently utilized by generative LLMs to augment the generation of answers to users' queries.

Therefore, the retriever, comprising the embedding model, constitutes a critical initial stage in determining the performance of a pipeline-structured RAG system.  Evidently, the retriever constitutes a pivotal initial component in determining the efficacy of a pipeline-structured RAG system. Optimization of the embedding model to improve the quality of retrieved context is imperative for the RAG systems.

\section{FedE4RAG}\label{sec:graphfordoc}
We develop and analyze the FedE4RAG framework, which is divided into two main components, as depicted in Figure~\ref{main_framework}:  upstream federated embedding fine-tuning and downstream private retrieval-augmented generation. 
The upstream process involves the general-purpose fine-tuning process based on the BGE embedding model. 
To ensure that each client \(\mathbf{C}_{i}\) performs upstream learning and response to downstream questions in a trusted localized environment, we employ homomorphic encryption for privacy-preserving federated learning to fine-tune the local embedding model \(\mathbf{M}_{i}\). Additionally, to address the issue of drift between local optimization and global convergence caused by data deficiencies in each client, we use a federated knowledge distillation mechanism for the embedding communication. This approach allows the server to aggregate the knowledgable data representations from all clients and distill them back to individual clients, improving the embedding models impacted by the data deficiencies of \(\mathbf{D}_{i}\) in the client \(\mathbf{C}_{i}\) and thus augmenting the local RAG retriever.

Specifically, from the perspective of an individual client, the complete process of privacy-preserving RAG can be summarized as follows: in the retrieval stage, the input consists of specific user queries and the RAG retriever utilizes localized embedding models to obtain query representations and domain text representations. Then RAG system conducts retrieval to obtain context relevant to the query. In the Q\&A stage, the LLM of RAG system in each client utilizes the query and its corresponding retrieved context to generate adequate answers.

FedE4RAG provides each client with a trusted local environment, ensuring that the entire RAG process occurs solely on the client side without transmitting local private data. Using a federated augmented method with a knowledge distillation mechanism, FedE4RAG addresses the issue of reduced performance in local embedding models caused by the lack of private data. This approach remains effective even in extreme cases where a client has no data available to fine-tune the retriever. Consequently, FedE4RAG ensures that each client obtains a robust retriever \(\mathbf{R}_{i}\) for the subsequent process of downstream localized RAG tasks.

\noindent\textbf{Threat Model.} 
We assume an "honest-but-curious" central aggregation server and participating clients, where they faithfully follow the specified protocols but may attempt to infer sensitive or proprietary client information from legitimately exchanged embeddings or model parameters~\cite{yang2019federated, mothukuri2021survey}. Additionally, we consider external adversaries capable of passively intercepting communicated data, conducting inference attacks to extract sensitive information~\cite{0003XS23,MorrisKSR23,HuangTHLL24,ZengZHLX000WYT24}. We do not consider active malicious adversaries or collusions and assume the presence of secure, authenticated communication channels. 
Our privacy-preserving framework explicitly addresses these threats through the use of secure aggregation and federated knowledge distillation to prevent passive inference attacks.

\subsection{Upstream Embedding Learning}\label{sec:UpFedE}
The problem to address is how to fine-tune local embedding models using insufficient local data.  It also explores the implementation of a privacy-preserving federated learning process that securely aggregates and transmits model gradients, thus enhancing the performance of local embedding models despite data constraints. To confront the prevailing challenges, a set of three indispensable modules have been developed for refining local embedding models. 
The first module of RAG fine-tuning, aptly named \textbf{RAG-FT},  is designed to fine-tune BGE embeddings specifically for RAG retrieval tasks. The second module of knowledge distillation between global and local embeddings, \textbf{KD-GLE} ensures that the local models benefit from the collective knowledge of the global model, leading to improved convergence and performance despite variations in local data distributions. Finally, the third module of homomorphic federated learning, \textbf{FED-HE}, ensures the privacy of local data by training embeddings in an encrypted state using homomorphic federated learning.


\subsubsection{RAG-FT}\label{sec:RAG-FT}
To develop privatized RAG-purpose embedding models, we fine-tune the BGE embedding model for each client using synthetic data which is introduced in Section~\ref{dataset}.  Specifically, the encoder of the BGE model processes pairs of user queries \(x^{q}\) and corresponding text chunks \(x^{c}\). The RAG-FT employs a contrastive optimization objective to predict similarity relationships between these pairs, thereby enhancing the model's ability to capture semantic similarities between query and contextual sentences. Thus, by leveraging contrastive learning, the model effectively learns to distinguish relevant context from irrelevant information, which is critical for improving the performance of downstream RAG systems. 
More concretely, during the synthetic data generation phase, we utilize Llamaindex\footnote{\url{https://github.com/run-llama}} modules to automatically create a set of questions from unstructured text chunks. These (question, chunk) pairs serve as positive examples for training the Federated embedding model, with negative examples randomly sampled from other chunks in the same mini-batch.

In the contrastive learning fine-tuning phase, we employ Information Noise-Contrastive Estimation  
 (InfoNCE) Loss~\cite{abs-1807-03748},\cite{ChenK0H20},\cite{He0WXG20},\cite{GaoYC21} as the training objective, utilizing the BGE-BASE model from HuggingFace\footnote{\url{https://huggingface.co/}} as the base model for optimization. We denote \(\textbf{h}^{q}=f_{\theta }(x^{q})\), and \(\textbf{h}^{c}=f_{\theta }(x^{c})\), where \(f_{\theta }\) is a contrastive-learning embedding model that employs RetroMAE encoders. By inputting question \(x^{q}\) and chunk \(x^{c}\) separately, the RetroMAE outputs the hidden state vectors \(\textbf{h}^{q}\) and \(\textbf{h}^{c}\). As the question and text chunk pairs in a mini-batch serve as positive training signals, while negative examples are randomly sampled from other chunks, resulting in a contrastive learning optimization strategy for optimization:

\begin{equation}
   \mathcal{L}(\textbf{h}^{q}, \textbf{h}^{c})=-\frac{1}{N}\sum_{i=1}^{N}log\frac{exp({\rm{sim}(\textbf{h}_{i}^{q}, \textbf{h}_{i}^{c})/\tau })}{\sum_{j=1}^{N}exp({\rm{sim}(\textbf{h}_{i}^{q}, \textbf{h}_{j}^{c})/\tau })}, 
\end{equation}

where \(sim(u,v)=u^\top v/\left ( \left \| u \right \|_{2}, \left \| v \right \|_{2} \right )\) denotes the cosine similarity of two text vector \(u\) and \(v\). \(\tau\) represents the temperature hyperparameter. In our experiments, we set \(\tau = 1\).

\subsubsection{KD-GLE}\label{sec:KD-GLE}
In federated learning settings, a pervasive challenge is data heterogeneity, which manifests as non-IID data distributed\footnote{\url{https://en.wikipedia.org/wiki/Federated_learning}} across different entities such as companies. Non-IID data~\cite{ZhangLDZZX25} refers to data that is not independent and identically distributed, often varying significantly in distribution, characteristics, and volume across clients. 
This heterogeneity introduces substantial difficulties during the model weight averaging phase, as traditional federated learning algorithms assume that data across clients follows an IID distribution. When this assumption is violated, the performance of the aggregated model can degrade significantly, leading to a notable reduction in accuracy compared to scenarios where data is IID.
To tackle this issue, we implement a sophisticated knowledge distillation strategy between the server's global embeddings and each client's local embeddings. The KD-GLE is designed to address the statistical heterogeneity by enabling a dual training process. On one hand, each client independently trains a personalized model that is finely tuned to the specific characteristics of its local data by comparative fine-tuning.  On the other hand,  clients collaboratively contribute to the training of a generalized server model that captures the collective patterns across all clients by knowledge distillation.

More concretely, we denote \(f_{\theta}^{g}\) is contrastive-learning global embedding model and \(f_{\theta}^{l}\) is contrastive-learning local embedding model. The penalty term \(\mathcal{L}(\textbf{z}^{l}, \textbf{z}^{g})\) is defined as the loss measuring the discrepancy between the similarity of the local model's representations. Specifically, the similarity of the local model's representations is defined as \(\textbf{z}^{l} = \text{sim}(f_{\theta}^{l}(x^{q}), f_{\theta}^{l}(x^{c}))\). Similarly, the similarity of the global model's representations is defined as \(\textbf{z}^{g} = \text{sim}(f_{\theta}^{g}(x^{q}), f_{\theta}^{g}(x^{c}))\), calculated over the synthetic privacy data. This loss is given by:

\begin{equation}
\mathcal{L}(\textbf{z}^{l}, \textbf{z}^{g}) = \frac{1}{N} \sum_{i=1}^{N} \left\| \textbf{z}_{i}^{l} - \textbf{z}^{g} \right\|_{2}^{2}.
\end{equation}


\subsubsection{FED-HE}\label{sec:FED-HE}
To enhance privacy in our proposed federated RAG framework, we propose using homomorphic encryption into federated learning to protect the exchanged local RAG embeddings.
Specifically, let $\mathcal{E}$ represent the  the employed fully homomorphic encryption (\textbf{FHE})~\cite{MarcollaSMBFA22} scheme  (in this work, we use the \textbf{CKKS}~\cite{CheonKKS17} scheme) and $[\![\cdot]\!]$ denote the ciphertext.

At communication round $t$, each client receives the aggregated and encrypted global gradients $[\![\nabla F(\mathbf{w}^{t-1})]\!]$ to train RAG embedding  from the previous round $t-1$.
Upon decrypting$[\![\nabla F(\mathbf{w}^{t-1})]\!]$ to get global RAG embedding gradients $\nabla F(\mathbf{w}^{t-1})=\mathcal{E}.\mathtt{D}([\![\nabla F(\mathbf{w}^{t-1})]\!])$ using the FHE scheme $\mathcal{E}$, each client $\mathbf{C}_i$ performs local embedding learning and generates an updated local RAG embedding gradients $\nabla F(\mathbf{w}_{\mathbf{C}_i}^{t})$.

 To ensure privacy, the client encrypts this updated gradients using the encryption function $\mathcal{E}.\mathtt{E}(\cdot)$, resulting in:  
\begin{equation}
  [\![\nabla F(\mathbf{w}_{\mathbf{C}_i}^{t})]\!] = \mathcal{E}.\mathtt{E}(\nabla F(\mathbf{w}_{\mathbf{C}_i}^{t})).
\end{equation}  

Once all encrypted local RAG embeddings gradients are collected from participating clients, the server performs secure aggregation to compute the new global RAG embedding gradients. This aggregation process operates directly on the encrypted embeddings gradients, ensuring that the server never accesses the plaintext embeddings gradients. The aggregated global RAG embedding gradients for round $t$ is computed as:  
\begin{equation}
[\![\nabla F(\mathbf{w}^{t})]\!] = \sum_{\mathbf{C}_i \in \mathcal{S}_t} \frac{n_{\mathbf{C}_i}}{N} [\![\nabla F(\mathbf{w}_{\mathbf{C}_i}^{t})]\!],
\end{equation}  
where $N = \sum_{\mathbf{C}_i \in \mathcal{S}_t} n_{\mathbf{C}_i}$ is the total number of data points contributed by the set of participating clients $\mathcal{S}_t$ in training round $t$.  
This approach ensures that the sensitive local embeddings remain encrypted throughout the entire training process, providing strong privacy guarantees while enabling collaborative learning.

\begin{table*}[tb]
  \centering
  \caption{Statistics of upstream and downstream datasets. Upstream training data refers to query-chunk pairs synthesized from documents for training.  Both upstream validation and downstream testing use the RAG paradigm. The upstream uses a smaller corpus for quick verification, while the downstream employs a full corpus for realistic assessment.
  }
  \renewcommand\arraystretch{1.6}
  \setlength{\tabcolsep}{1.2mm}{
  \begin{tabular}{|ccccccc|cccccc|}
    \specialrule{1pt}{0pt}{0pt}
  \multicolumn{7}{|c|}{\textbf{Upstream}}                                                       & \multicolumn{6}{c|}{\textbf{Downstream}}                        \\ \hline
  \multicolumn{1}{|l}{\textbf{Dataset}} & \multicolumn{1}{|l|}{\textbf{Train} (pairs)} & 
  \multicolumn{1}{l|}{\cellcolor{Gray}\textbf{Valid} (queries)} & 
  \cellcolor{Gray}\textbf{NumRS} & 
  \cellcolor{Gray}\textbf{InfoEX} & 
  \multicolumn{1}{l|}{\cellcolor{Gray}\textbf{LogcRS}} & 
  \multicolumn{1}{l|}{\textbf{Corpus}} & 
  \multicolumn{1}{l|}{\textbf{Dataset}} & 
  \multicolumn{1}{l|}{\cellcolor{Gray}\textbf{Test} (queries)} & 
  \cellcolor{Gray}\textbf{NumRS} & 
  \cellcolor{Gray}\textbf{InfoEX} & 
  \multicolumn{1}{l|}{\cellcolor{Gray}\textbf{LogcRS}} & 
  \multicolumn{1}{l|}{\textbf{Corpus}} 
  \\ \hline
  \multicolumn{1}{|c|}{\textbf{FedE4FIN}} & \multicolumn{1}{c|}{43,658} & 
  \multicolumn{1}{c|}{\cellcolor{Gray}50} & 
  \cellcolor{Gray}34\% &
  \cellcolor{Gray}28\% &
  \multicolumn{1}{c|}{\cellcolor{Gray}16\%}
  & 6,656 & \multicolumn{1}{c|}{\textbf{RAG4FIN}} & 
  \multicolumn{1}{c|}{\cellcolor{Gray}100} &
  \cellcolor{Gray}47\% &
  \cellcolor{Gray}23\% &
  \multicolumn{1}{c|}{\cellcolor{Gray}14\%} 
  & 24,323 \\ 
  \specialrule{1pt}{0pt}{0pt}
  \end{tabular}
  }
  \label{datasets_1}
\end{table*}

\begin{figure*}
  \centering
  \begin{minipage}[t]{0.22\linewidth}
    \centering
    \subfigure[Query-Chunk Pairs (Training)]{
      \includegraphics[width=\linewidth]{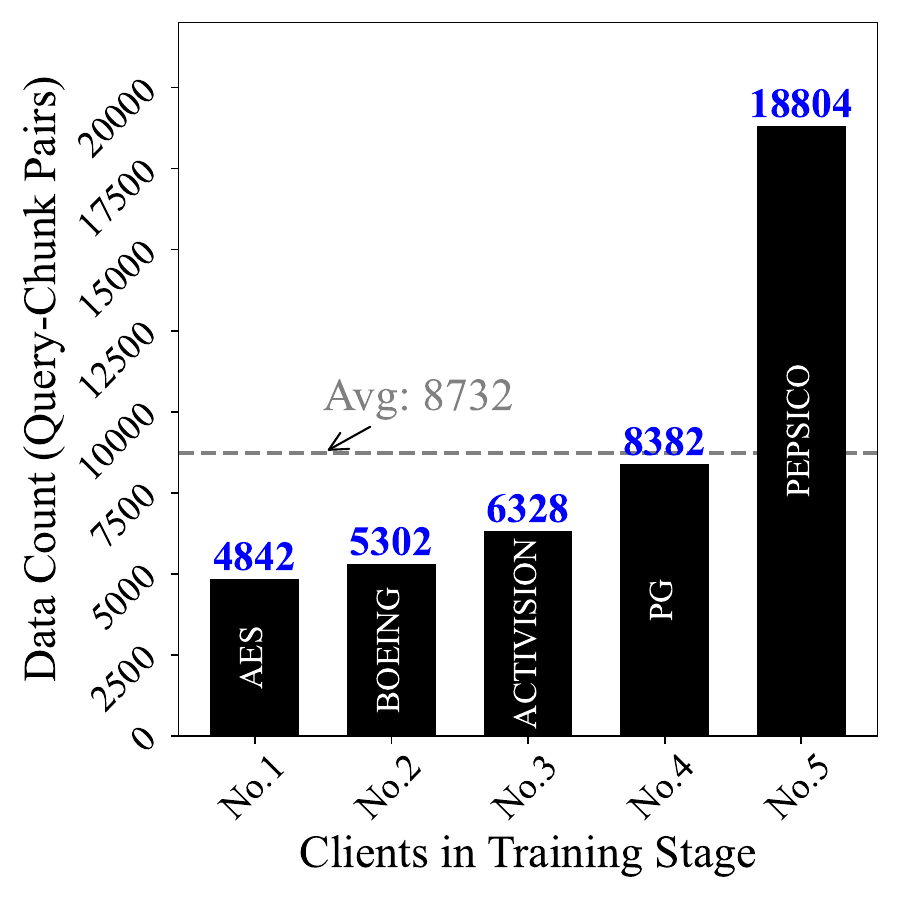}
    }
    \label{clients_in_training_data}
  \end{minipage}
  \hfill 
  \begin{minipage}[t]{0.22\linewidth}
    \centering
    \subfigure[Documents (Training)]{
      \includegraphics[width=\linewidth]{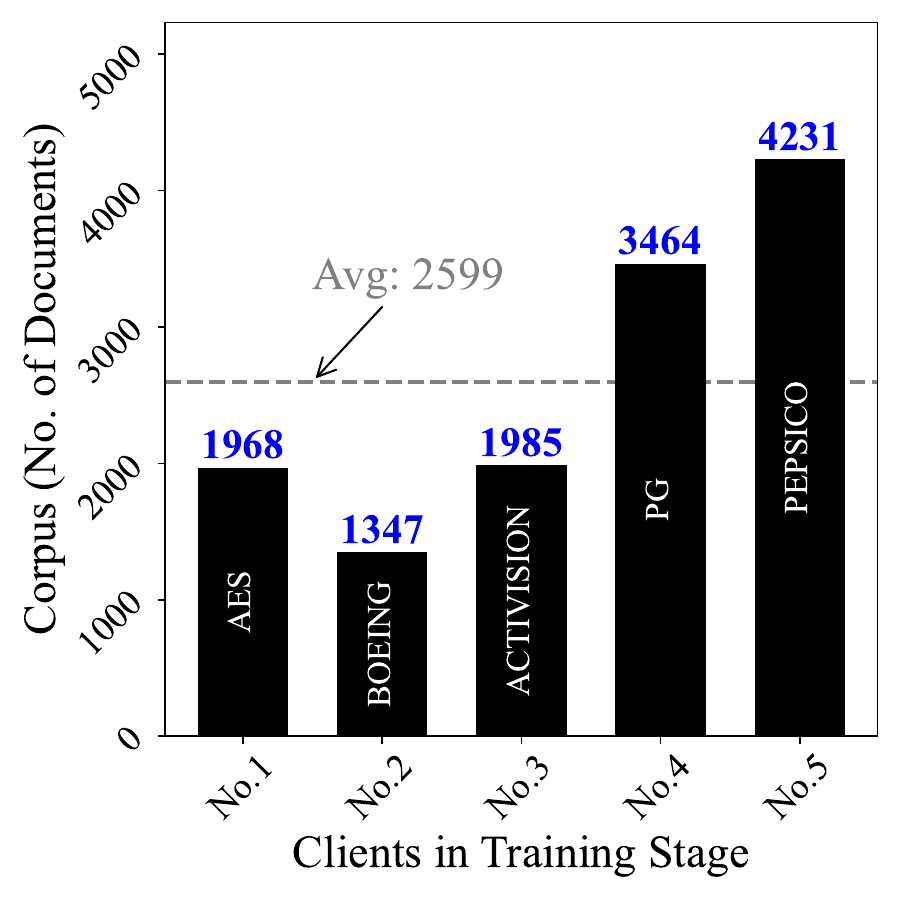}
    }
    \label{clients_in_training_corpus}
  \end{minipage}
  \hfill
  \begin{minipage}[t]{0.22\linewidth}
    \centering
    \subfigure[No. Query Types (Validation)]{
      \includegraphics[width=\linewidth]{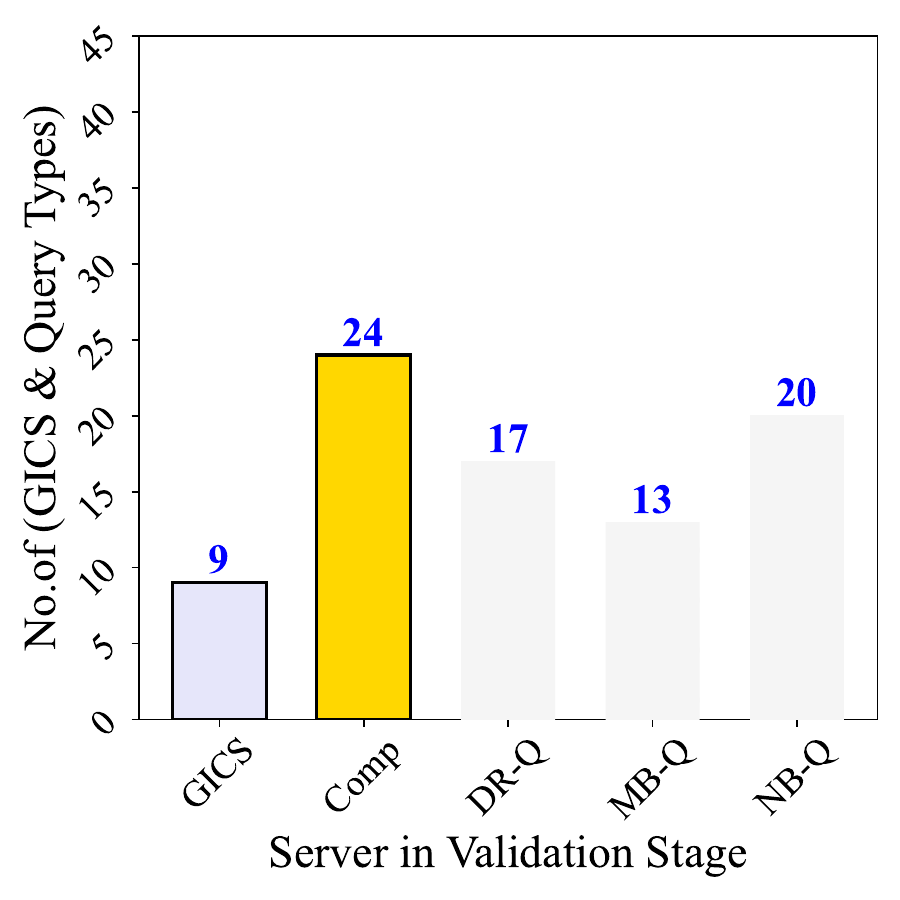}
    }
    \label{sever_in_validation}
  \end{minipage}
  \hfill 
  \begin{minipage}[t]{0.22\linewidth}
    \centering
    \subfigure[No. Query Types (Validation)]{
      \includegraphics[width=\linewidth]{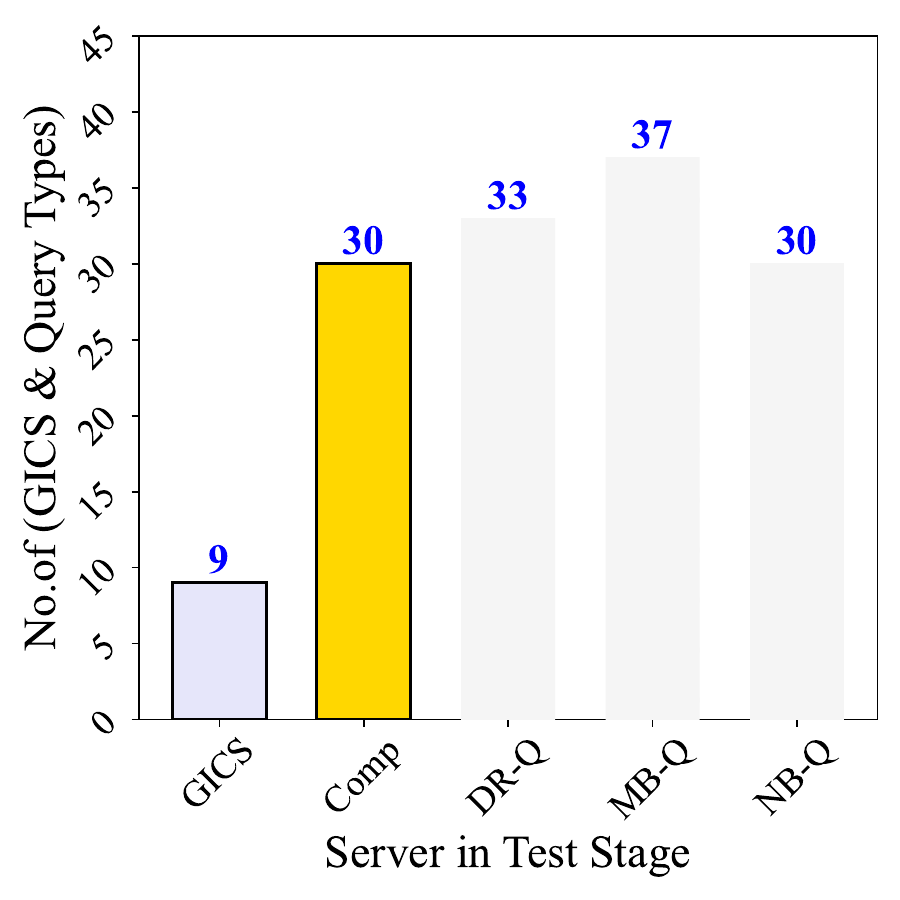}
    }
    \label{sever_in_test}
  \end{minipage}
  \caption{The number of query-chunk pairs in upstream embedding pretraining, and the number of documents provided by each client company. In the training data (subfigures a \& b), there are 5 companies (clients), namely AES (No.1), BOEING (No.2), ACTIVISION (No.3), PG (No.4), and PEPSICO (No.5). These companies provide 8, 8, 9, 13, and 13 documents for federated learning respectively, with a total of 51 documents.
  }
  \label{datasets_distribution}
\end{figure*}


\subsection{Downstream Question \& Answer}\label{sec:DnPRAG}
The Retrieval-Augmented Generation (RAG) framework in the private local environment consists of three primary components: a large language model \(\mathbf{M}\) and its model parameters \(\mathbf{W}\), the retrieval corpora \(\mathbf{D}\), and the retriever \(\mathbf{R}\). 
The system is designed to produce an user answer \(x^{a}\) to the user query \(x^{q}\). The RAG process begins with the retriever \(\mathbf{R}\) identifying the top-\(k\) relevant documents from \(\mathbf{D}\) corresponding to the query \(x^{q}\). Mathematically, this can be represented as:

\begin{equation}
    \mathcal{R}(x^{q}, \mathbf{D})= \left\{ d_{1},d_{2},...,d_{k} \right\}\subseteq \mathbf{D}.
\end{equation}
    
This step typically involves calculating the similarity or distance between the query's embedding \(\textbf{z}^{q}\),  and the embeddings of stored document chunks \(\textbf{z}^{c}_{d_{i}}\). 
For instance, using a k-NN (k-Nearest Neighbors) retriever, the retrieval step can be formulated as: 
\begin{equation}
    \mathcal{R}(x^{q}, \mathbf{D})= \left\{d_{i}\in \mathbf{D} \mbox{ }|\mbox{ }dist (\textbf{z}^{q},\textbf{z}^{c}_{d_{i}}) \mbox{ } \mbox{is in the top} \mbox{ } k \right\}.
\end{equation}

Here, dist\((\textbf{z}_{q},\textbf{z}_{d_{i}})\) quantifies the distance between two embeddings of query and relevant textual chunks. 
The top-\(k\) textual chunks with the smallest distances are then retrieved. Once the relevant textual chunks are retrieved, the RAG integrates the retrieved context \(\mathcal{R}(x^{q}, \mathbf{D})\) with the query  \(x^{q}\) to generate an answer. 
This integration is achieved by concatenating the retrieved documents with the query, forming a combined input for the language model \(\mathbf{M}\). Finally, the system produces the output from \(\mathbf{M}\) as follows:
\begin{equation}
    x^{a}=\mathcal{M}(\mathcal{R}(x^{q}, \mathbf{D}) \parallel x^{q})
\end{equation}

\section{EXPERIMENTS}\label{sec:experi}
In this section, we conduct experiments to evaluate the performance of the proposed framework. 
We will first introduce the used datasets, the evaluation metrics, methods for comparison, and experimental settings. 
Then, we will compare our methods with baselines in upstream retrieval, and provide the analysis in the downstream generation.

\subsection{Datasets Setup}
\subsubsection{Datasets used for Training}
\label{dataset}
In the upstream contrastive learning pretraining process, each client holds financial document data from their respective enterprises (in principle, documents of any content are considered viable to meet the requirements of a real federated learning (FL) setting). The FL process begins with synthetic data generation, where training data is synthesized from the limited local data available, resulting in training pairs of the form [{\footnotesize\texttt{user query, text chunk}}], designed to emulate real-world query-context interactions. Subsequently, the synthetic data is used to fine-tune the BGE embedding model within a contrastive learning framework. We enhance the upstream pre-training dataset by generating synthetic questions from client-provided document texts using the GPT-4o mini model. 
As illustrated in Figure~\ref{datasets_distribution}, the upstream pretraining involves five (No.1 to No.5) corporate clients: AES Corporation,  Boeing Company, Activision Blizzard, PG, and PepsiCo. These clients collectively provide approximately 51 documents, comprising a total of 23,123 document pages. The synthetic data generation process results in 43,658 query-chunk document pairs as shown in Table~\ref{datasets_1} and Figure~\ref{datasets_distribution}.

\subsubsection{Datasets used for Validation and Test}
In practical application, the server-side model can serves as the main model for the FL system~\cite{SoltaniZHL23}, managing and deploying the model for use by companies beyond the initial 5 companies (clients). The experimental analysis focuses solely on the server-side model, which serves as a representation of the average performance of the entire federated framework. This approach enables the evaluation of the model's generalization capability across different companies and industries.

In assessing the performance of pretrained models, we adopt validation and testing methods analogous to those used in RAG retrieval tasks. This necessitates the construction of a vector database based on pretrained embedding models. For the retrieval question answering, we reference the Financebench~\cite{abs-2311-11944}\footnote{~\url{https://github.com/patronus-ai/financebench}} dataset, standardize it to the RAG format, and retain the metadata. 
As shown in Table~\ref{datasets_1}, we utilize the high-quality human-annotated 50 queries from Financebench and pair them with a smaller corpus of approximately 6,656 document pages for validation. Upstream validation employs a smaller corpus for faster indexing, while downstream testing uses the full corpus, increasing retrieval difficulty.

In the test set, we select 100 high-quality human-annotated queries from Financebench and expand the corpus to a larger scale of 24,323 document pages, providing a more comprehensive evaluation environment. 
We designate the upstream pre-training dataset used for contrastive learning to generate the upstream embedding as \textbf{RAG4FIN}, and the downstream RAG dataset as \textbf{FedE4FIN}. 
The upstream data is formatted for contrastive learning as [{\footnotesize \texttt{user query}, \texttt{text chunk}}], while the downstream data is standardized into the RAG format [{\footnotesize\texttt{question}, \texttt{retrieval-context}, \texttt{retrieval-context.ids}, \texttt{golden-contexts},  \texttt{golden-context.ids}, \texttt{expected-answer}, \texttt{actual-response}}]. The golden contexts represent the retrieval annotations, specifically the evidence text for each query, and during testing, the evaluation is based on whether the page number (retrieval-context.ids) containing the evidence is retrieved. The expected answer is human-annotated, while the actual response is generated by the RAG system.

To highlight the non-IID distribution of the data, the selected enterprises in the downstream validation and testing datasets cover 9 GICS\footnote{\url{https://en.wikipedia.org/wiki/Global_Industry_Classification_Standard}} classifications, as shown in Figure~\ref{datasets_distribution}. The validation set consists of 24 companies, while the testing set comprises 30 companies. A comparison of Figure~\ref{datasets_distribution} indicates that the upstream pretraining involves 5 companies across 4 GICS sectors\footnote{These 4 GICS (global industry classification standard) sectors are utilities, industrials, communication services, and consumer staples.}. In contrast, the validation set includes 24 companies and the testing set includes 30 companies, both covering 9 GICS sectors\footnote{These 9 GICS (global industry classification standard) sectors are consumer staples, industrials, health care, consumer discretionary, communication services, financials, information technology, utilities, and materials.}.

According to the metadata of Financebench, we categorize queries into three types as depicted in Figure~\ref{datasets_distribution} (c) \& (d): Domain-related query (DR-Q), Metric-based query  (MB-Q) and Novel-based query (NB-Q). Additionally, we classify query-reasoning into types listed in Table~\ref{datasets_1}: information extraction queries (InfoEX) and complex reasoning queries, which include numerical reasoning (NumRS) and logical reasoning (LogcRS). Notably, a single query that involves both numerical and logical reasoning is categorized in both the NumRS and LogcRS categories. 
Domain-related queries typically pertain to public or financial status information of a company, with answers that include professional explanations, as exemplified by "Does American Water Works have positive working capital based on FY2022 data?".
Metric-based queries encompass 18 specific metrics relevant to the financial sector, such as net income and total revenue, derived from the three principal financial statements in 10Ks (income statement, balance sheet, and cash flow statement). 
Some questions are purely extractive, for instance, "What is Amazon's FY2019 unadjusted operating income (as reported by management)?" Others require additional calculations, involving either a single financial statement or multiple statements, as exemplified by "What is PepsiCo's FY2021 total D\&A (as shown in the cash flow statement) as a percentage of total revenue?"
Novel-based queries present a challenge as they necessitate reasoning beyond mere extraction. Financial experts annotate these queries, leveraging their knowledge and experience to formulate realistic questions, as exemplified by "What drove the reduction in SG\&A expense as a percent of net sales in FY2023?". 

Given that the query type is user-determined and the reasoning process is model-decided (unknown to the user), we configure Q\&A prompts based solely on the query type, as shown in the following three code blocks.


\begin{figure}[!h]
  \begin{tcolorbox}[
    colback=gray!10,
    colframe=gray!166,
    width=1\linewidth, 
    arc=1mm, 
    auto outer arc,
    title={Prompt for Domain-Relevant Query (DR-Q)},
    fonttitle=\footnotesize\bfseries, 
    fontupper=\scriptsize, 
    fontlower=\scriptsize, 
    boxsep=2pt, 
    left=2pt, 
    right=2pt, 
    top=2pt, 
    bottom=2pt, 
    ]
          \scriptsize{ 
            Below is the context information: \{context\_str\}      
      
            Based solely on the above context, not prior knowledge, please answer the following question: \{query\_str\}  
    
            Instructions:
            \begin{itemize}
                \item First, determine if the statement is correct (Yes/No).
                \item Second, provide a concise justification for your determination.
                \item Third, if the statement is false, provide the correct information.
            \end{itemize}
          }
    \end{tcolorbox}
      \vspace{-0.3in}
\end{figure}

\begin{figure}[!h]
  \begin{tcolorbox}[
    colback=gray!10,
    colframe=gray!166,
    width=1\linewidth, 
    arc=1mm, 
    auto outer arc,
    title={Prompt for Metrics-based Query (MB-Q)},
    fonttitle=\footnotesize\bfseries, 
    fontupper=\scriptsize, 
    fontlower=\scriptsize, 
    boxsep=2pt, 
    left=2pt, 
    right=2pt, 
    top=2pt, 
    bottom=2pt, 
    ]
          \scriptsize{ 
            Below is the context information: \{context\_str\}
    
            Based solely on the above context, not prior knowledge, please answer the following question: \{query\_str\}            
  
            Instructions:
            \begin{itemize}
                \item Output ONLY the final numerical result.
                \item Strictly maintain two decimal places.
                \item Use numeric format.
                \item Do not include explanations or other text.
                \item Ensure complete calculation before rounding.
            \end{itemize}
          }
    \end{tcolorbox}
    \vspace{-0.3in}
\end{figure}

\begin{figure}[!h]
    \begin{tcolorbox}[
      colback=gray!10,
    colframe=gray!166,
    width=1\linewidth, 
    arc=1mm, 
    auto outer arc,
    title={Prompt for Novel-based Query (DR-Q)},
    fonttitle=\footnotesize\bfseries, 
    fontupper=\scriptsize, 
    fontlower=\scriptsize, 
    boxsep=2pt, 
    left=2pt, 
    right=2pt, 
    top=2pt, 
    bottom=2pt, 
      ]
          \scriptsize{
            Below is the context information:  \{context\_str\}
    
            Based solely on the above context, not prior knowledge, please answer the following question: \{query\_str\}    
          
            Instructions:
            Keep your answer extremely brief. Focus only on the most essential information from the context.
          }
    \end{tcolorbox}
    \vspace{-0.3in}
\end{figure}

\subsection{Evaluation Setup}
\subsubsection{Upstream Evaluation Metrics} For upstream RAG evaluation, we  evaluate retrieval performance using metrics adapted from XRAG~\cite{abs-2412-15529}\footnote{~\url{https://github.com/DocAILab/XRAG}}, categorized as follows:
\begin{itemize}[leftmargin=*]
  \item \textbf{Presence-Based Metrics}: Assess the presence of correct results in the top k outputs, regardless of ranking, including Hit@1, Hit@10, EM (Exact Match). We usemploy  `extract match' as the retrieval metric, indicating success if any result matches the ground truth ({\footnotesize\texttt{golden-context.ids}}), irrespective of the ranking order. 
  \item \textbf{Order-Sensitive Metrics}: Evaluate the quality of ranked results, emphasizing positional relevance, including  Mean Reciprocal Rank (MRR~\cite{RadevQWF02}), Mean Average Precision (MAP~\cite{CarteretteV11}), and DCG family, which evaluates the effectiveness of ranking models by assessing the quality of ordered results ({\footnotesize \texttt{retrieval-context.ids}}). This family includes Discounted Cumulative Gain (DCG), Normalized Discounted Cumulative Gain (NDCG), and Ideal Discounted Cumulative Gain (IDCG). IDCG represents the maximum DCG achievable if the results are ideally ranked in descending order ({\footnotesize \texttt{golden-context.ids}}) of relevance.
  \item \textbf{Threshold-Driven Metrics}: Balance precision and recall using similarity thresholds (e.g., cosine similarity). They include F1, Accuracy@k (Acc@k): the percentage of top-k results where the cosine similarity exceeds a predefined threshold (indicating correct predictions), Recall@k (Rec@k): the proportion of all truly relevant documents captured within the top-k results (reflecting coverage), and Precision@k (Pre@k): the fraction of the top-k results that are truly relevant (indicating result quality).
\end{itemize}

\subsubsection{Downstream Evaluation Metrics} For downstream RAG testing in the RAG4FIN dataset, we employ a range of natural language generation metrics implemented by XRAG~\cite{abs-2412-15529}. These metrics are categorized as follows:
\begin{itemize}[leftmargin=*]
  \item \textbf{N-gram Overlap based Metrics}: Measure lexical or character-level overlap between generated and reference texts, commonly used in tasks like machine translation and text generation BLEU~\cite{PapineniRWZ02} computes n-gram precision between generated and reference texts, emphasizing lexical overlap. ChrF~\cite{Popovic15} and ChrF\(^{++}\)~\cite{Popovic17}  use the F-score statistic for character n-gram matches. ChrF++ additionally incorporates word n-grams to improve correlation with human assessments. METEOR~\cite{BanerjeeL05} evaluates unigram precision, recall, and fragmentation while incorporating synonym matching and word order alignment. ROUGE~\cite{lin2004rouge} measures overlap through unigrams (ROUGE-1), bigrams (ROUGE-2), and the longest common subsequence (ROUGE-L).
\item \textbf{LLM Intrinsic Evaluation}: Assess the predictive quality of language models themselves. PPL (Perplexity)~\cite{jelinek1977perplexity} quantifies the model’s uncertainty via the exponentiated average negative log-likelihood of the generated sequence. 
\item \textbf{Edit Distance based Metrics}: These quantify the number of edits (insertions, deletions, substitutions) required to align generated text with references, including WER (word error rate), word-level edit distance and CER (character error rate), character-level edit distance. 
\end{itemize}


\begin{table*}[t]
  \centering
  \scriptsize
  \caption{Comparative \colorbox{blue!20}{\textcolor{blue}{\textbf{test}}} performance of baselines and our FedE4RAG on the \textcolor{blue}{\textbf{upstream RAG retrieval}} tasks. The \textbf{best} results for each metric are highlighted in bold. 
  `\underline{X}' indicates results exceeding ScenT, the strongest non-federated model trained centrally with all data.}
  \renewcommand\arraystretch{1.6}
  \setlength{\tabcolsep}{1.mm}{
    \begin{tabular}{|l|c|c|c|c|c|c|c|c|c|c|c|c|c|c|c|c|c|c|c|c|c|c|c|c|c|c|}
      \specialrule{1pt}{0pt}{0pt}
      \textbf{Methods} & \multicolumn{3}{c|}{\textbf{Presence-Based}} & \multicolumn{5}{c|}{\textbf{Order-Sensitive}} & \multicolumn{10}{c|}{\textbf{Threshold-Driven}} \\
      \cline{2-19}
      & \textbf{Hit@1} & \textbf{Hit@10} & \textbf{EM} & \textbf{MRR} & \textbf{MAP} & \textbf{NDCG} & \textbf{DCG}  & \textbf{IDCG} & \textbf{F1} & \textbf{Acc@1}  & \textbf{Acc@5} &  \textbf{Acc@10} & \textbf{Rec@1} & \textbf{Rec@5} & \textbf{Rec@10} & \textbf{Pre@1} & \textbf{Pre@5} & \textbf{Pre@10} \\
      \specialrule{1pt}{0pt}{0pt}
      \textbf{SCenT}   
      & \textit{70.00} & \textit{77.00} & \textit{36.00} & \textit{51.81} & \textit{85.70} & \textit{57.51} & \textit{64.09} & \textit{53.54} & \textit{40.73} & \textit{43.00} & \textit{63.00} & \textit{77.00} & \textit{37.50} & \textit{57.83} & \textit{72.17} & \textit{43.00} & \textit{13.85} & \textit{09.08} \\
      \textbf{ICMaxD} 
      & 68.00 & 74.00 & 31.00 & 49.58 & 82.83 & 54.89 & 61.78 & 51.03 & 36.73 & 40.00 & 61.00 & 74.00 & 34.50 & 56.67 & 70.67 & 40.00 & 14.05 & 08.83 \\
      \textbf{ICMinD}  
      & 68.00 & 72.00 & 30.00 & 48.77 & 81.33 & 53.71 & 60.92 & 50.11 & 35.73 & 39.00 & 60.00 & 72.00 & 33.50 & 56.17 & 69.17 & 39.00 & 14.05 & 08.57 \\
      \textbf{SVanE1.5\(_{\rm base}\)} 
      & 69.00 & 72.00 & 29.00 & 49.95 & 81.96 & 53.37 & 61.22 & 49.95 & 35.23 & 38.00 & 59.00 & 72.00 & 32.50 & 55.67 & 69.67 & 38.00 & 13.85 & 08.68 \\
      \textbf{SVanE1.5\(_{\rm large}\)} 
      & 61.00 & 70.00 & 33.00 & 48.48 & 75.67 & 53.15 & 57.61 & 49.20 & 37.33 & 40.00 & 58.00 & 70.00 & 36.00 & 54.33 & 64.33 & 40.00 & 12.85 & 07.81 \\ 
      \textbf{FedAvg} 
      & 68.00 & 72.00 & 31.00 & 49.29 & 81.33 & 54.15 & 61.39 & 50.61 & 36.73 & 40.00 & 61.00 & 72.00 & 34.50 & 56.67 & 69.17 & 40.00 & 14.05 & 08.57 \\
      \rowcolor{Gray} \textbf{FedE4RAG} 
      & \textbf{\underline{73.00}} & \textbf{\underline{79.00}} & \textbf{\underline{36.00}} & \textbf{\underline{55.39}} & \textbf{\underline{90.22}} & \textbf{\underline{60.34}} & \textbf{\underline{69.19}} & \textbf{\underline{57.86}} & \textbf{\underline{42.30}} & \textbf{\underline{45.00}} & \textbf{\underline{71.00}} & \textbf{\underline{79.00}} & \textbf{\underline{38.50}} & \textbf{\underline{65.83}} & \textbf{\underline{75.67}} & \textbf{\underline{45.00}} & \textbf{\underline{16.25}} & \textbf{\underline{09.70}} \\
      \specialrule{1pt}{0pt}{0pt}
    \end{tabular}
    }
    \label{main_retrieval_test}
\end{table*}

\begin{table*}[t]
  \centering
  \scriptsize
  \caption{Comparative \colorbox{gray!10}{\textcolor{blue}{\textbf{validation}}} performance of baselines and our FedE4RAG on the \textcolor{blue}{\textbf{upstream RAG retrieval}} tasks. The \textbf{best} results for each metric are highlighted in bold. 
  `\underline{X}' indicates results exceeding ScenT, the strongest non-federated model trained centrally with all data.}
  \renewcommand\arraystretch{1.6}
  \setlength{\tabcolsep}{1.mm}{
    \begin{tabular}{|l|c|c|c|c|c|c|c|c|c|c|c|c|c|c|c|c|c|c|}
      \specialrule{1pt}{0pt}{0pt}
      \textbf{Methods} & \multicolumn{3}{c|}{\textbf{Presence-Based}} & \multicolumn{5}{c|}{\textbf{Order-Sensitive}} & \multicolumn{10}{c|}{\textbf{Threshold-Driven}} \\
      \cline{2-19}
      & \textbf{Hit@1} & \textbf{Hit@10} & \textbf{EM} & \textbf{MRR} & \textbf{MAP} & \textbf{NDCG} & \textbf{DCG}  & \textbf{IDCG} & \textbf{F1} & \textbf{Acc@1}  & \textbf{Acc@5} &  \textbf{Acc@10} & \textbf{Rec@1} & \textbf{Rec@5} & \textbf{Rec@10} & \textbf{Pre@1} & \textbf{Pre@5} & \textbf{Pre@10} \\
      \specialrule{1pt}{0pt}{0pt}
      \textbf{SCenT}   
      & \textit{87.00} & \textit{89.00} & \textit{51.00} & \textit{69.13} & \textit{72.73} & \textit{73.52} & \textit{86.51} & \textit{1.0481} & \textit{58.33} & \textit{60.00} & \textit{82.00} & \textit{89.00} & \textit{53.00} & \textit{79.17} & \textit{88.00} & \textit{60.00} & \textit{20.25} & \textit{11.63} \\
      \textbf{ICMaxD} 
      & 86.00 & 88.00 & 49.00 & 68.31 & 71.94 & 72.70 & 85.39 & 1.0381 & 56.33 & 58.00 & 81.00 & 88.00 & 51.00 & 77.67 & 87.00 & 58.00 & 19.65 & 11.54 \\
      \textbf{ICMinD}  
      & 86.00 & 88.00 & 48.00 & 68.15 & 71.85 & 72.89 & 85.82 & 1.0381 & 55.83 & 57.00 & 80.00 & 88.00 & 50.00 & 76.67 & 87.00 & 57.00 & 19.45 & 11.63 \\
      \textbf{SVanE1.5\(_{\rm base}\)} 
      & \textbf{87.00} & \textbf{89.00} & 48.00 & 67.20 & 70.93 & 72.28 & 85.19 & \textbf{1.0481} & 54.83 & 56.00 & 81.00 & \textbf{89.00} & 49.00 & 77.67 & \textbf{88.00} & 56.00 & 19.65 & 11.65 \\
      \textbf{SVanE1.5\(_{\rm large}\)} 
      & 86.00 & \textbf{89.00} & 50.00 & 70.69 & 73.50 & 73.83 & 85.86 & 1.0354 & 58.56 & \textbf{\underline{62.00}} & 81.00 & \textbf{89.00} & \textbf{\underline{54.33}} & 78.00 & 87.00 & 62.00 & 19.60 & 11.30 \\
      \textbf{FedAvg} 
      & 86.00 & 88.00 & 49.00 & 68.35 & 72.02 & 72.83 & 85.68 & 1.0381 & 56.33 & 58.00 & 81.00 & 88.00 & 51.00 & 77.67 & 87.00 & 58.00 & 19.65 & 11.54 \\
      \rowcolor{Gray} \textbf{FedE4RAG} 
      & \textbf{87.00} & \textbf{89.00} & \textbf{\underline{52.00}} & \textbf{\underline{71.07}} & \textbf{\underline{74.94}} & \textbf{\underline{75.00}} & \textbf{\underline{88.60}} & \textbf{1.0481} & \textbf{\underline{60.16}} & 61.00 & \textbf{\underline{84.00}} & \textbf{89.00} & 53.50 & \textbf{\underline{81.67}} & \textbf{88.00} & \textbf{\underline{61.00}} & \textbf{\underline{20.90}} & \textbf{\underline{11.69}} \\
      \specialrule{1pt}{0pt}{0pt}
    \end{tabular}
    }
    \label{main_retrieval_validation}
\end{table*}

\subsection{Implementation Details}
We utilize FLGo~\cite{abs-2306-12079}\footnote{\url{https://github.com/WwZzz/easyFL}} to construct a communication topology between five enterprise clients and the server for federated learning. In the upstream federated embedding pre-training phase, each client trains on local data in batches. Our experiments employ a batch size of 16, determined as optimal from trials with sizes of 8, 16, 32. The learning rate is established at 1e-5, and the number of federated communication rounds is set to 25, with 22 rounds generally found optimal based on validation set outcomes. During each mini-batch of upstream embedding pre-training, we choose one question along with its corresponding single golden context to serve as a positive sample. Concurrently within the same mini-batch, the context paired with other questions serves as negative samples for the contrastive learning training.

For financial dataset pre-training, each client uses the pre-trained BGE  ({\footnotesize \texttt{BAAI/bge-base-en-v1.5}}\footnote{\url{https://huggingface.co/BAAI/bge-base-en-v1.5}}) model. We prefer the base model over the large model ({\footnotesize \texttt{BAAI/bge-large-en-v1.5}}\footnote{\url{https://huggingface.co/BAAI/bge-large-en-v1.5}}) due to their comparable performance (promoted as having abilities comparable to the BGE-large by Xiao et al.,~\cite{xiao2023c}) and the base model's smaller size (438 MB, 0.32 times the BGE-large's 1.34 GB), which reduces training costs and time. 

During downstream generation, the input to the LLM supports maximum sentence chunks of 2048 tokens for each query-retrieved context. 
The model uses a low temperature (0.1) for deterministic output and a top-p value of 0.9 to balance diversity and plausibility. 
Repetition is not penalized (repeat penalty \(=\) 1.0) within the last 64 tokens, and the model generates up to 512 tokens per output.


\subsection{Compared Methods for Upstream Retrieval}
We use the following comparative strategies: (i) Independent Client with 
Maximum Data Volume (\(\rm{\textbf{ICMaxD}}\)): Performance of the client with the most data, trained solely on its data. (ii)  Independent Client with  Minimum Data Volume (\(\rm{\textbf{ICMinD}}\)): Performance of the client with the least data, trained solely on its data. Monotonic results for (i) and (ii) indicate the upper and lower limits of individual performance. (iii) Sever with Vanilla BGE (BGE V1.5 \& BGE M3), refer to (\(\rm{\textbf{SVanE1.5\(_{\rm base}\)}} \& \rm{\textbf{SVanE1.5\(_{\rm large}\)}}\)): Performance of sever is conducted testing using an unadjusted model, applying BGE directly without modifications. 
(iv) Sever with Centralized Training (\(\rm{\textbf{SCenT}}\)): Performance of a model trained by aggregating all client data. This method theoretically achieves optimal performance by using the entire dataset. (v) \(\rm{\textbf{FedAvg}}\)~\cite{WangF0WWY21}: the Federated average embedding aggregation method. (vi) \(\rm{\textbf{FedE4RAG}}\): Our framework tested on the server model representing average client performance.

\subsection{Compared Methods for Downstream Generation}
To evaluate downstream question-answering generation performance, we select LLMs that are compatible with most local deployment scenarios due to their open-source nature and reduced parameter count. We employ our FedE4RAG as the fixed retrieval model. The selected large models include \textbf{Llama3.1-8B}\footnote{\url{https://huggingface.co/meta-llama}} from Meta AI, which are foundational models with 8 billion parameters. Additionally, \textbf{GPT-4o Mini}\footnote{\url{https://openai.com/index/gpt-4o-system-card/}}, a hypothetical model from OpenAI, represents a smaller version of a GPT model intended for wider applications. 
We have also introduced two reasoning LLMs to address queries involving mathematical and logical reasoning. \textbf{MathStral}\footnote{\url{https://huggingface.co/mistralai/Mathstral-7B-v0.1}}, a 7 billion parameter model developed by Mistral AI, is tailored for mathematical reasoning and scientific discovery. Additionally, \textbf{Deepseek R1 7B}\footnote{\url{https://huggingface.co/deepseek-ai/DeepSeek-R1}} from DeepSeek AI, with 7 billion parameters, specializes in reasoning LLM.


\section{Experimental Analysis}
To facilitate performance comparison, we select the server-side model of the FedE4RAG framework for validation and testing, consistent with other federated frameworks~\cite{SoltaniZHL23}. We establish five comparison strategies: 
\begin{itemize}[leftmargin=*]
  \item \texttt{Strategy A}: FedE4RAG vs. ICMaxD \& ICMinD,
  \item \texttt{Strategy B}: FedE4RAG vs. SVanE1.5\(_{\rm base}\),
  \item \texttt{Strategy C}: FedE4RAG vs. SVanE1.5\(_{\rm large}\),
  \item \texttt{Strategy D}: FedE4RAG vs. SCenT, 
  \item \texttt{Strategy E}: FedE4RAG vs. FedAvg.
\end{itemize}

Comparisons with \texttt{Strategy A} evaluate whether the server-side model effectively leverages local data while maintaining performance comparable to individual client training. \texttt{Strategy B} assesses whether our framework exceeds the performance of the BGE model. Conversely, \texttt{Strategy C} examines if employing BGE-base as the backbone for FedE4RAG yields superior results compared to the standard large version of BGE. 
\texttt{Strategy D} determines whether our model matches the performance of non-federated learning models by comparing it to a centralized approach that trains a global embedding using aggregated local datasets from all clients, without data privacy protection. \texttt{Strategy E} evaluates our federated learning method, which includes teacher-student distillation and homomorphic encryption, against the traditional FedAvg approach. The goal is to demonstrate our method's advantages in performance and privacy. If FedE4RAG matches FedAvg's performance, it shows the effectiveness of teacher-student distillation in FedE4RAG's communication and optimization.

\begin{figure*}[tb]
  \hspace{-0.07cm}
  \subfigure[Hit@1]{
    \begin{minipage}[t]{0.12\linewidth}
    \includegraphics[width=0.8in]{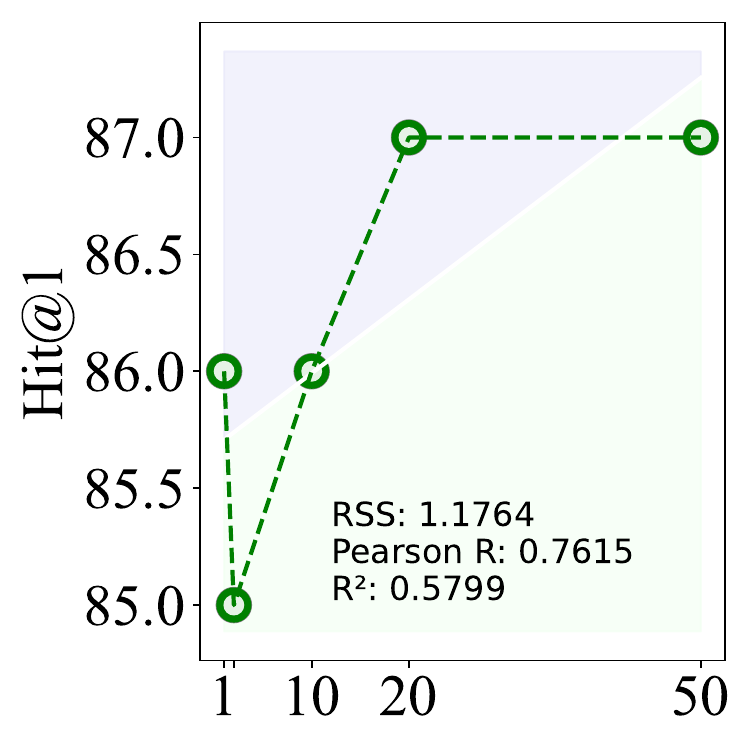}
    \end{minipage}%
    }%
  \subfigure[F1]{
  \begin{minipage}[t]{0.12\linewidth}
  \includegraphics[width=0.8in]{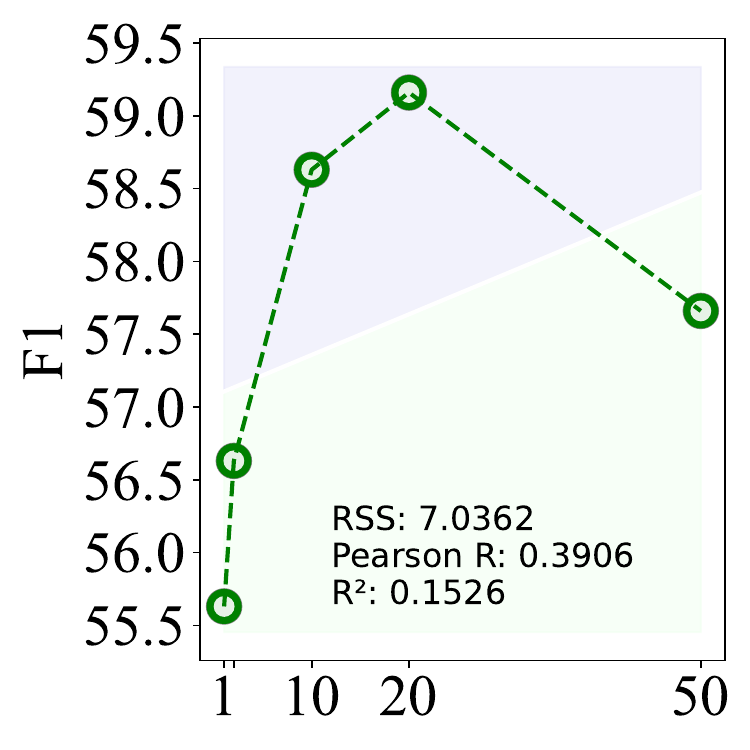}
  \label{fig:side:F1}
  \end{minipage}%
  }%
  \subfigure[MRR@10]{
  \begin{minipage}[t]{0.12\linewidth}
  \includegraphics[width=0.8in]{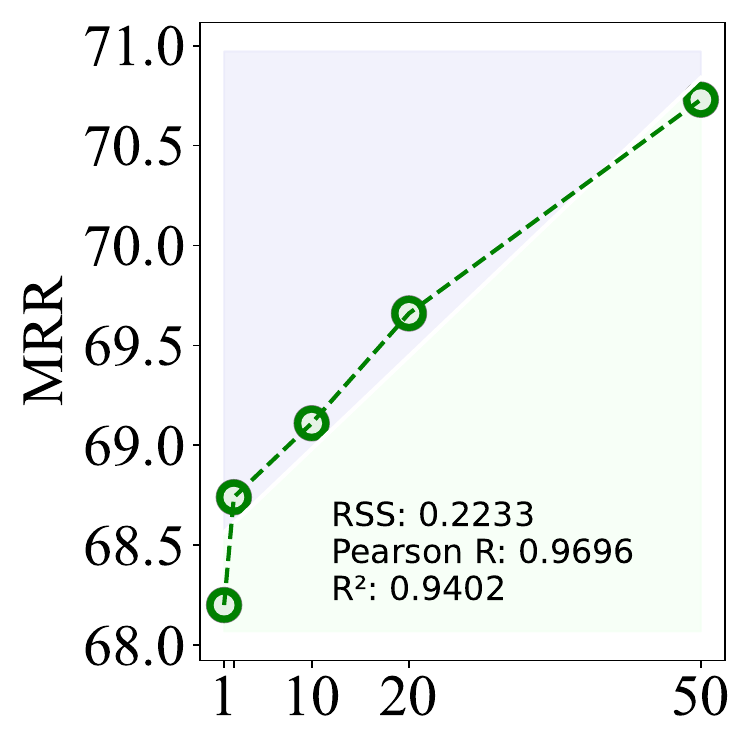}
  \label{fig:side:MRR}
  \end{minipage}%
  }%
  \subfigure[EM]{
  \begin{minipage}[t]{0.12\linewidth}
  \includegraphics[width=0.8in]{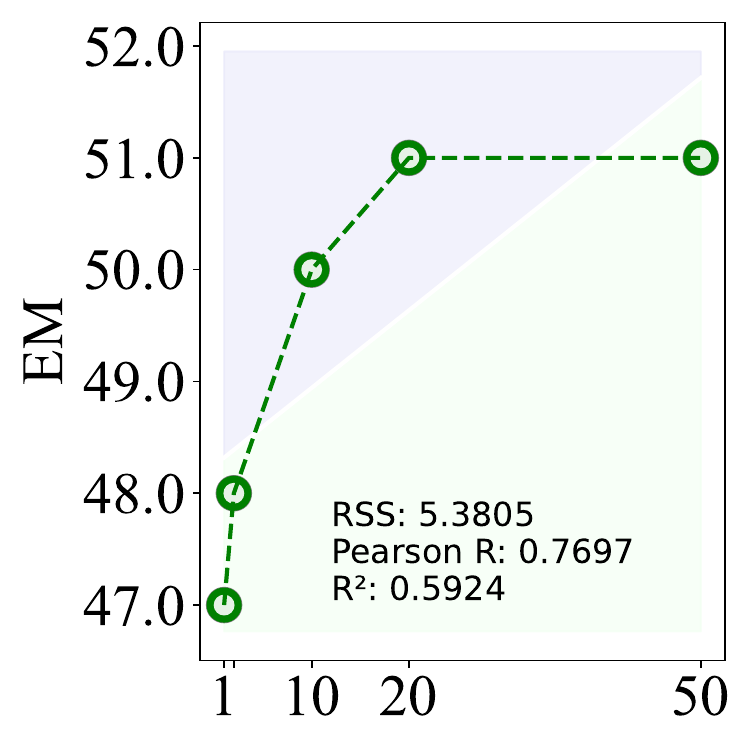}
  \end{minipage}%
  }%
  \subfigure[DCG@10]{
  \begin{minipage}[t]{0.12\linewidth}
  \includegraphics[width=0.8in]{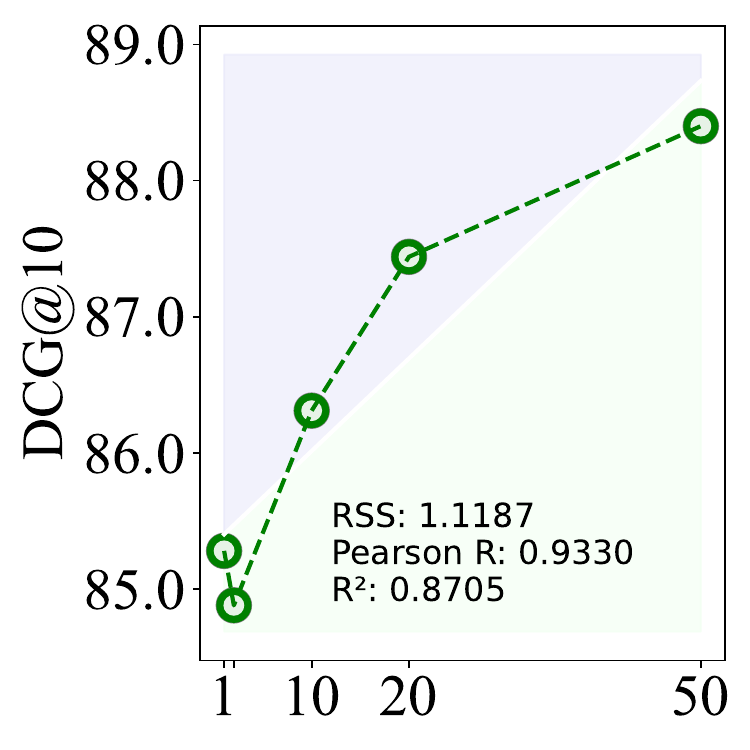}
  \label{fig:side:DCG}
  \end{minipage}%
  }%
  \subfigure[NDCG@10]{
  \begin{minipage}[t]{0.12\linewidth}
  \includegraphics[width=0.8in]{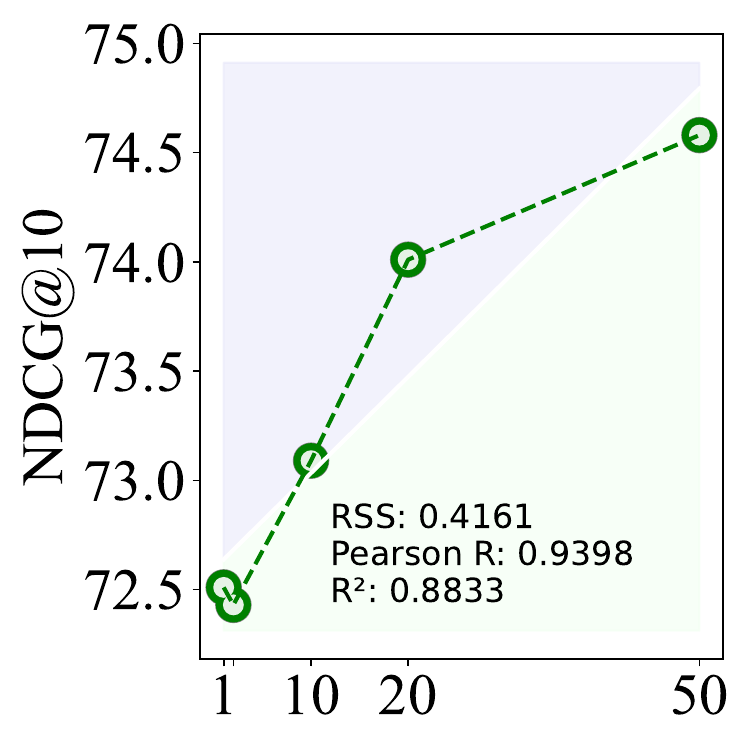}
  \label{fig:side:NDCG}
  \end{minipage}%
  }%
  \subfigure[MAP]{
  \begin{minipage}[t]{0.12\linewidth}
    \includegraphics[width=0.8in]{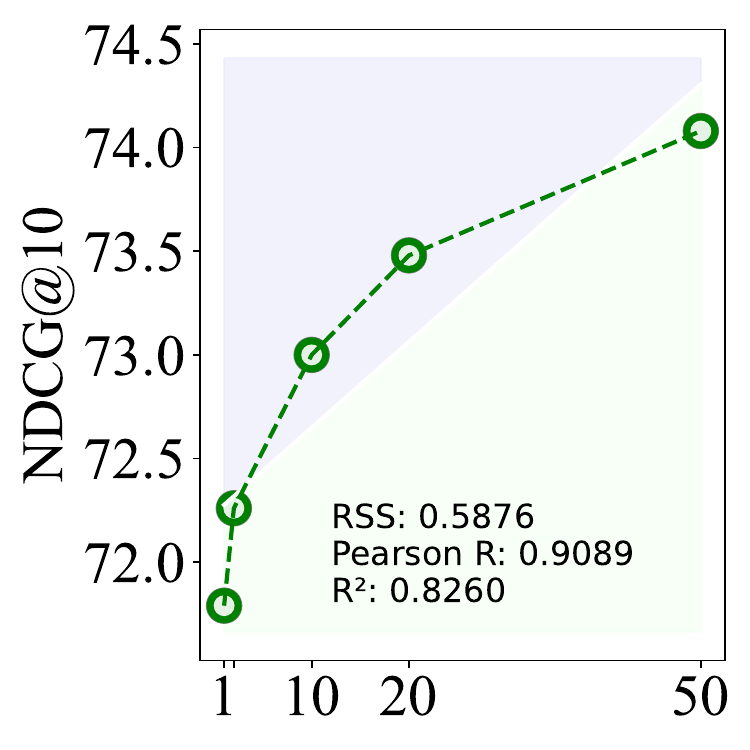}
    \end{minipage}%
  }%
  \subfigure[Acc@1]{
  \begin{minipage}[t]{0.12\linewidth}
      \includegraphics[width=0.8in]{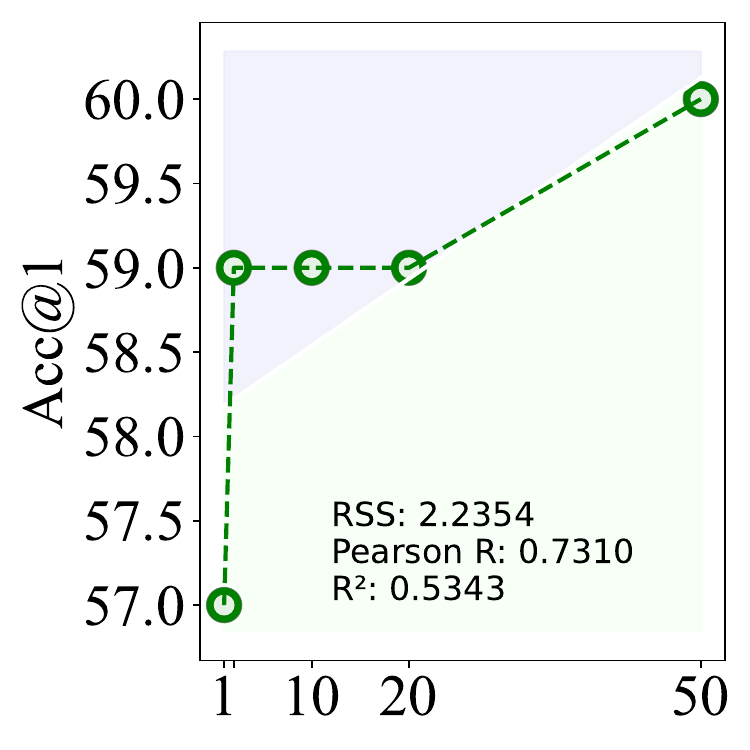}
      \label{fig:side:Acc@1}
  \end{minipage}%
  }%

  \subfigure[Acc@5]{
  \begin{minipage}[t]{0.12\linewidth}
        \includegraphics[width=0.8in]{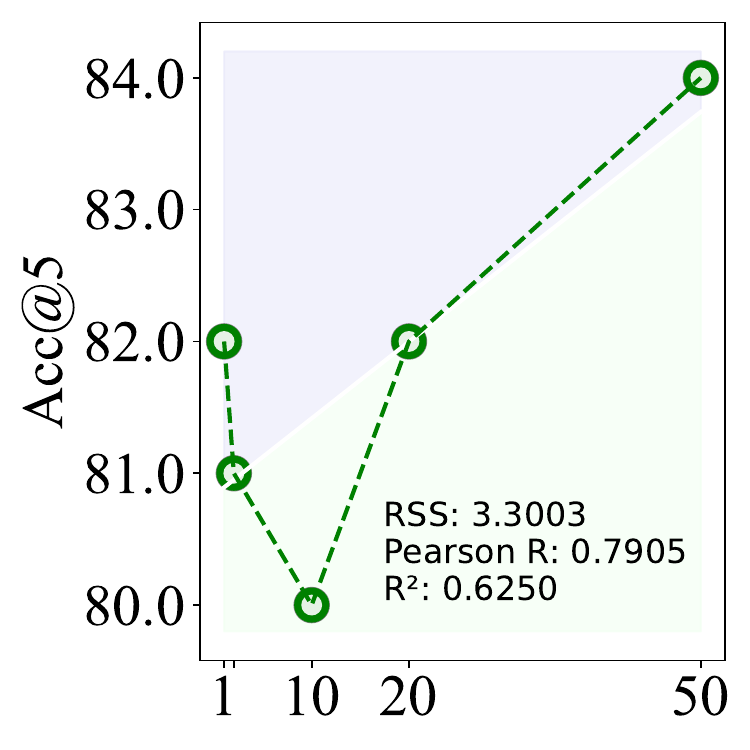}
        \label{fig:side:Acc@5}
  \end{minipage}%
  }%
  \subfigure[Acc@10]{
  \begin{minipage}[t]{0.12\linewidth}
          \includegraphics[width=0.8in]{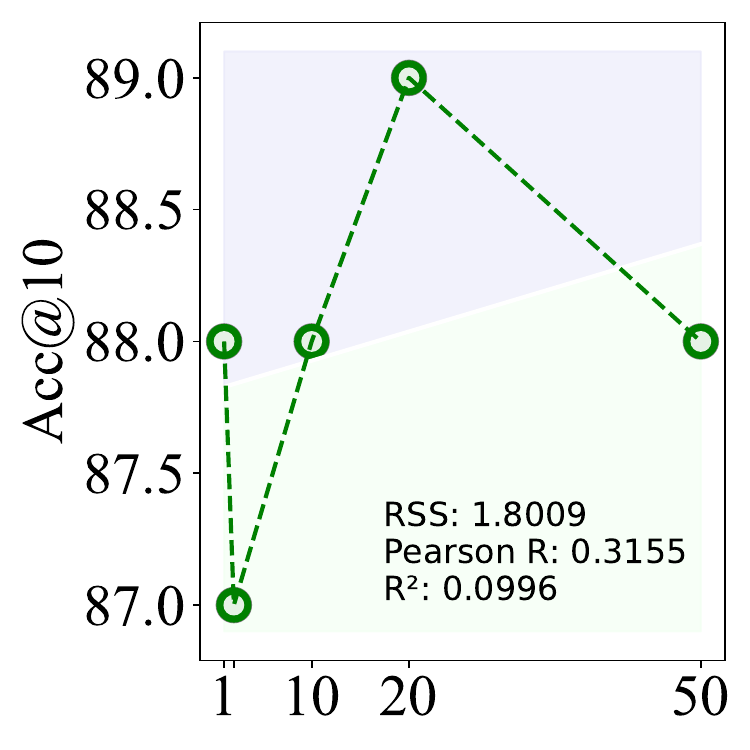}
          \label{fig:side:Acc@10}
  \end{minipage}%
  }%
  \subfigure[Rec@1]{
  \begin{minipage}[t]{0.12\linewidth}
  \includegraphics[width=0.8in]{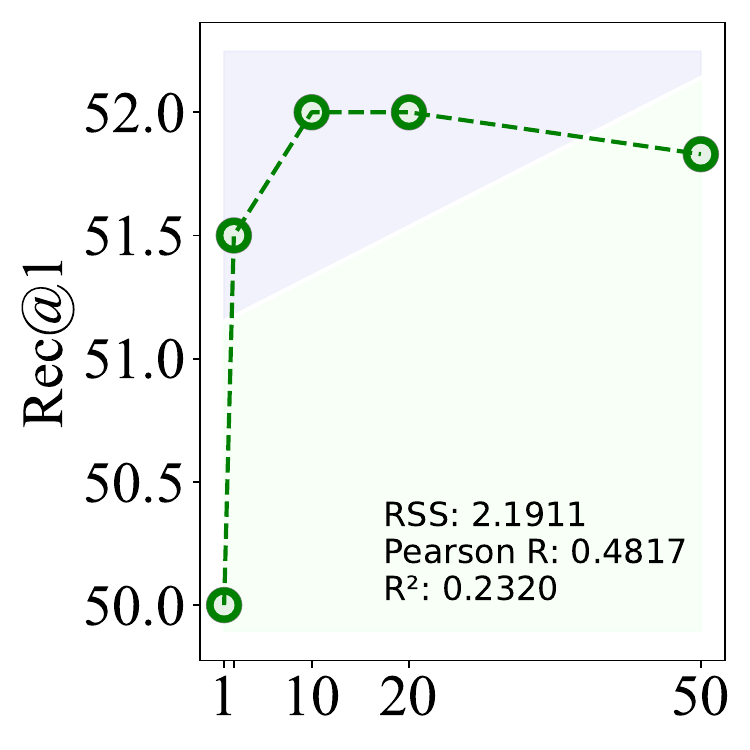}
            \end{minipage}%
  }%
  \subfigure[Rec@5]{
  \begin{minipage}[t]{0.12\linewidth}
  \includegraphics[width=0.8in]{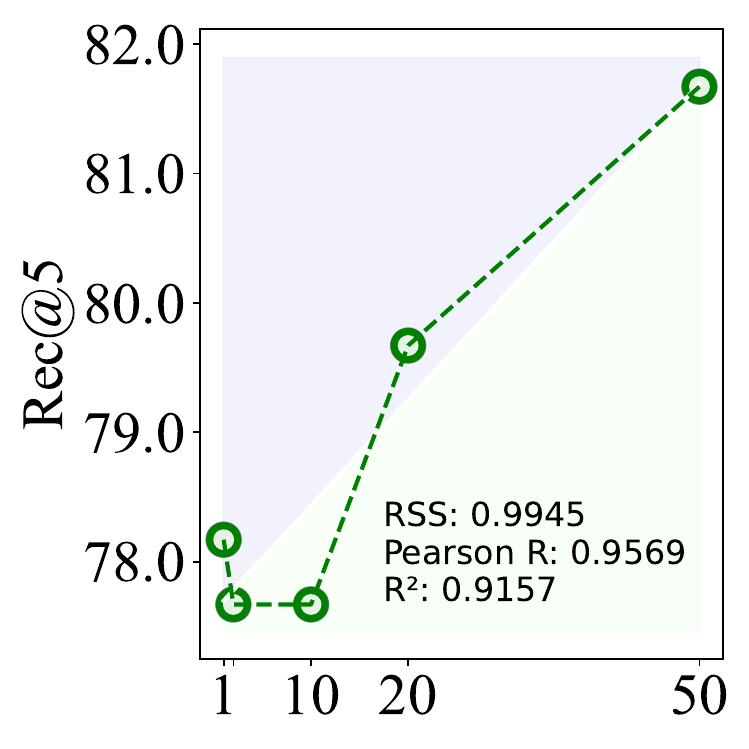}
            \label{fig:side:rec5}
            \end{minipage}%
  }%
  \subfigure[Rec@10]{
  \begin{minipage}[t]{0.12\linewidth}
  \includegraphics[width=0.8in]{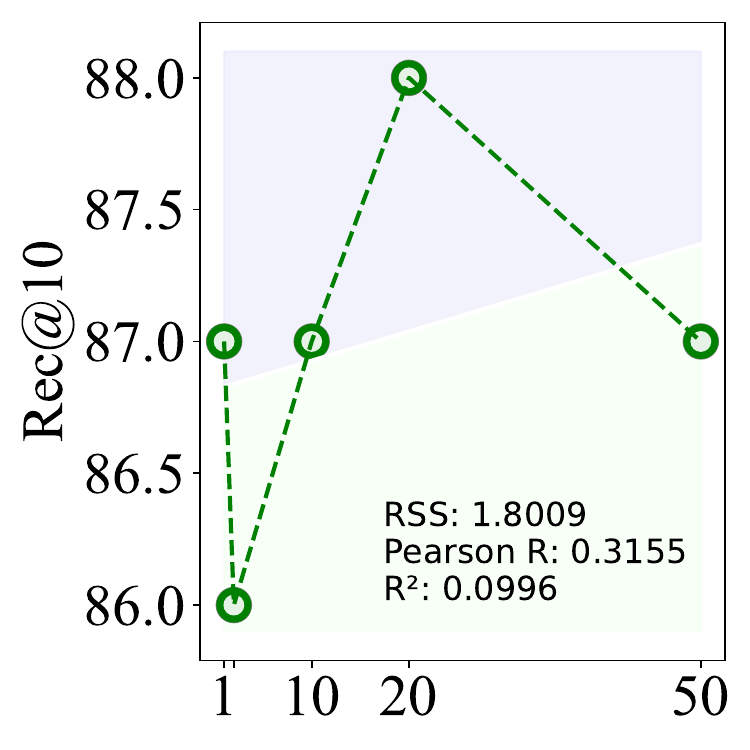}
            \end{minipage}%
  }%
  \subfigure[Pre@1]{
  \begin{minipage}[t]{0.12\linewidth}
  \includegraphics[width=0.8in]{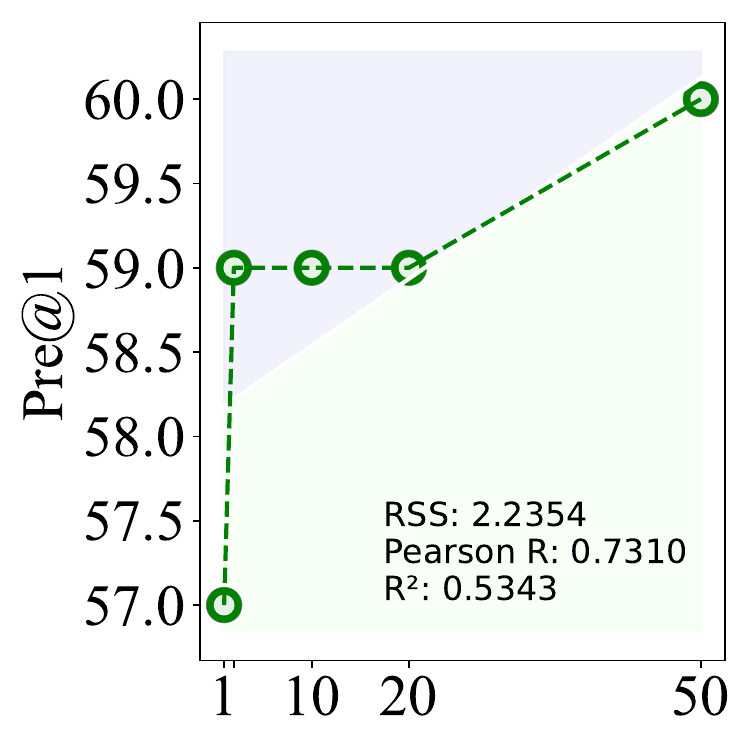}
            \end{minipage}%
  }%
  \subfigure[Pre@5]{
  \begin{minipage}[t]{0.12\linewidth}
  \includegraphics[width=0.8in]{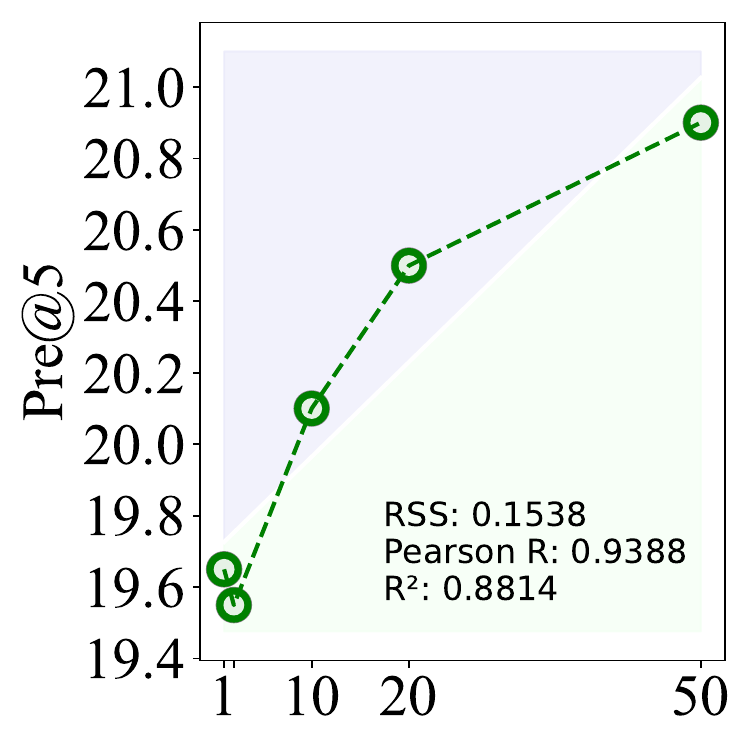}
            \label{fig:side:pre5}
            \end{minipage}%
  }%
  \subfigure[Pre@10]{
  \begin{minipage}[t]{0.12\linewidth}
  \includegraphics[width=0.8in]{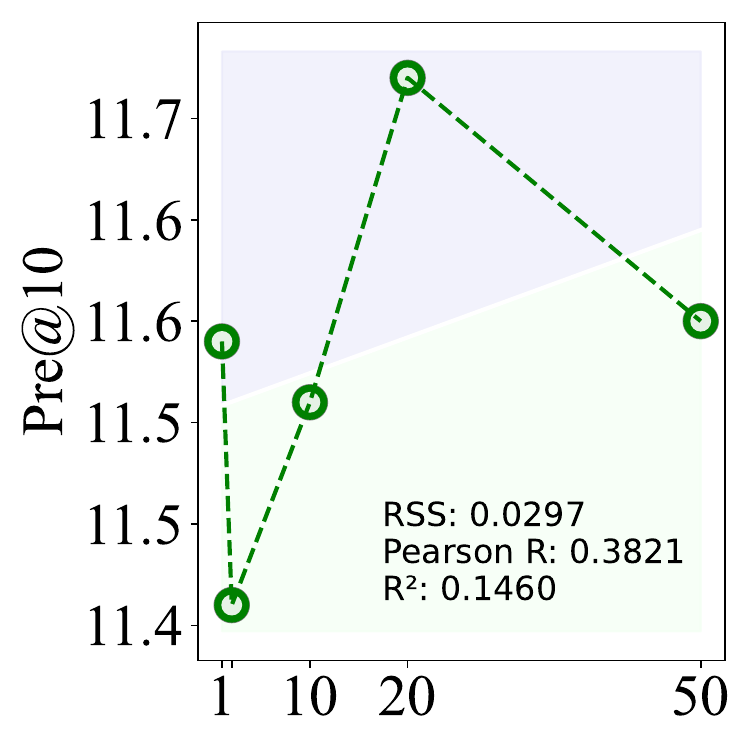}
            \end{minipage}%
  }%
  \vspace{-0.1in}
\caption{Performance comparison in validation dataset: Each enterprise provides training data to the client in a fixed proportion. Each enterprise provides training data, with the total data volume incrementally reaching 1K, 2K, 10K, 20K, and ultimately 50K. The model training process employs a batch size of 8 and 20 rounds. Each figure box is divided into two regions by its regression line. Each evaluation metric displays the Residual Sum of Squares (\texttt{RSS}), Pearson coefficient (\texttt{Pearson R}), and R-squared (\texttt{R}\(^{2}\)). These represent the total deviation of data points from the regression line, the strength of the linear relationship, and the proportion of variance explained by the model, respectively.}
\label{fig_otherproperties_data_increment}
\vspace{-0.2in}
\end{figure*}

\subsection{Main Results of Upstream RAG Retrieval}
\label{Main_Results_upstream}
The experimental results presented in Tables~\ref{main_retrieval_test} and ~\ref{main_retrieval_validation} highlight the outstanding performance of FedE4RAG across various metrics in the UPSTREAM RAG retrieval tasks. Overall, FedE4RAG demonstrates a comprehensive superiority in both test and validation settings. It consistently achieves the highest scores across a wide range of metrics,

In terms of Presence-Based Metrics, which focus on the presence of correct results in the top k outputs regardless of ranking, FedE4RAG achieves the highest scores in Hit@1 and Hit@10. This indicates that FedE4RAG is highly effective at retrieving relevant documents within the top ranks, which is crucial for user satisfaction and efficiency in information retrieval tasks. For Order-Sensitive Metrics, which evaluate the quality of ranked results with an emphasis on positional relevance, FedE4RAG stands out with scores of 55.39 in Mean Reciprocal Rank (MRR) and 60.34 in Normalized Discounted Cumulative Gain (NDCG)  in test. These metrics underscore FedE4RAG's ability to retrieve relevant documents and rank them effectively according to their relevance. Furthermore, FedE4RAG's superior performance in the DCG family metrics (DCG, NDCG, IDCG) underscores its efficacy in ranking models through the evaluation of ordered result quality. FedE4RAG attains a DCG score of 69.19, surpassing the centrally trained model SCenT by 5.1 percentage points and the non-federated baseline model SVanE1.5\(_{\rm base}\) by 7.97 percentage points. Additionally, FedE4RAG achieves a MAP score of 90.22, exceeding SCenT by 4.52 percentage points and SVanE1.5\(_{\rm base}\) by 8.26 percentage points, demonstrating its robust capability in prioritizing relevant documents. Regarding Threshold-Driven Metrics, which balance precision and recall using similarity thresholds, FedE4RAG shows a balanced performance in F1 scores, indicating a strong balance between precision and recall. Its high accuracy scores (Acc@k) across different settings demonstrate its reliability in correctly identifying relevant documents. Furthermore, FedE4RAG's strong performance in recall and precision metrics reflects its comprehensive coverage and quality of the retrieved results.

FedE4RAG's superior performance across all three categories of metrics—Presence-Based, Order-Sensitive, and Threshold-Driven—clearly positions it as the most effective model for the UPSTREAM RAG retrieval tasks. The comparison with ICMaxD and ICMinD (Strategy A) evaluates FedE4RAG's capacity to leverage IID client data effectively. FedE4RAG exceeds both in all metrics, showcasing its superior use of local data while maintaining high performance.

\begin{figure}[tb]
  \centering
  \begin{minipage}[t]{1\linewidth}
  \centering
  \includegraphics[width=\linewidth]{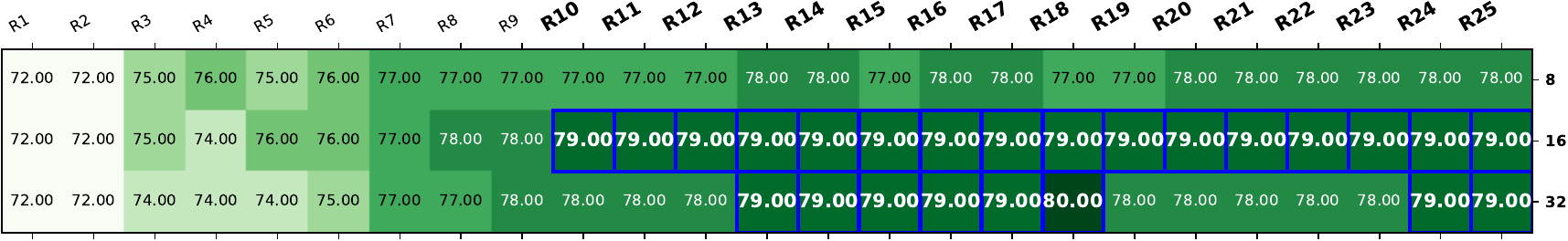}
  \end{minipage}%
  \\
  \begin{minipage}[t]{1\linewidth}
  \centering
  \includegraphics[width=\linewidth]{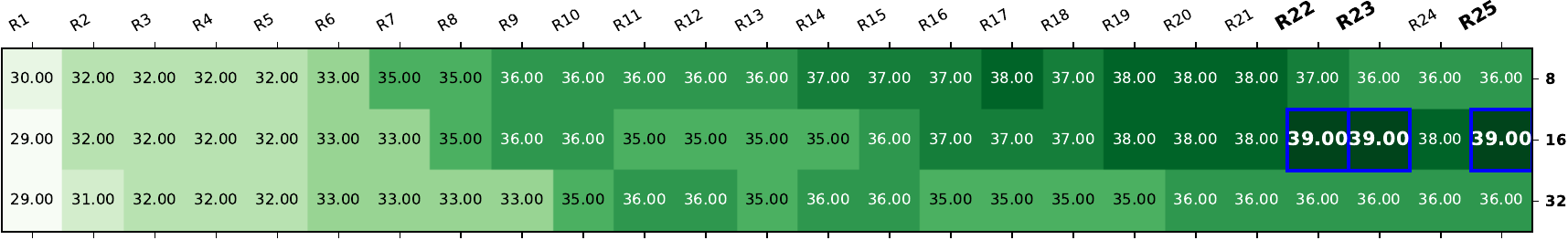}
  {\footnotesize Presence-Based Metrics: Hit@10, EM}
  \end{minipage}%
  \caption{Heatmap comparison of model performance with variations in number of training Rounds and Batch Sizes on Presence-Based Metrics.}
  \label{fig:hyperparameters}
\end{figure}

\subsection{Factors Discussions }
\subsubsection{Impact of RAG-FT} 
As shown in Tables~\ref{main_retrieval_test} and ~\ref{main_retrieval_validation}, our FedE4RAG consistently outperform the baseline backbone SVanE1.5\(_{\rm base}\) (\texttt{Strategy B}) in RAG retrieval testing, narrowing the performance gap between federated and centralized training. 
This demonstrates the effectiveness of federated learning in enhancing performance through  our upstream embedding learning from multiple clients. Evaluation against SVanE1.5\(_{\rm large}\) (\texttt{Strategy C}) assesses whether FedE4RAG can surpass the BGE-large model. Notably, our framework employs BGE-base, and the achieved advantage demonstrates that FedE4RAG with RAG-FT not only effectively utilizes local data but also outperforms the existing BGE-large model in terms of performance.

\subsubsection{Impact of FED-HE}  Our results indicate that the integration of FED-HE into the FedE4RAG framework has a negligible impact on the overall performance of the model. As shown in Table 2, the performance of the FedE4RAG model with FED-HE enabled is comparable to the model without encryption, demonstrating that the encryption process does not significantly hinder the model's ability to learn and generalize from the data. Moreover, the use of FED-HE provides an additional layer of security, ensuring that the data remains confidential throughout the training process. This is particularly important in scenarios where data privacy is a critical concern, such as in the financial domains. In terms of computational efficiency, we observed a slight increase in training time when using FED-HE due to the overhead of homomorphic encryption operations. However, this increase was found to be acceptable, given the prominent benefits in terms of data privacy and security. 
In conclusion, the integration of FED-HE into the FedE4RAG framework provides a robust solution for privacy-preserving federated learning in RAG systems. It allows for the collaborative training of models while maintaining the confidentiality of sensitive data, making it a valuable addition to the framework, especially for applications in domains with strict privacy requirements.

\begin{figure}[tb]
  \centering
  \begin{minipage}[t]{1\linewidth}
  \centering
  \includegraphics[width=\linewidth]{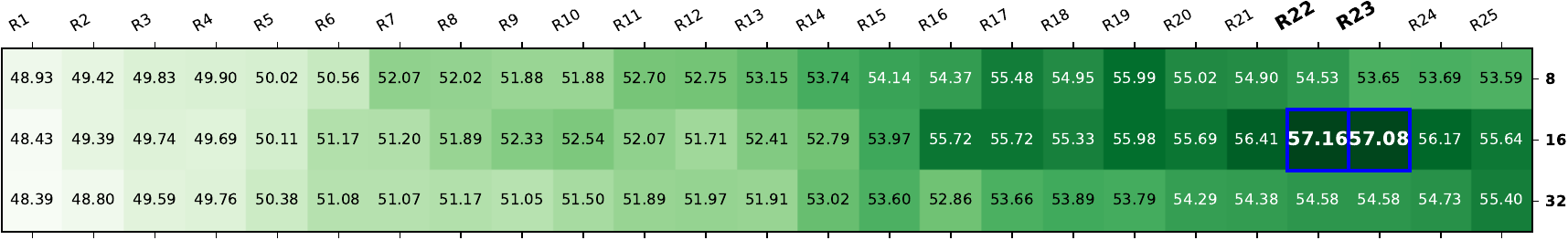}
  \end{minipage}%
  \\
  \begin{minipage}[t]{1\linewidth}
  \centering
  \includegraphics[width=\linewidth]{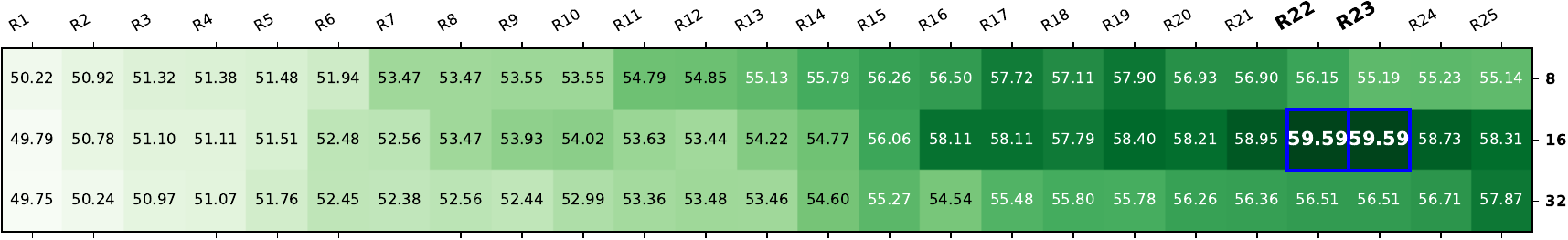}
  \end{minipage}%
  \\
  \begin{minipage}[t]{1\linewidth}
  \centering
  \includegraphics[width=\linewidth]{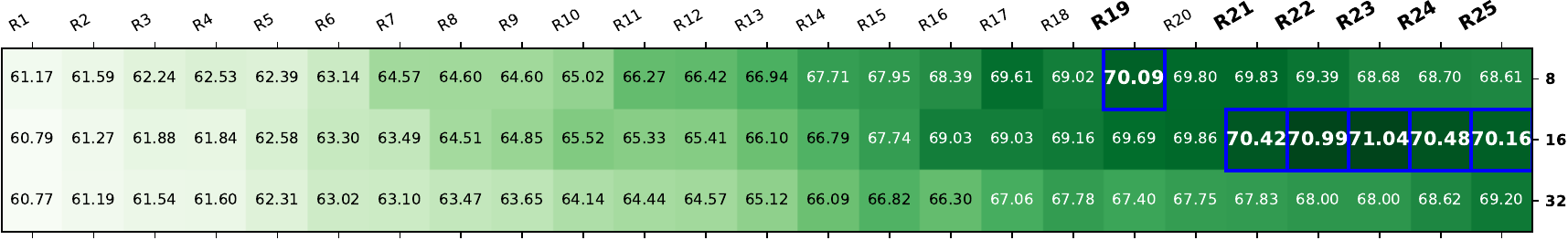}
  \end{minipage}%
  \\
  \begin{minipage}[t]{1\linewidth}
  \centering
  \includegraphics[width=\linewidth]{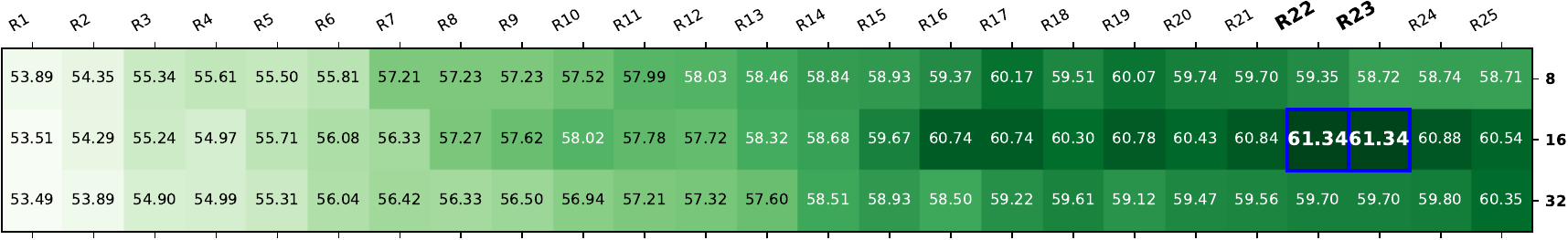}
  \end{minipage}%
  \\
  \begin{minipage}[t]{1\linewidth}
  \centering
  \includegraphics[width=\linewidth]{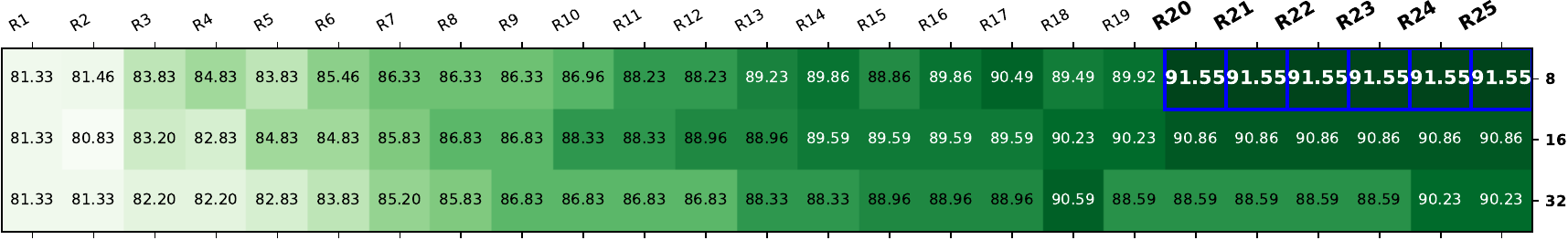}
  {\footnotesize Order-Sensitive Metrics: MRR, MAP, DCG, NDCG, IDCG}
  \end{minipage}%
  \caption{Heatmap comparison of model performance with variations in number of training Rounds and Batch Sizes on Order-Sensitive Metrics. }
  \label{hyperparameters}
\end{figure}

\begin{figure}[tb]
  \centering
  \begin{minipage}[t]{1\linewidth}
  \centering
  \includegraphics[width=3.45in]{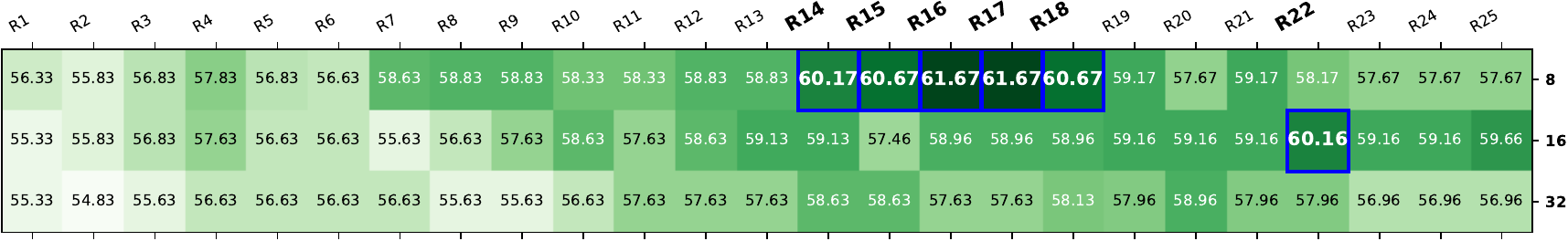}
  \end{minipage}%
  \\
  \begin{minipage}[t]{1\linewidth}
  \centering
  \includegraphics[width=3.45in]{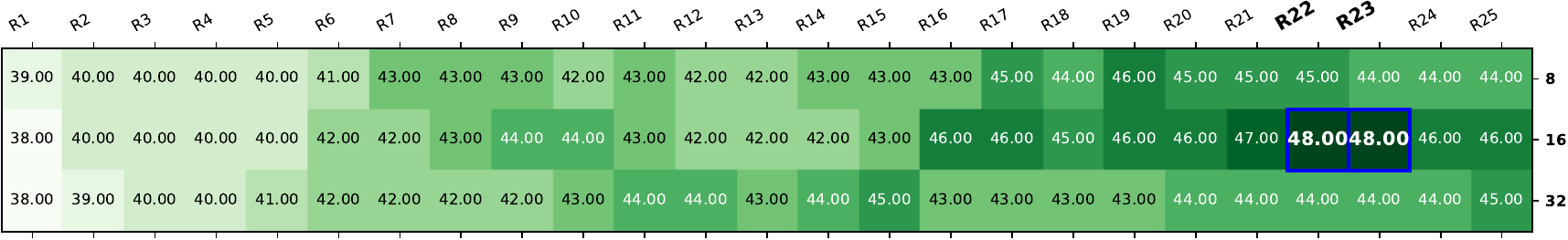}
  \end{minipage}%
  \\
  \begin{minipage}[t]{1\linewidth}
  \centering
  \includegraphics[width=3.45in]{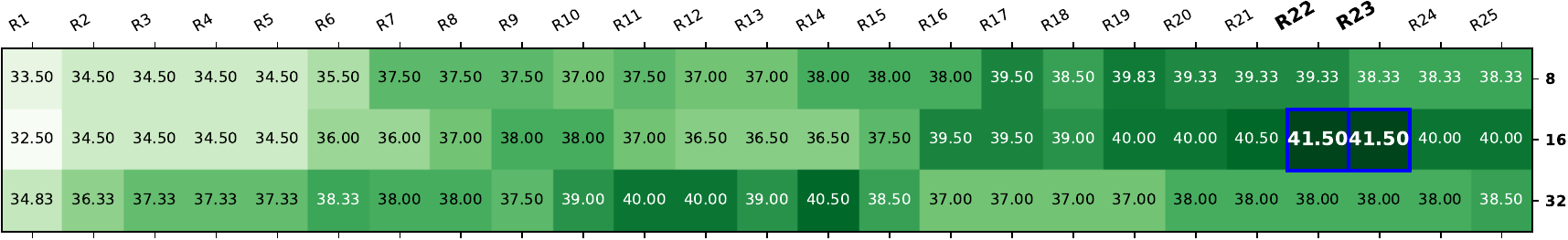}
  \end{minipage}%
  \\
  \begin{minipage}[t]{1\linewidth}
  \centering
  \includegraphics[width=3.45in]{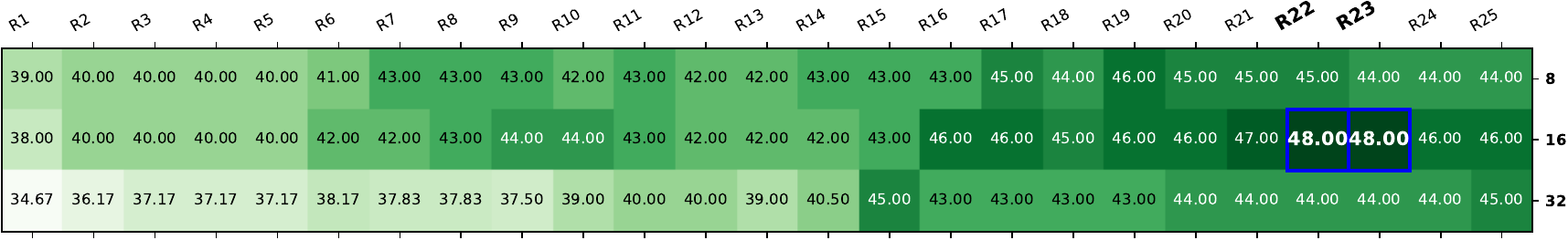}
  {\footnotesize Threshold-Driven Metrics: F1, Accuracy@1, Recall@1, Precision@1}
  \end{minipage}%
  \caption{Heatmap comparison of model performance with variations in number of training Rounds and Batch Sizes on Threshold-Driven Metrics.}
  \label{hyperparameters}
\end{figure}

\subsubsection{Impact of KD-GLE} 
As shown in Tables~\ref{main_retrieval_test} and ~\ref{main_retrieval_validation},  the non-distilled model (SCent \& FedAvg) performs worse than our FedE4RAG. 
Comparison with SCenT (\texttt{Strategy D}) reveals that FedE4RAG achieves superior performance to centrally trained models without compromising data privacy. This is of great importance for both data privacy protection and model performance enhancement. 
Assessment against FedAvg (\texttt{Strategy E}) evaluates the performance advantage of FedE4RAG's federated learning approach, which combines knowledge  distillation and homomorphic encryption, over the standard FedAvg method. This underscores the remarkable effectiveness of our FedE4RAG, which demonstrates substantial performance gains in both secure communication and performance optimization.

\subsection{Other Upstream Analysis}
\subsubsection{Impact of Federated Training Data Scale}  
In a federated learning of our FedE4RAG, as the amount of data provided by clients (each companies) increases from 1K to 50K, most performance metrics exhibit an upward trend as shown in Figure~\ref{fig_otherproperties_data_increment}. Specifically, metrics such as MRR@10, EM, DCG@10, NDCG@10, and MAP improve with increasing data volume, indicating that more data enhances the model's retrieval performance.
Moreover, the high correlation coefficients depicted in Figures~\ref{fig:side:MRR}, \ref{fig:side:DCG}, \ref{fig:side:NDCG}, \ref{fig:side:rec5}, and \ref{fig:side:pre5}—with Pearson R values of 0.9696 for MRR, 0.9350 for DCG@10, 0.9398 for NDCG@10, 0.9569 for Rec@5, and 0.9388 for Pre@5—indicate a robust association between these metrics and data volume. An increase in data volume consistently enhances the performance indicated by these metrics.

\begin{table*}[h]
  \centering
  \scriptsize
  \caption{Comparative  \colorbox{blue!20}{\textcolor{blue}{\textbf{test}}}  performance of RAG systems integrated with our FedE4RAG but using different LLMs on the \textcolor{blue}{\textbf{\cellcolor{gray!20}downstream RAG generation}} tasks. The \textbf{best} results for each metric are highlighted in bold.}
  \renewcommand\arraystretch{1.6}
  \setlength{\tabcolsep}{1.2mm}
  \begin{tabular}{|l|c|c|c|c|c|c|c|c|c|c|c|c|c|c|c|}
    \specialrule{1pt}{0pt}{0pt}
  \textbf{LLM} & \multicolumn{12}{c|}{\textbf{N-gram overlap}} & \multicolumn{1}{c|}{\textbf{Intrinsic}} & \multicolumn{2}{c|}{\textbf{Edit Distance}}  \\ 
  \hline
  \textcolor{red}{+} \textbf{FedE4RAG} & \textbf{Chrf} & \textbf{Chrf\(^{++}\)} & \textbf{METEOR} & \textbf{R-1 Pre} & \textbf{R-2 Pre} & \textbf{R-L Pre} & \textbf{R-1 Rec} & \textbf{R-2 Rec} & \textbf{R-L Rec} & \textbf{R-1 F1} & \textbf{R-2 F1} & \textbf{R-L F1} & \textbf{PPL\(_{\downarrow}\)} & \textbf{CER\(_{\downarrow}\)} & \textbf{WER\(_{\downarrow}\)}  \\ \hline
  Llama3.1-8B & 17.50 & 16.10 & 16.05 & 11.45 & 05.24 & 08.51 & 51.93 & 18.46 & 42.52 & 15.88 & 07.16 & 11.93 & 45.74 & 36.18 & 28.85  \\ 
  GPT-4o Mini & \textbf{24.40} & \textbf{22.81} & \textbf{23.33} & \textbf{24.06} & \textbf{14.63} & \textbf{21.15} & \textbf{54.59} & \textbf{28.34} & \textbf{49.79} & \textbf{25.85} & \textbf{15.62} & \textbf{22.71} & 59.96 & 21.13 & 16.19   \\ \hline
  MathStral-7B & 18.15 & 16.77 & 15.11 & 19.97 & 07.28 & 17.31 & 49.28 & 19.73 & 41.05 & 22.38 & 07.97 & 18.85 & 100.95 & \textbf{15.63} & \textbf{12.79}  \\ 
  DeepSeek R1-7B & 18.12 & 16.37 & 17.63 & 09.72 & 04.23 & 07.22 & 51.18 & 20.13 & 41.37 & 14.84 & 06.51 & 11.16 & \textbf{36.19} & 31.09 & 27.66  \\ 
  \specialrule{1pt}{0pt}{0pt}
  \end{tabular}
  \label{LLM_TEST}
\end{table*}

\begin{table*}[h]
  \centering
  \scriptsize
  \caption{Comparative \colorbox{gray!20}{\textcolor{blue}{\textbf{validation}}} performance of RAG systems integrated with our FedE4RAG but using different LLMs on the \textcolor{blue}{\textbf{\cellcolor{gray!20}downstream RAG generation}} tasks. The \textbf{best} results for each metric are highlighted in bold.}
  \renewcommand\arraystretch{1.6}
  \setlength{\tabcolsep}{1.2mm}
  \begin{tabular}{|l|c|c|c|c|c|c|c|c|c|c|c|c|c|c|c|}
    \specialrule{1pt}{0pt}{0pt}
  \textbf{LLM} & \multicolumn{12}{c|}{\textbf{N-gram overlap}} & \multicolumn{1}{c|}{\textbf{Intrinsic}} & \multicolumn{2}{c|}{\textbf{Edit Distance}}  \\ 
  \hline
  \textcolor{red}{+} \textbf{FedE4RAG} & \textbf{Chrf} & \textbf{Chrf\(^{++}\)} & \textbf{METEOR} & \textbf{R-1 Pre} & \textbf{R-2 Pre} & \textbf{R-L Pre} & \textbf{R-1 Rec} & \textbf{R-2 Rec} & \textbf{R-L Rec} & \textbf{R-1 F1} & \textbf{R-2 F1} & \textbf{R-L F1} &\textbf{PPL\(_{\downarrow}\)} & \textbf{CER\(_{\downarrow}\)} & \textbf{WER\(_{\downarrow}\)}   \\ \hline
  Llama3.1-8B & 21.21 & 19.50 & \textbf{20.43} & 13.24 & 06.24 & 10.22 & \textbf{54.83} & \textbf{21.91} & \textbf{46.13} & 19.14 & 08.92 & 14.95 & 49.16 & 26.41 & 18.75 \\ 
  GPT-4o Mini & \textbf{22.57} & \textbf{20.95} & 17.57 & \textbf{33.87} & \textbf{13.85} & \textbf{31.76} & 39.27 & 16.55 & 36.77 & \textbf{24.93} & \textbf{11.36} & \textbf{22.74} & 123.03 & \textbf{12.77} & \textbf{09.52} \\ \hline
  MathStral-7B & 18.77 & 16.72 & 14.51 & 16.90 & 06.07 & 14.31 & 47.92 & 18.94 & 41.24 & 19.62 & 07.14 & 16.32 & 111.96 & 19.21 & 11.06 \\ 
  DeepSeek R1-7B & 15.95 & 14.41 & 16.23 & 08.23 & 03.32 & 06.14 & 51.29 & 21.43 & 43.82 & 12.87 & 05.33 & 09.81 & \textbf{38.41} & 47.72 & 31.68 \\ \hline
  \end{tabular}
  \label{LLM_validation}
\end{table*}

\begin{figure*} 
  \begin{minipage}[t]{0.14\linewidth} 
  \subfigure[{\scriptsize Llama3.1-8B:DR-Q}]{
    \includegraphics[width=1.in]{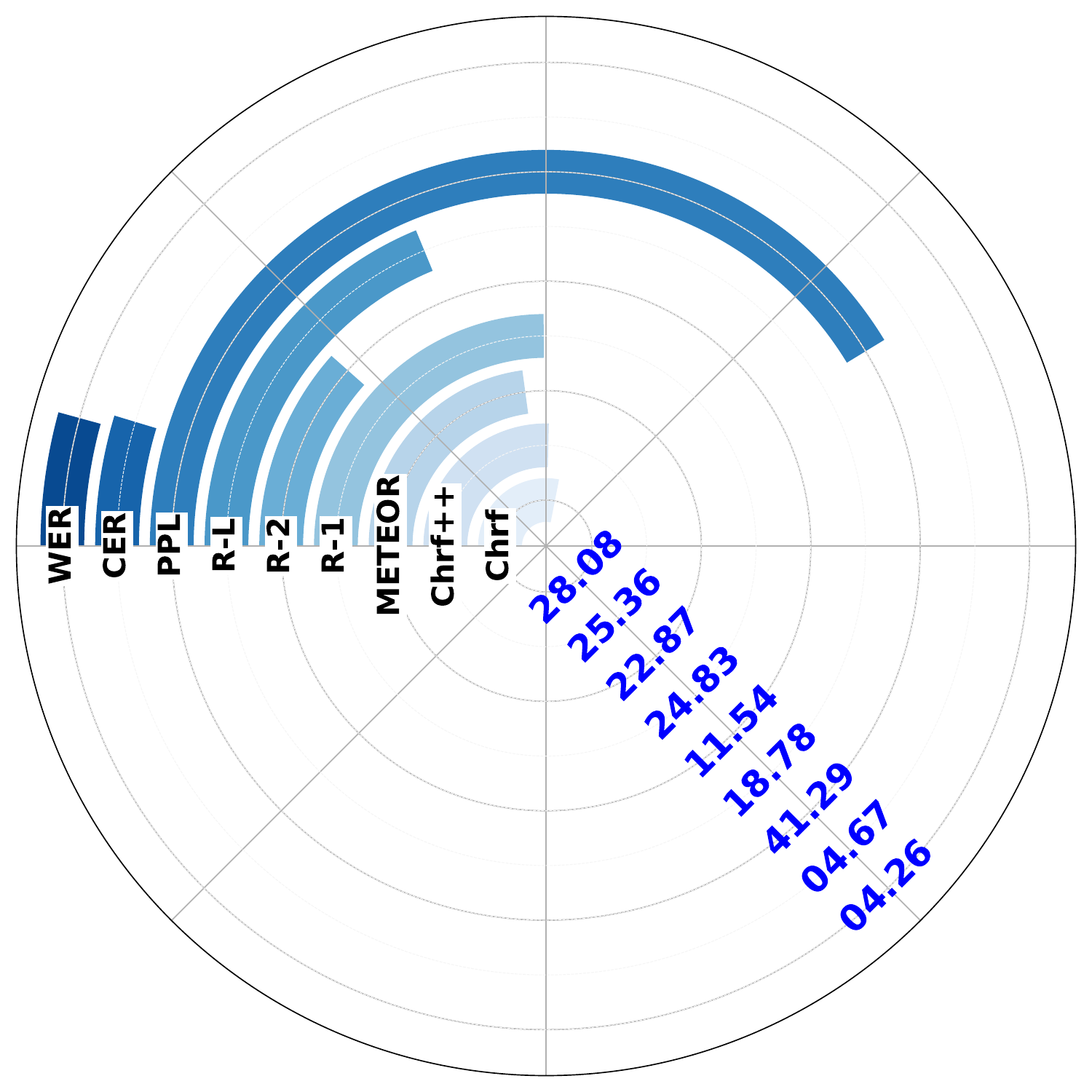}}
  \end{minipage}\hspace{0.025\linewidth}%
  \begin{minipage}[t]{0.14\linewidth} 
    \subfigure[{\scriptsize Llama3.1-8B:MB-Q}]{
      \includegraphics[width=1.in]{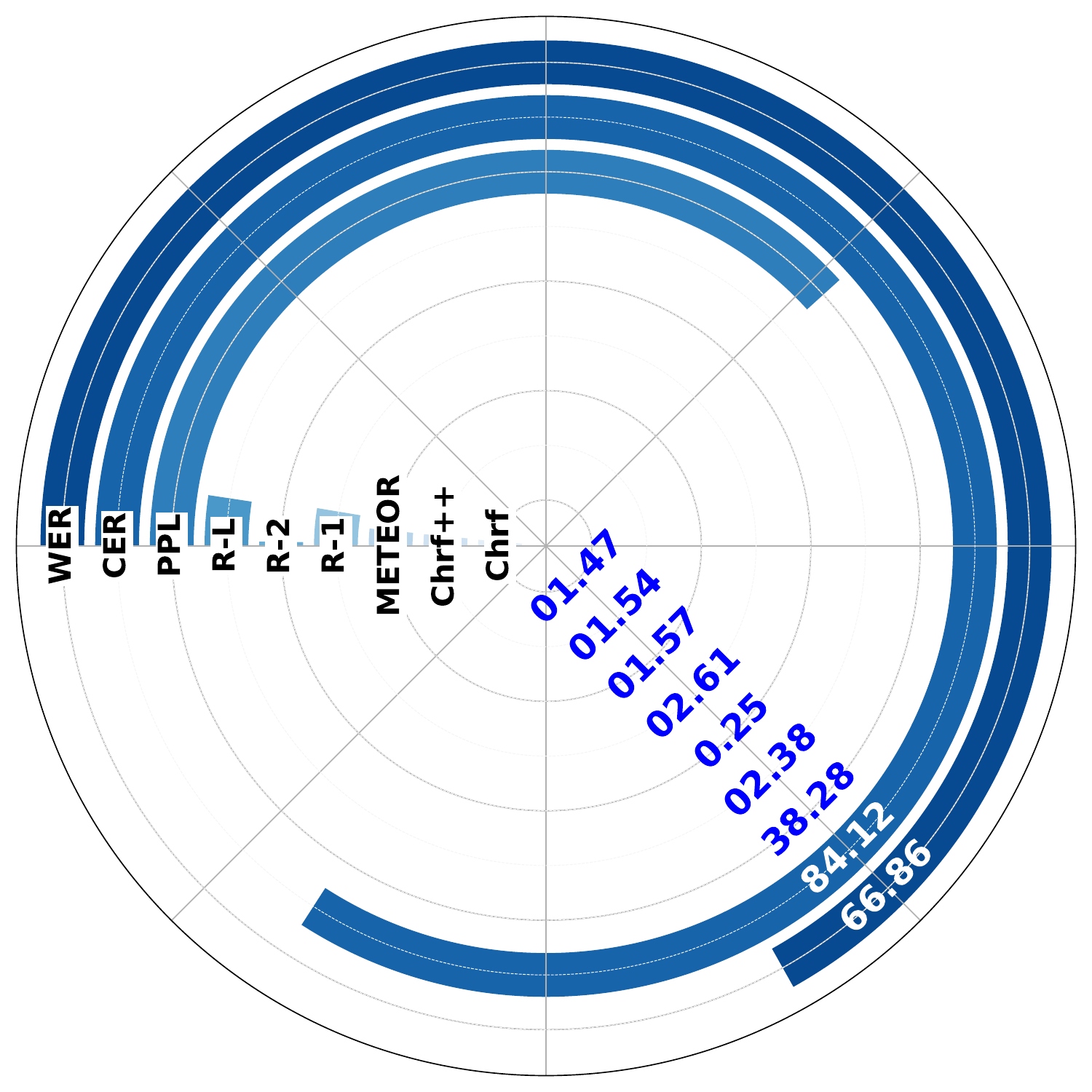}}
    \end{minipage}\hspace{0.025\linewidth}%
    \begin{minipage}[t]{0.14\linewidth} 
      \subfigure[{\scriptsize Llama3.1-8B:NB-Q}]{
        \includegraphics[width=1.in]{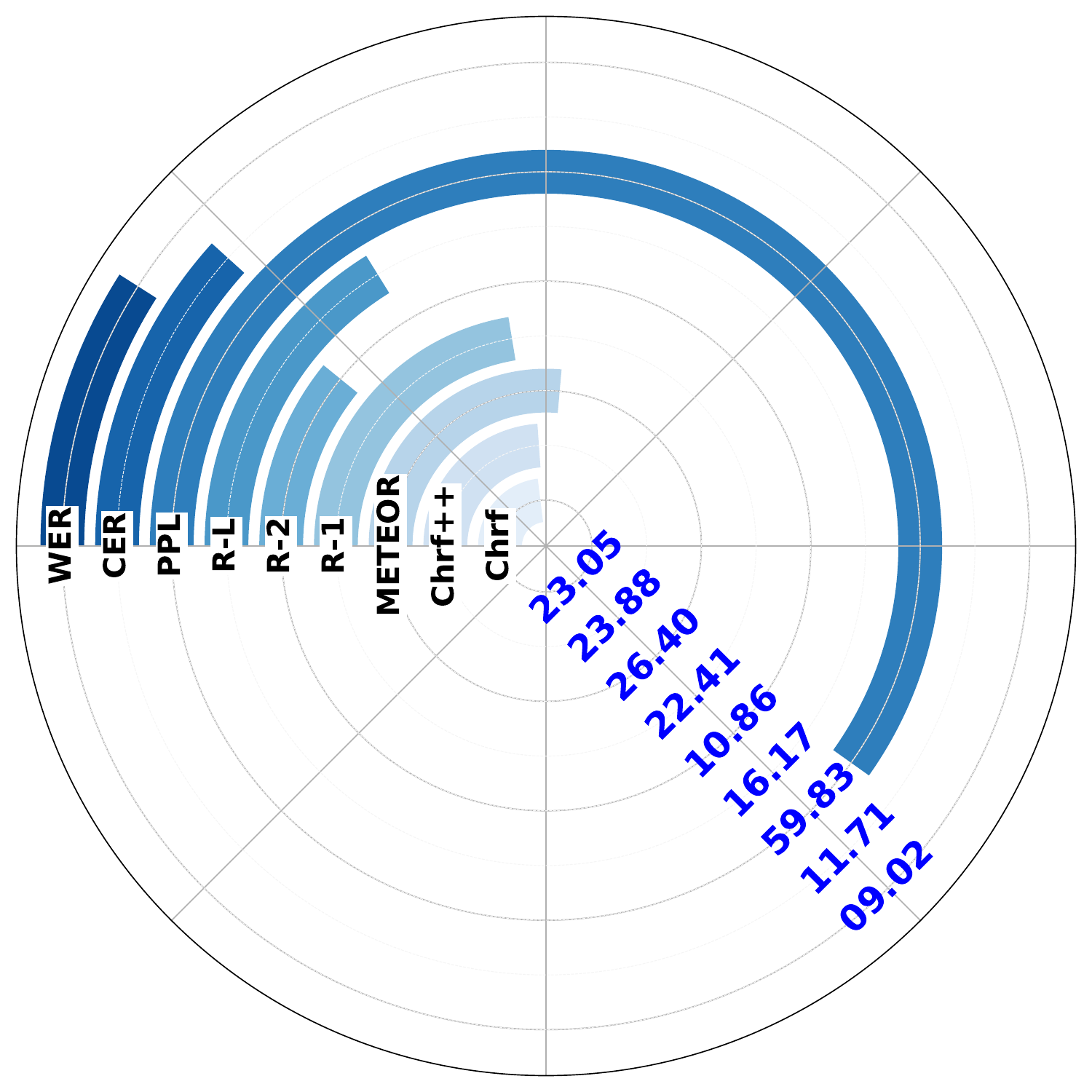}}
      \end{minipage}\hspace{0.025\linewidth}%
      \begin{minipage}[t]{0.14\linewidth} 
        \subfigure[{\scriptsize GPT-4o Mini:DR-Q}]{
          \includegraphics[width=1.in]{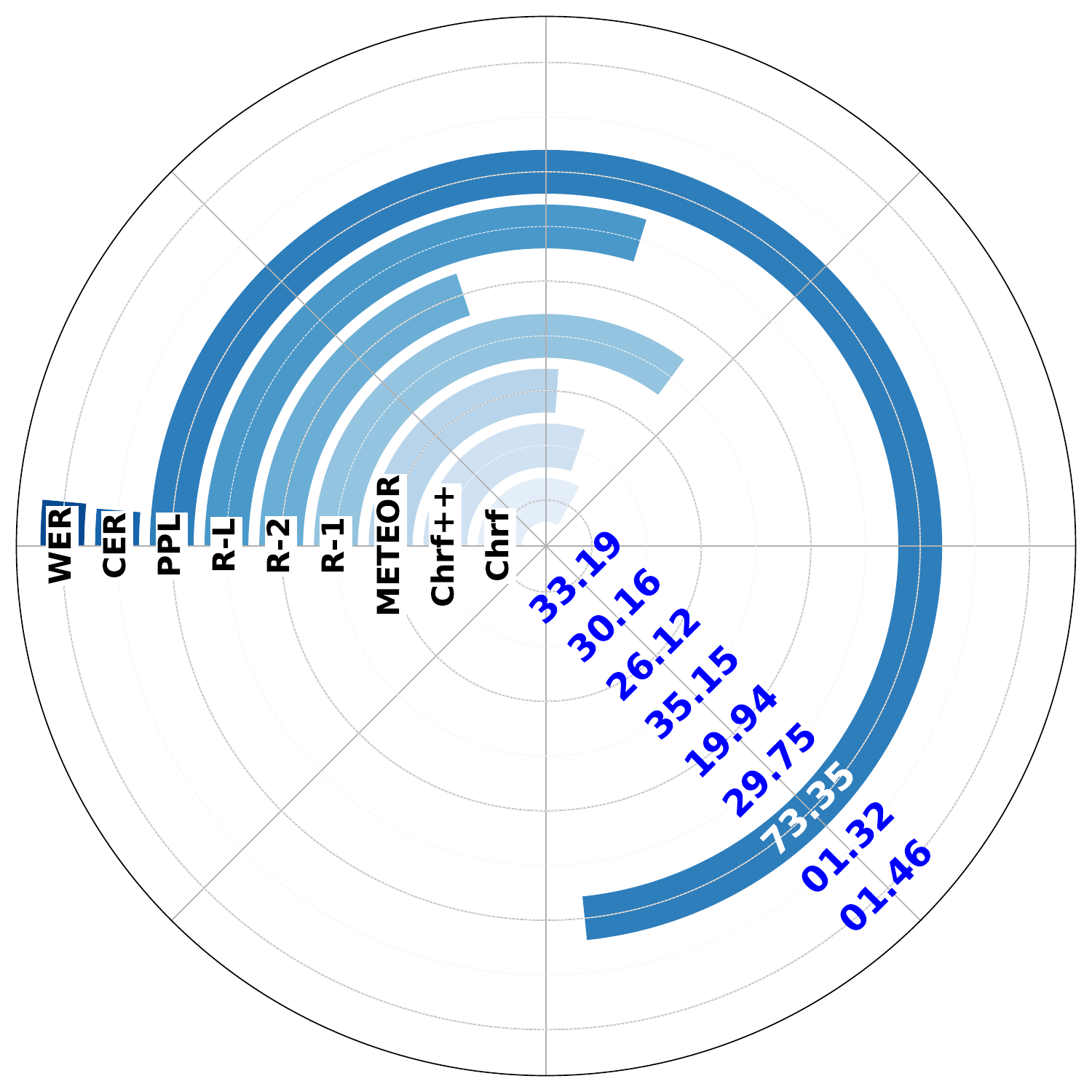}}
        \end{minipage}\hspace{0.025\linewidth}%
        \begin{minipage}[t]{0.14\linewidth} 
          \subfigure[{\scriptsize GPT-4o Mini:MB-Q}]{
            \includegraphics[width=1.in]{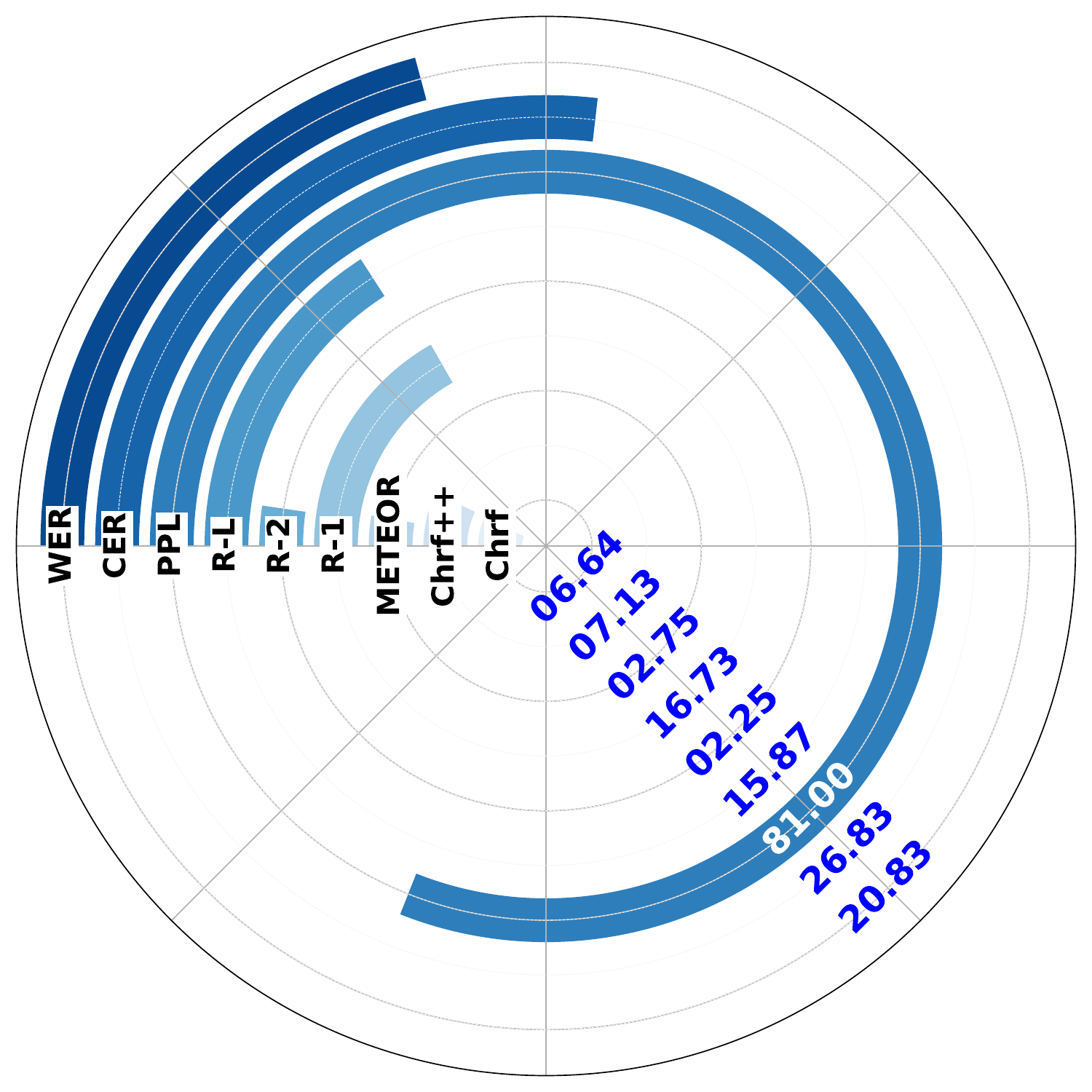}}
          \end{minipage}\hspace{0.025\linewidth}%
 \begin{minipage}[t]{0.14\linewidth} 
        \subfigure[{\scriptsize GPT-4o Mini:NB-Q}]{
          \includegraphics[width=1.in]{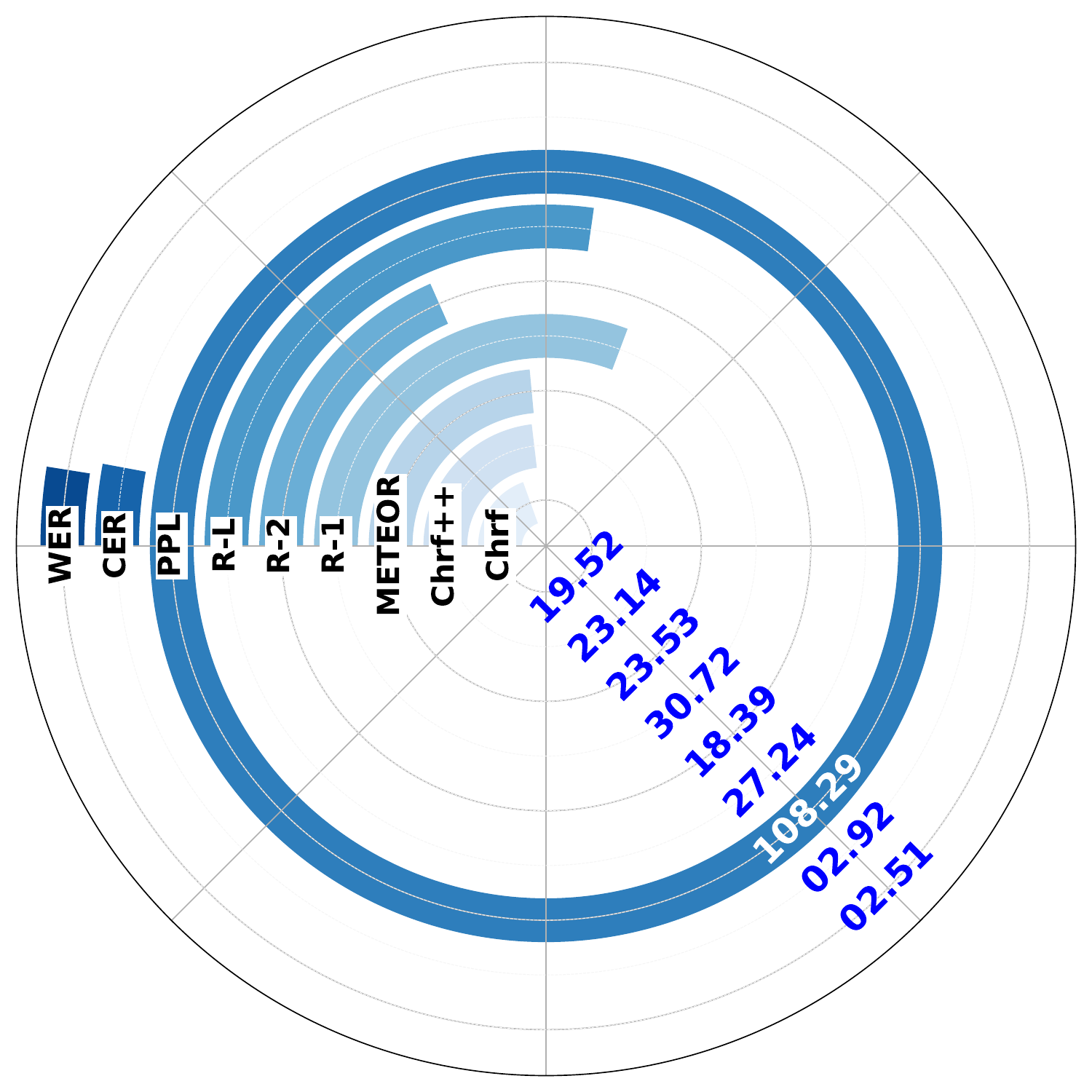}}
        \end{minipage}\hspace{0.025\linewidth}%

\begin{minipage}[t]{0.14\linewidth} 
          \subfigure[{\scriptsize MathStral-7B:DR-Q}]{
            \includegraphics[width=1.in]{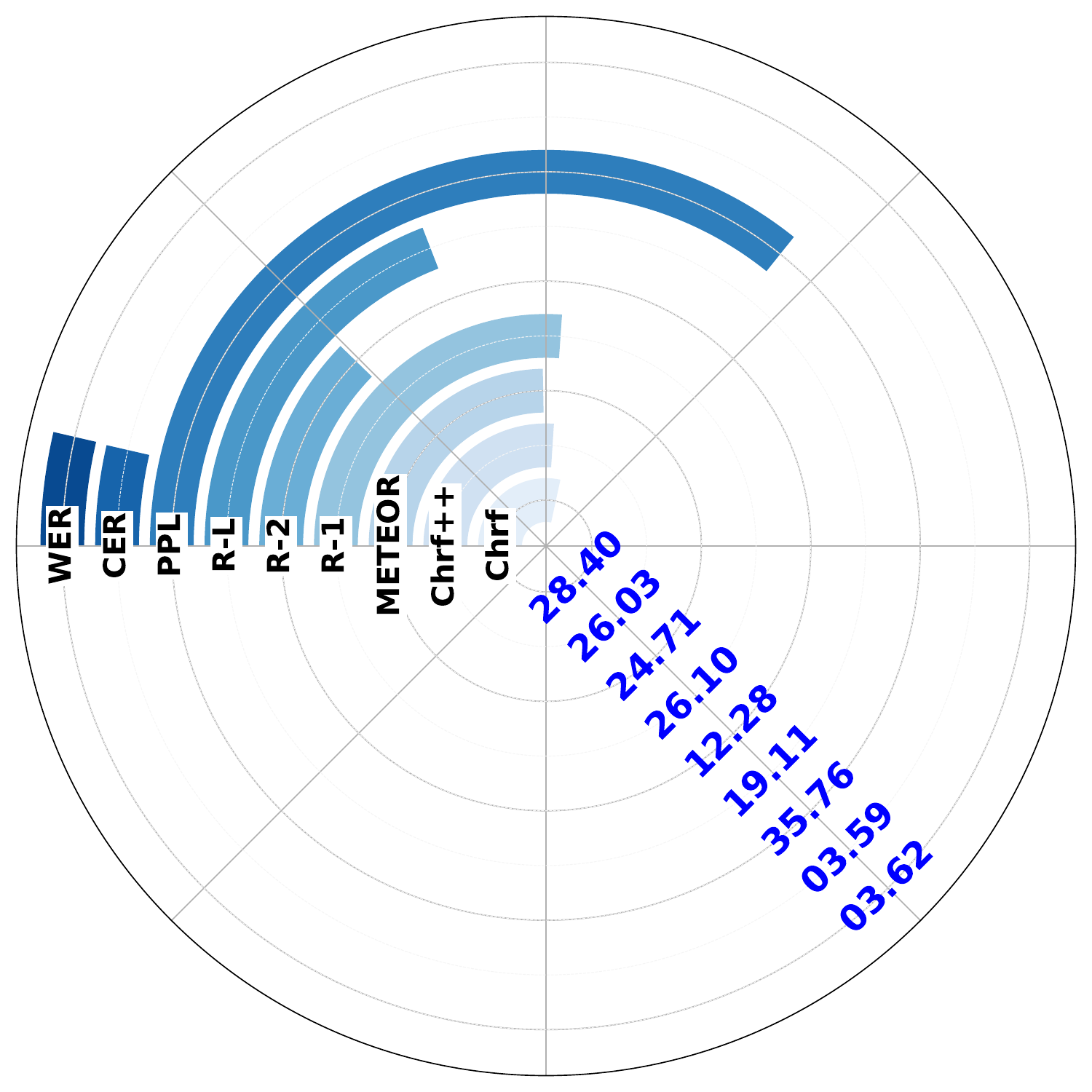}}
\end{minipage}\hspace{0.025\linewidth}%
\begin{minipage}[t]{0.14\linewidth} 
\subfigure[{\scriptsize MathStral-7B:MB-Q}]{
              \includegraphics[width=1.in]{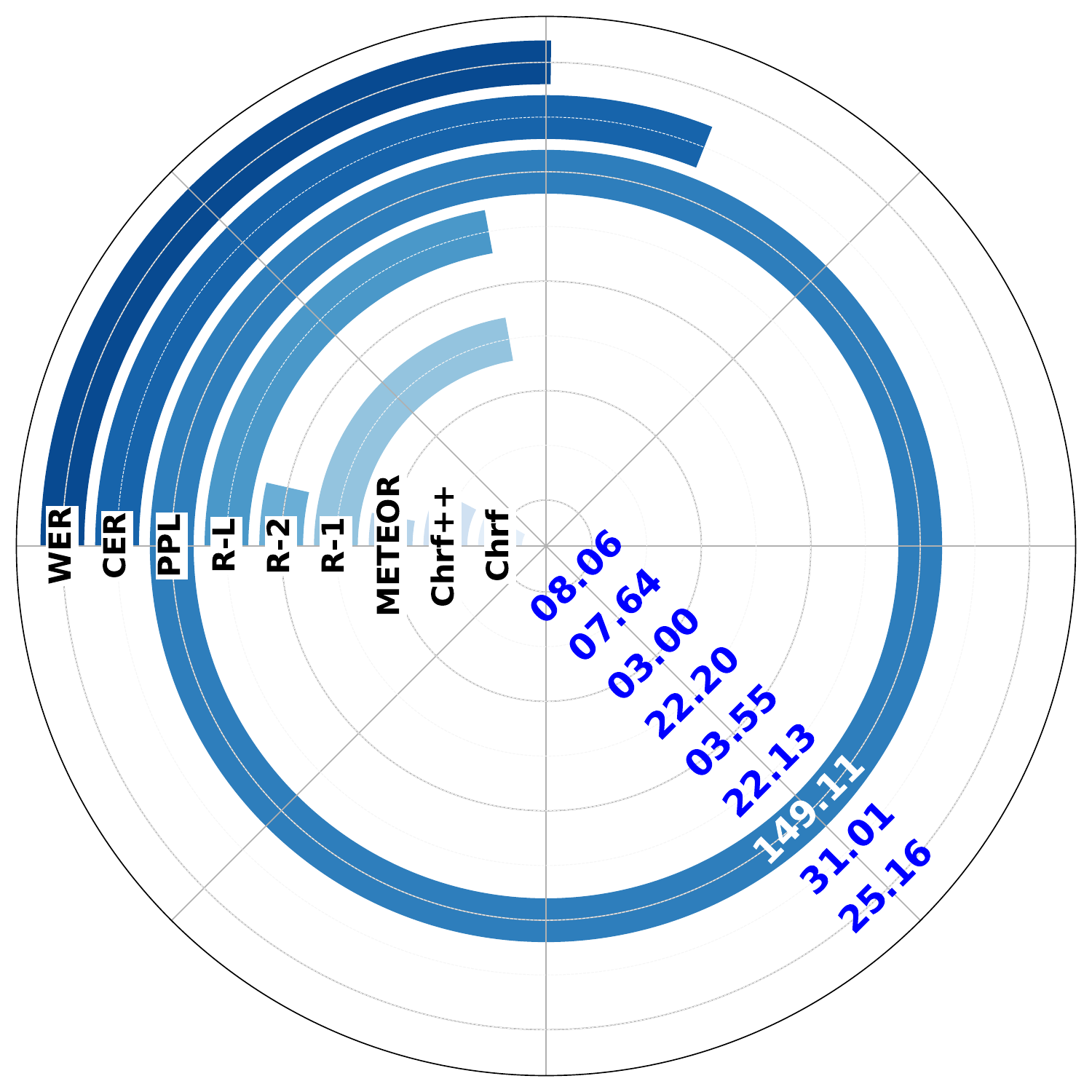}}
            \end{minipage}\hspace{0.025\linewidth}%
            \begin{minipage}[t]{0.14\linewidth} 
              \subfigure[{\scriptsize MathStral-7B:NB-Q}]{
                \includegraphics[width=1.in]{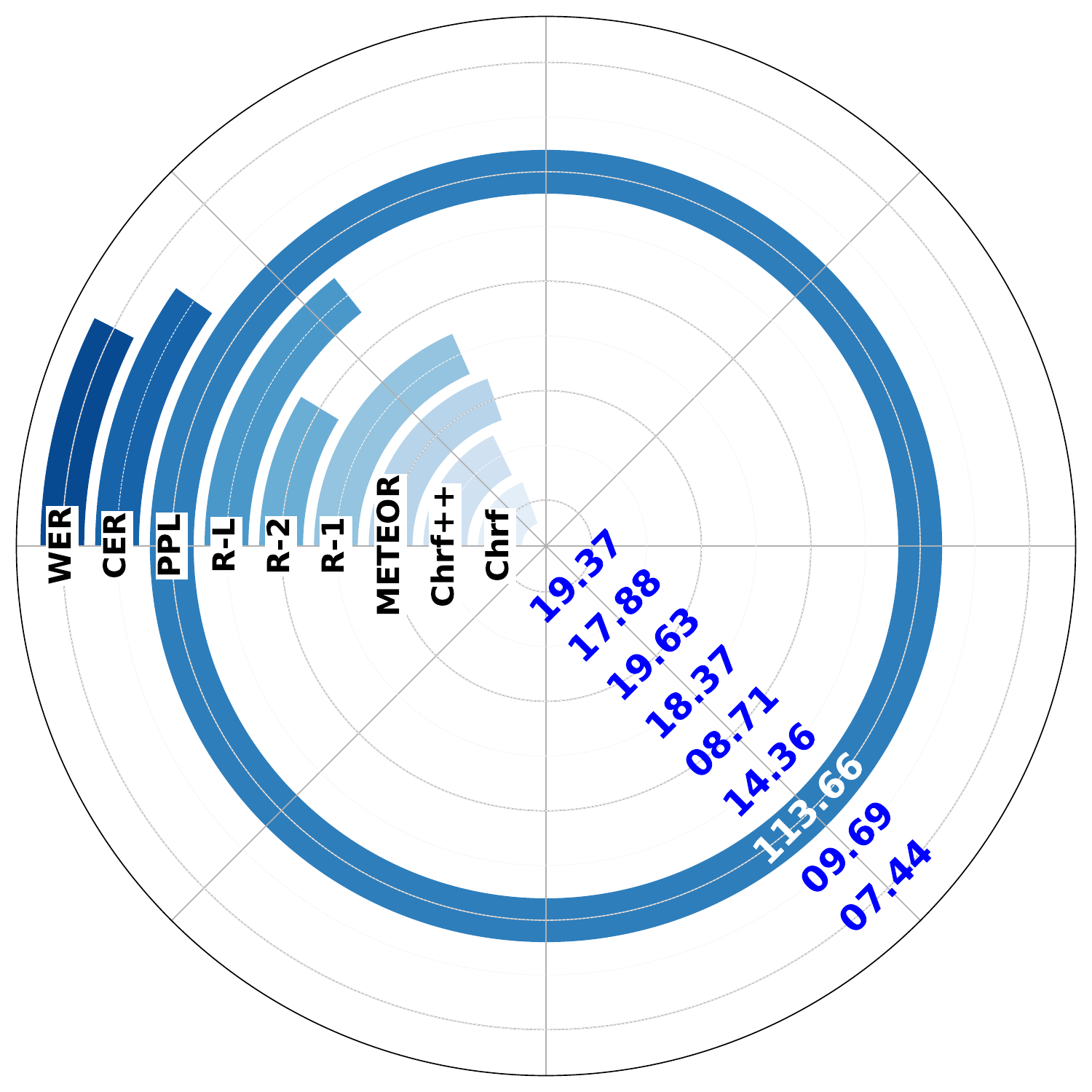}}
              \end{minipage}\hspace{0.025\linewidth}%
              \begin{minipage}[t]{0.14\linewidth} 
                \subfigure[{\scriptsize DS R1-7B:DR-Q}]{
                  \includegraphics[width=1.in]{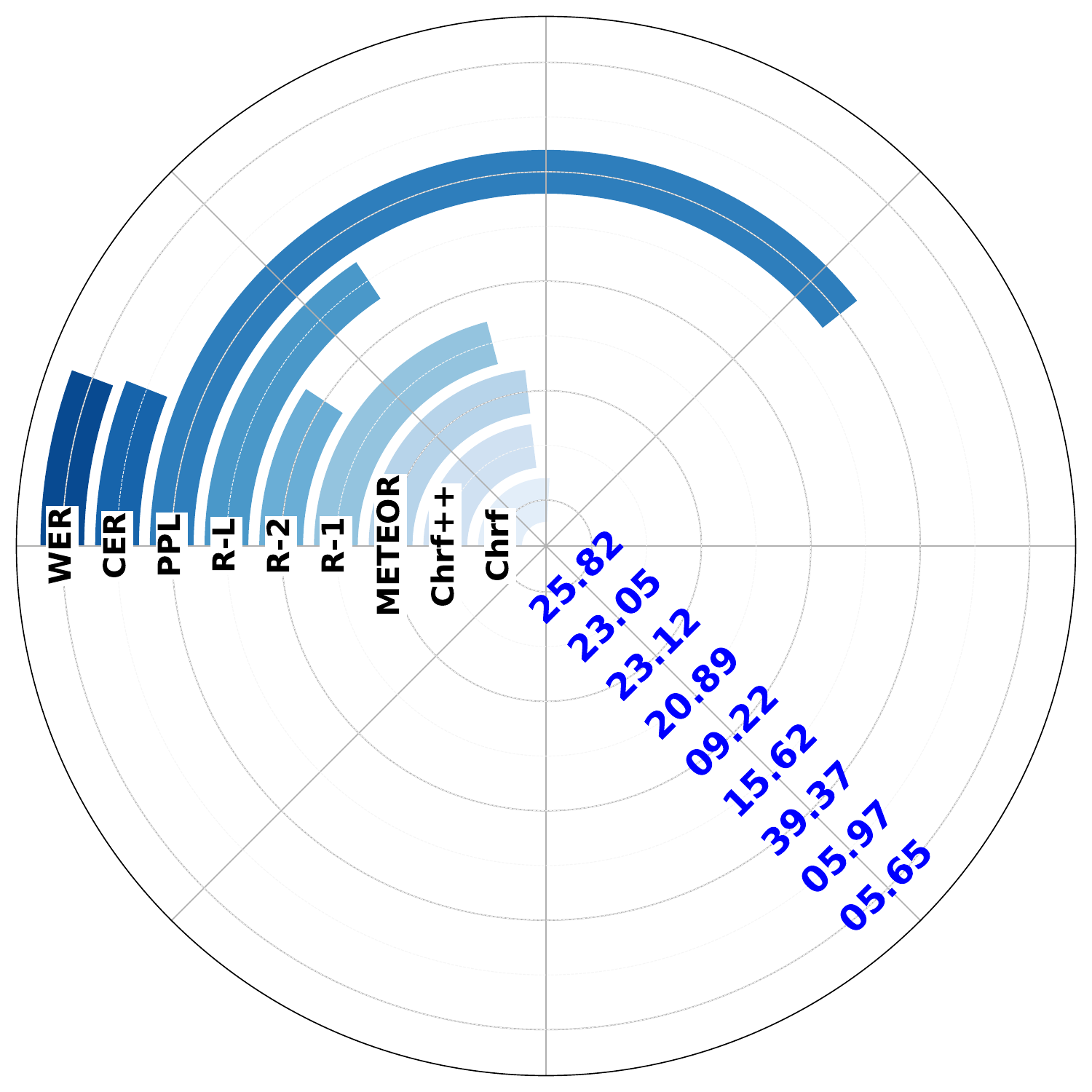}}
                \end{minipage}\hspace{0.025\linewidth}%
                \begin{minipage}[t]{0.14\linewidth} 
                  \subfigure[{\scriptsize DS R1-7B:MB-Q}]{
                    \includegraphics[width=1.in]{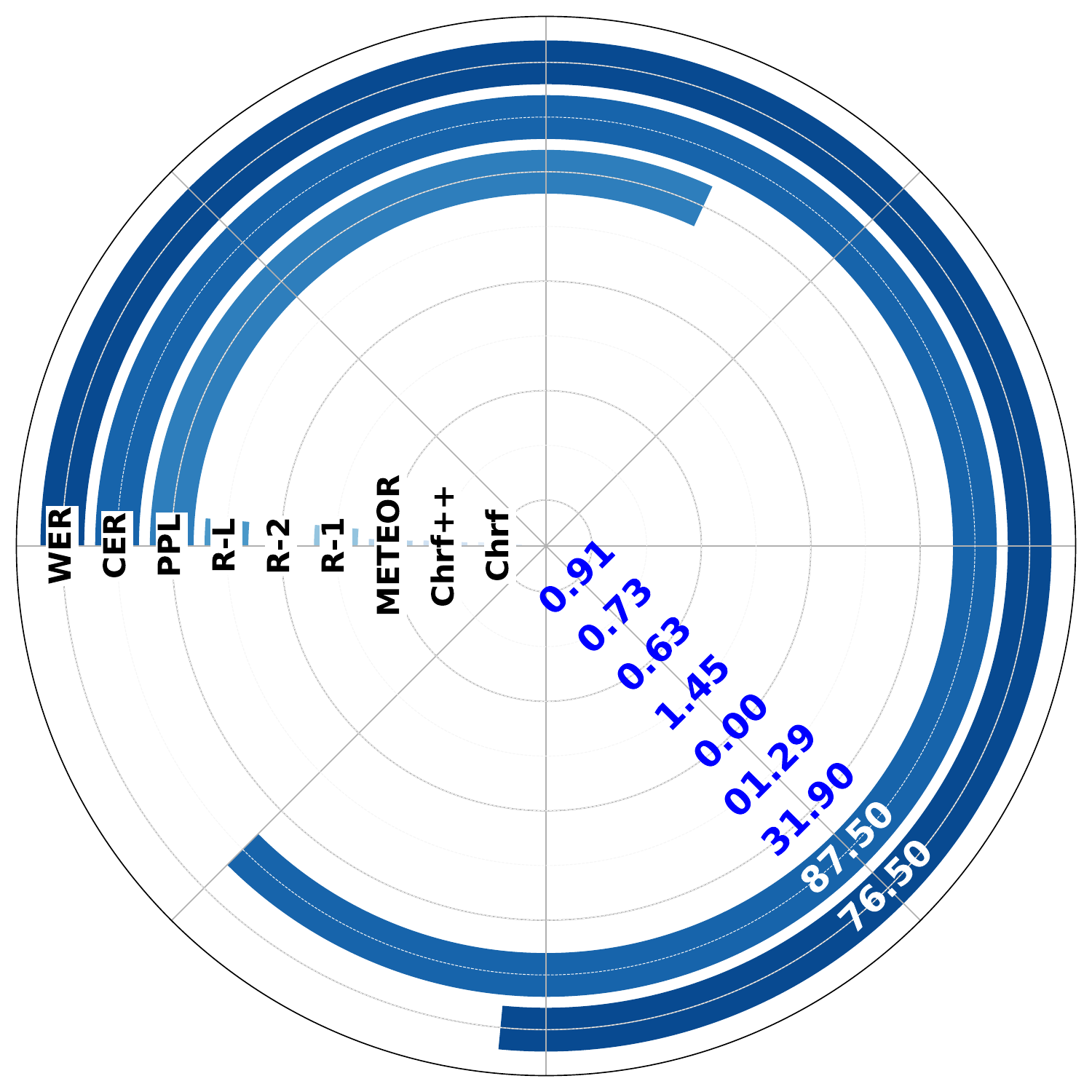}}
                  \end{minipage}\hspace{0.025\linewidth}%
         \begin{minipage}[t]{0.14\linewidth} 
                \subfigure[{\scriptsize DS R1-7B:NB-Q}]{
                  \includegraphics[width=1.in]{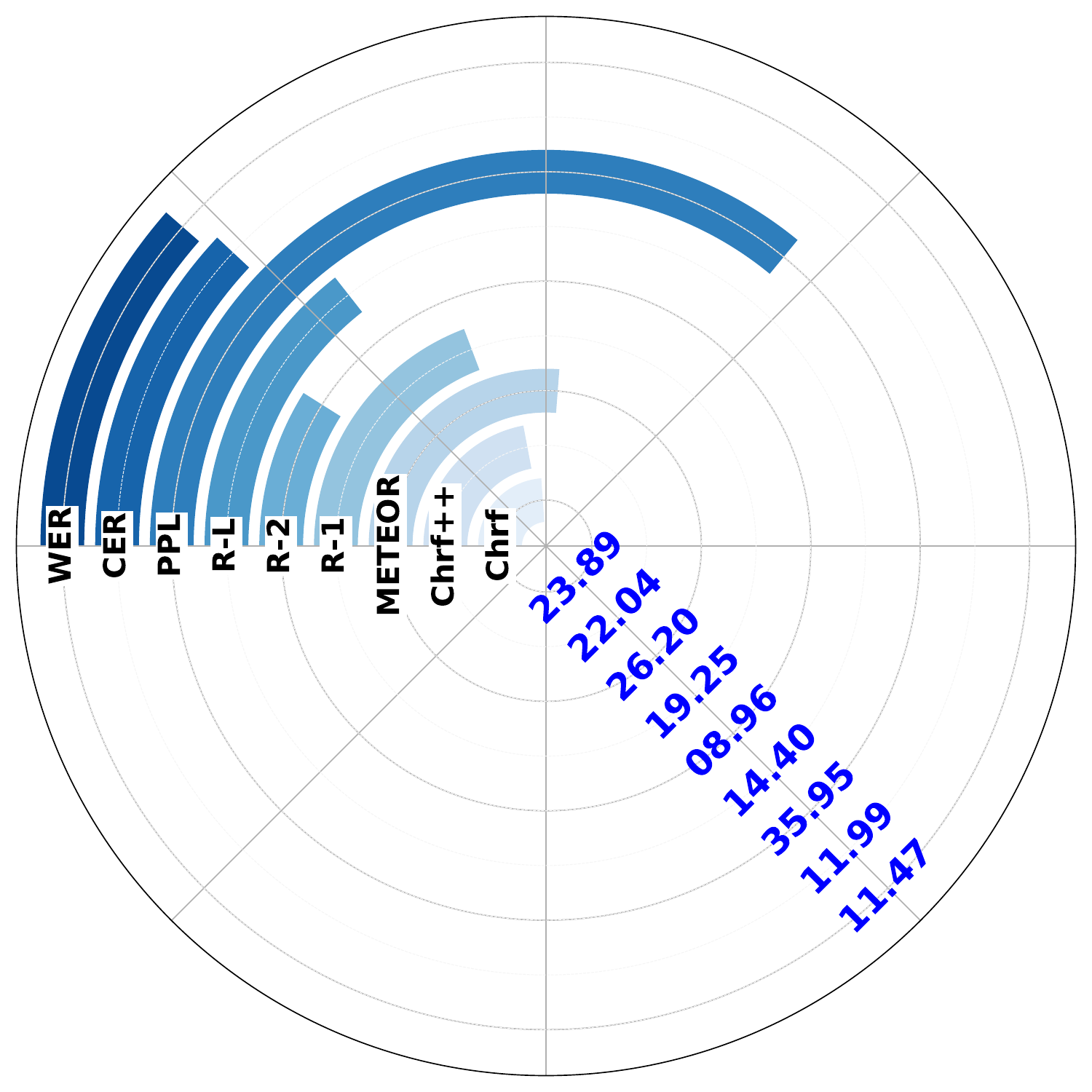}}
                \end{minipage}\hspace{0.025\linewidth}%
    \caption{Circular bar chart illustrations of question-answering performance on the test set, delineating various question reasoning categories. The chart categorizes question reasoning into information extraction queries (InfoEX) and complex reasoning queries, which comprise numerical reasoning (NumRS) and logical reasoning (LogcRS). The values of Perplexity (PPL), Character Error Rate (CER), and Word Error Rate (WER) are optimal when minimized.
    }
    \label{graphmodels_A}
\end{figure*}

Furthermore, certain metrics, such as F1 (Figure~\ref{fig:side:F1}), demonstrate greater sensitivity to changes in data distribution, potentially resulting in fluctuations as data volume increases. This sensitivity may arise because, despite the proportional increase in data from each client, imbalances in data distribution persist across clients. Such variations in data distribution can affect the model's performance across different categories. Additionally, achieving higher accuracy is more feasible with Acc@5 than with Acc@1, and with Acc@10 than with Acc@5, as evidenced by a comparison of the vertical axes in Figures~\ref{fig:side:Acc@1}, ~\ref{fig:side:Acc@5}, and ~\ref{fig:side:Acc@10}. Observations from the performance of Acc@10, Rec@10, and Pre@10 does not increase linearly when the data volume reaches 50k. This suggests that while increasing data volume enhances model performance, it also introduces changes in data distribution, which in turn impacts the model's ability to recognize different categories.

\subsubsection{Batch Size Sensitivity}  
Larger batch sizes typically offer more stable gradient estimates for client computation, which can enhance model generalization. However, they also increase memory usage and may extend training time. Figure 3 shows that while our method's performance improves with larger batch sizes, the rate of improvement diminishes. When the batch size reaches a certain point, the positive effects begin to plateau. This may occur because larger batch sizes can make it harder for the model to capture small-scale variations in the data, limiting its ability to learn detailed features from different data distributions.

\begin{figure*} 
  \begin{minipage}[t]{0.14\linewidth} 
  \subfigure[{\scriptsize Llama3.1-8B:NumRS}]{
    \includegraphics[width=1.in]{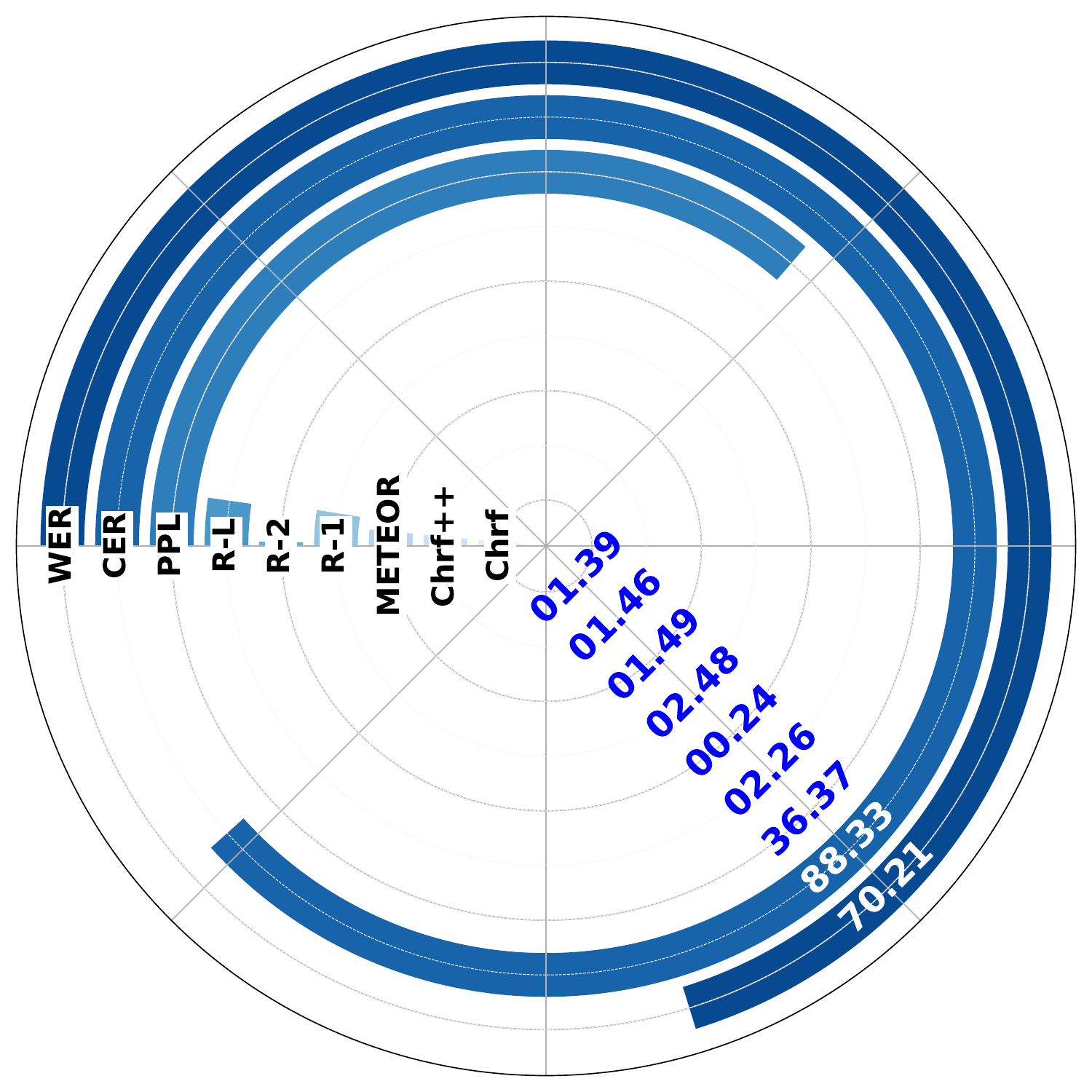}}
  \end{minipage}\hspace{0.025\linewidth}%
  \begin{minipage}[t]{0.14\linewidth} 
    \subfigure[{\scriptsize Llama3.1-8B:InfoEX}]{
      \includegraphics[width=1.in]{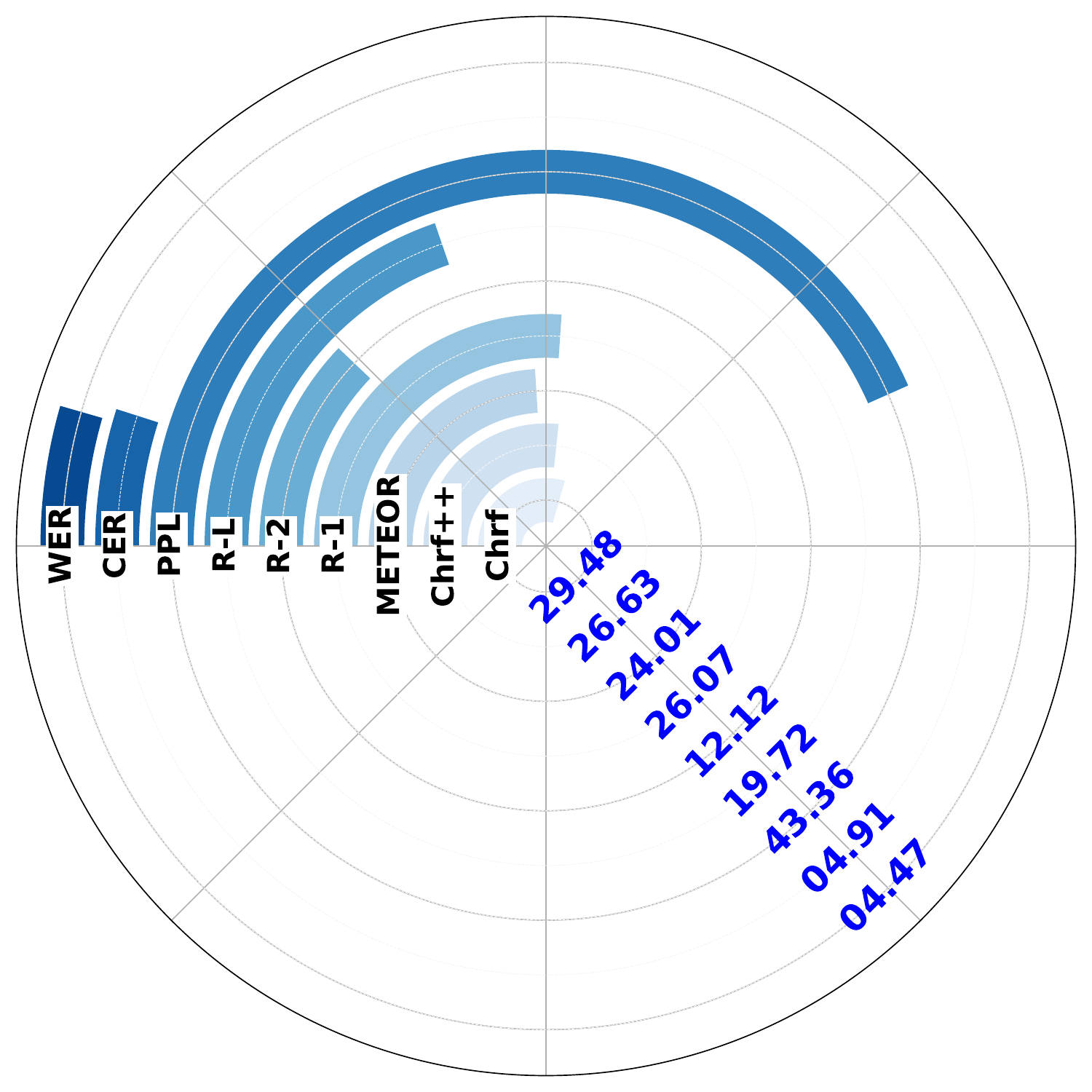}}
    \end{minipage}\hspace{0.025\linewidth}%
    \begin{minipage}[t]{0.14\linewidth} 
      \subfigure[{\scriptsize Llama3.1-8B:LogcRS}]{
        \includegraphics[width=1.in]{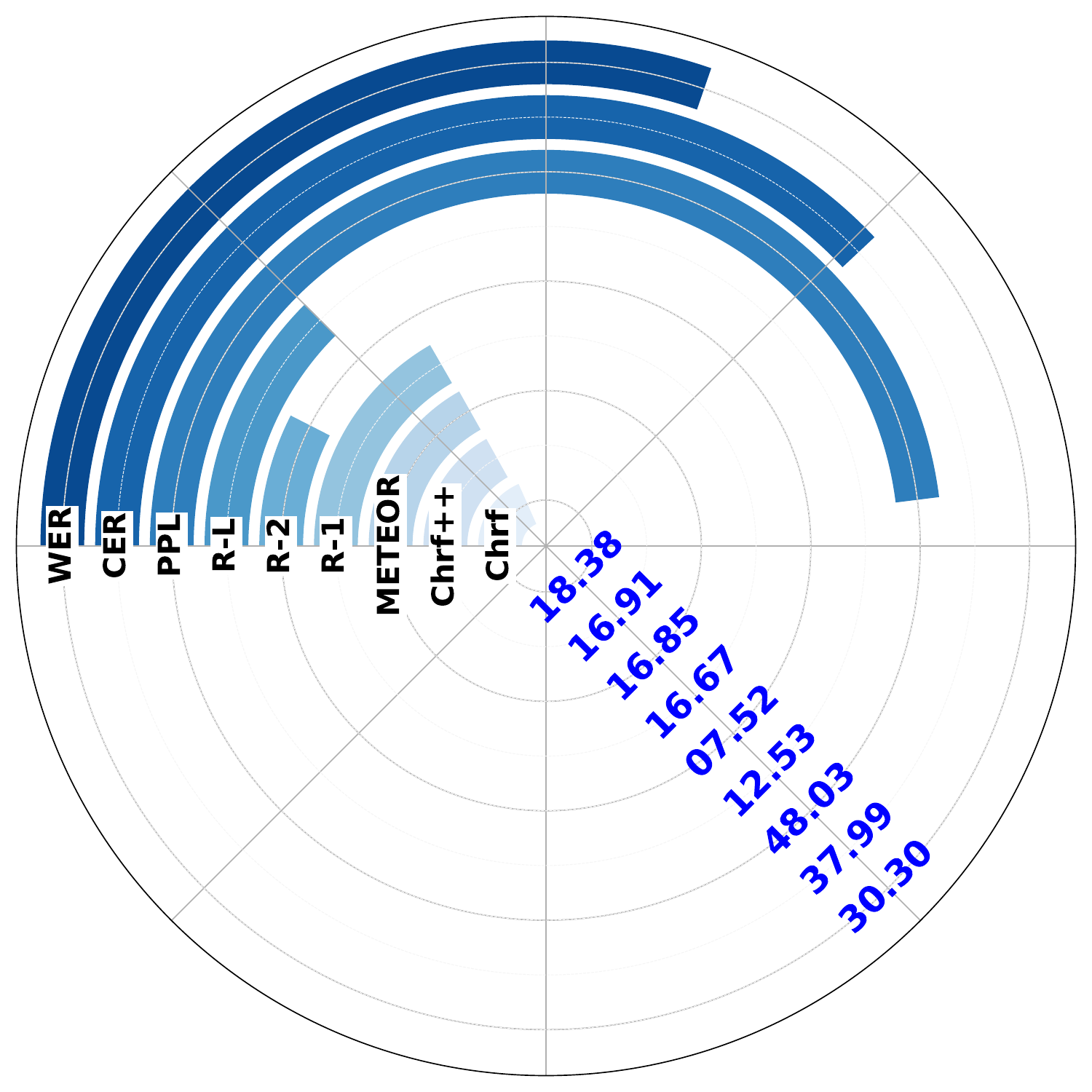}}
      \end{minipage}\hspace{0.025\linewidth}%
      \begin{minipage}[t]{0.14\linewidth} 
        \subfigure[{\scriptsize GPT-4o Mini:NumRS}]{
          \includegraphics[width=1.in]{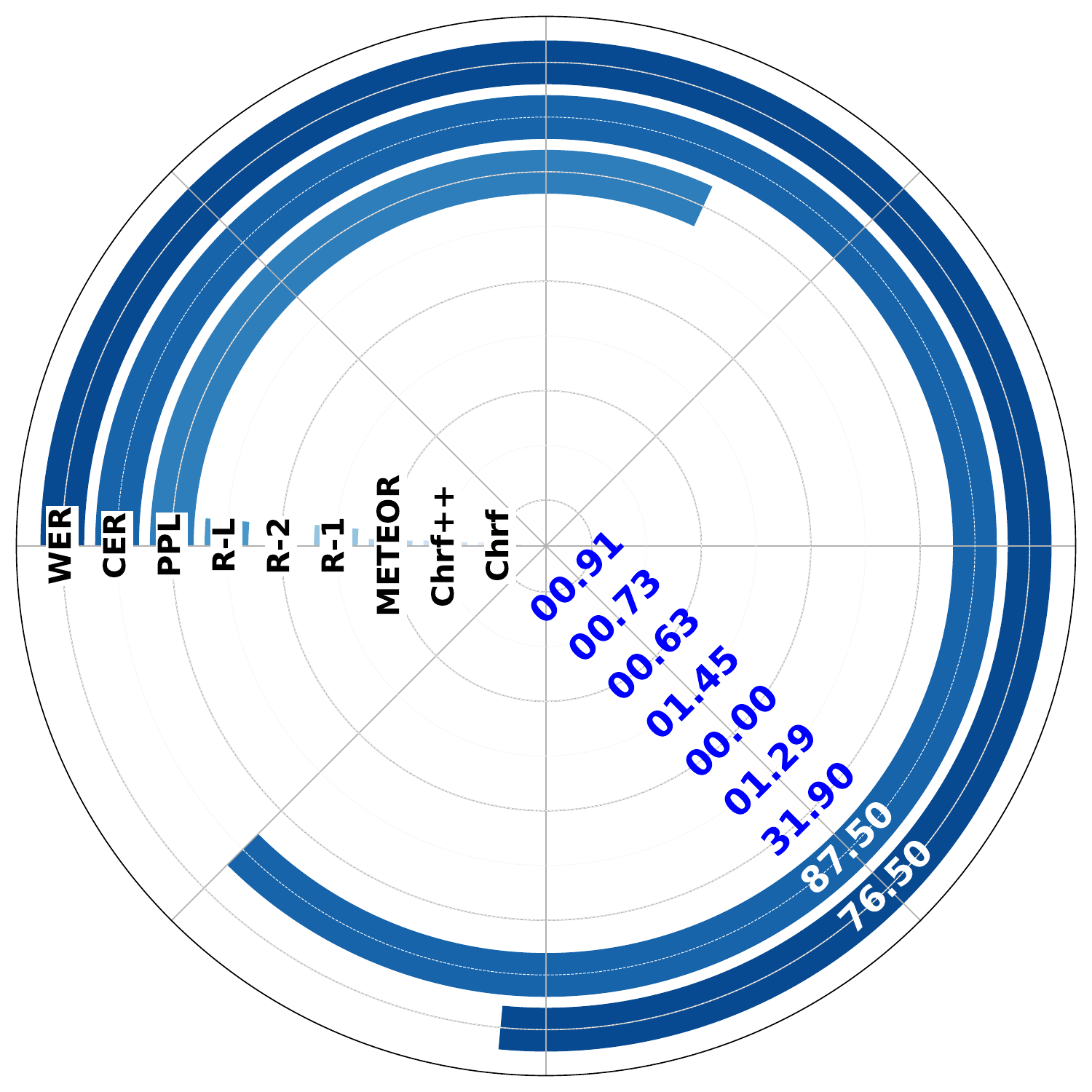}}
        \end{minipage}\hspace{0.025\linewidth}%
        \begin{minipage}[t]{0.14\linewidth} 
          \subfigure[{\scriptsize GPT-4o Mini:InfoEX}]{
            \includegraphics[width=1.in]{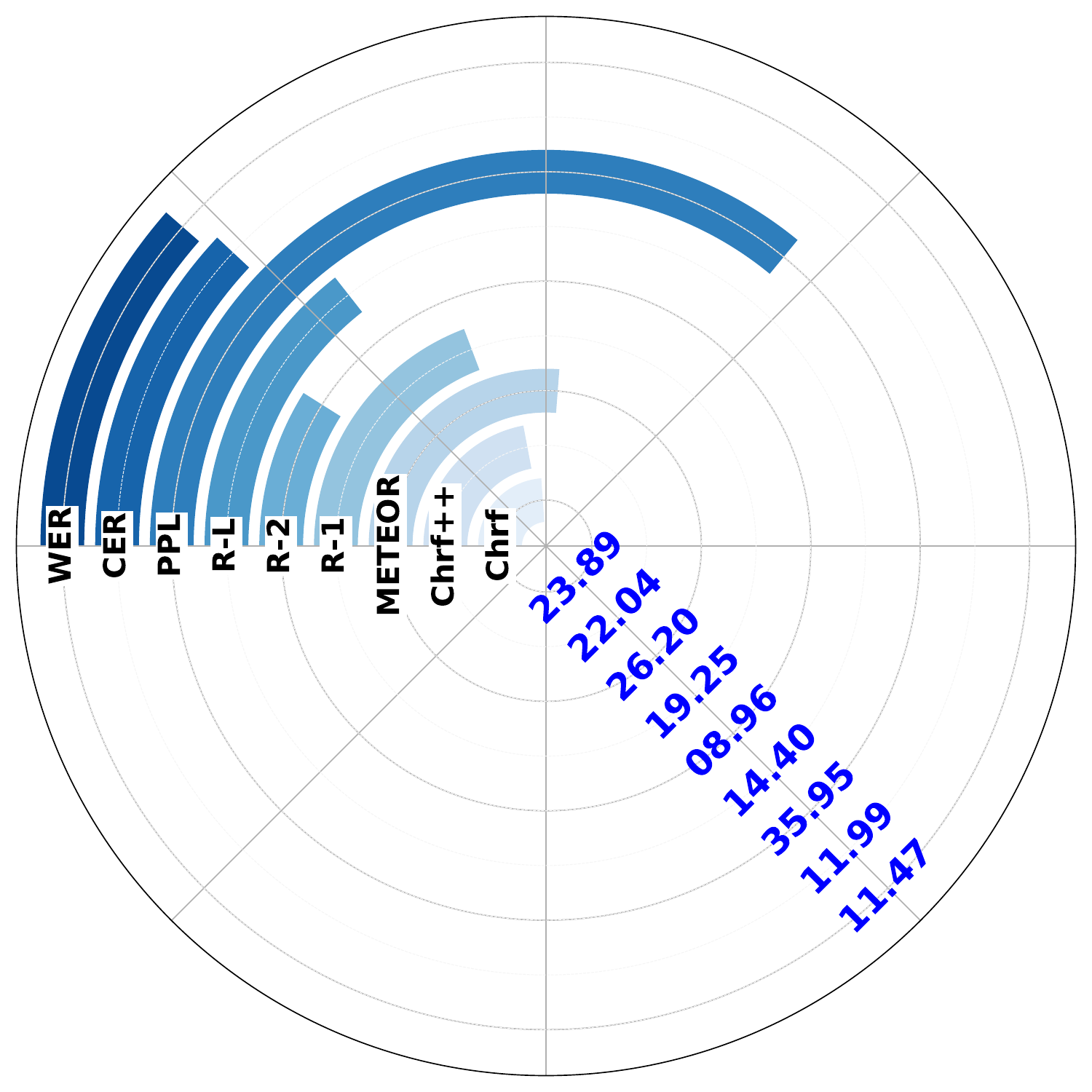}}
          \end{minipage}\hspace{0.025\linewidth}%
 \begin{minipage}[t]{0.14\linewidth} 
        \subfigure[{\scriptsize GPT-4o Mini:LogcRS}]{
          \includegraphics[width=1.in]{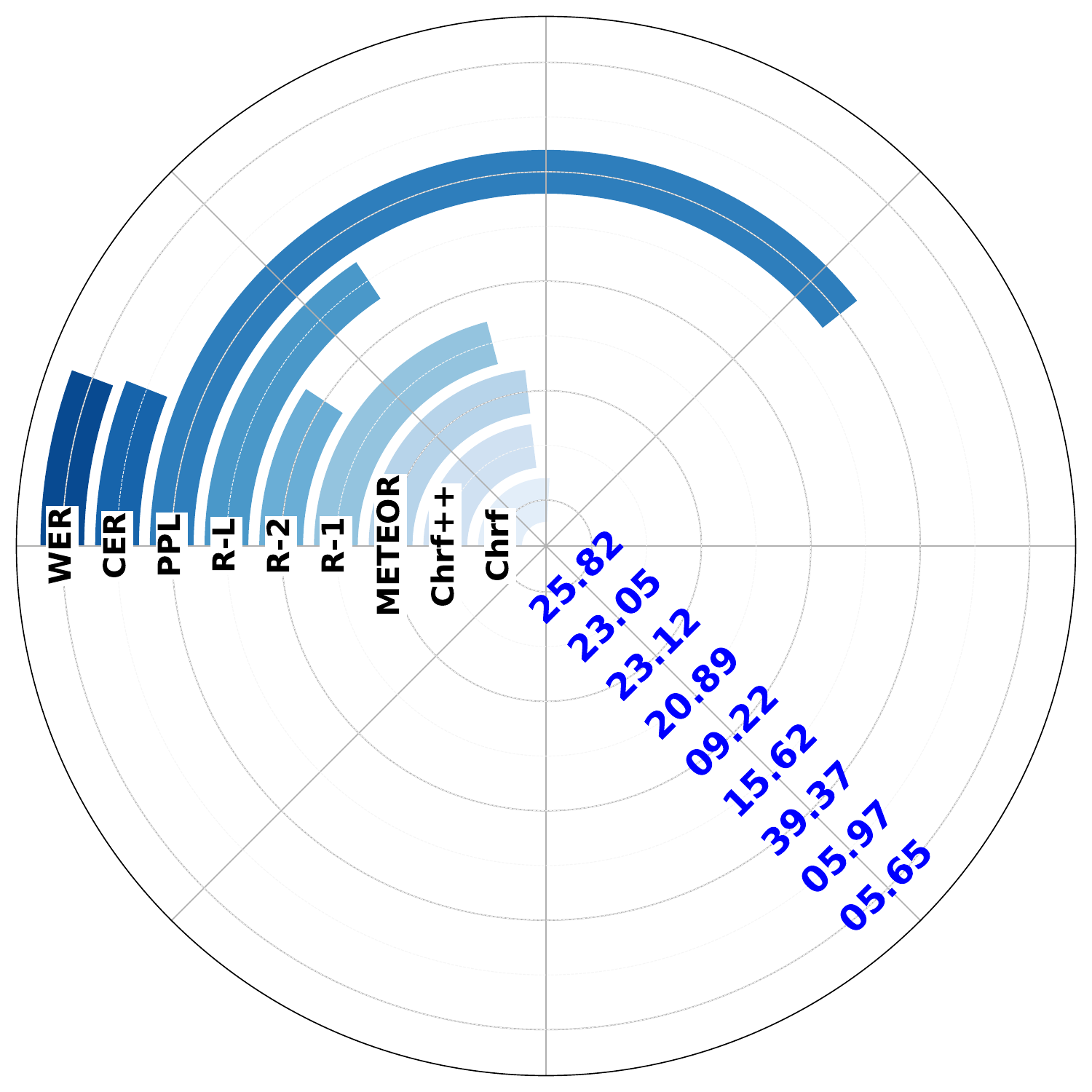}}
        \end{minipage}\hspace{0.025\linewidth}%

        \begin{minipage}[t]{0.14\linewidth} 
          \subfigure[{\scriptsize MathStral-7B:NumRS}]{
            \includegraphics[width=1.in]{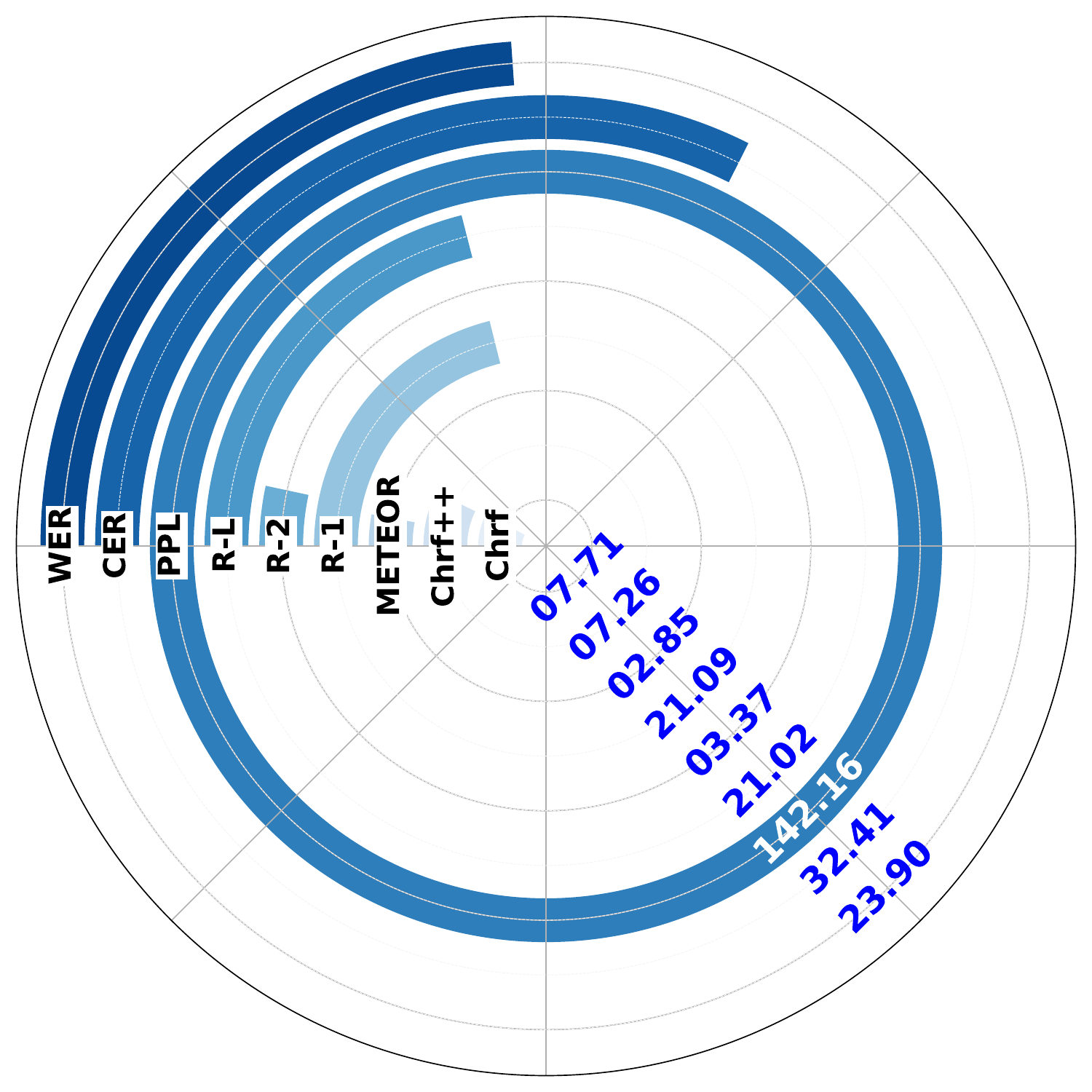}}
          \end{minipage}\hspace{0.025\linewidth}%
          \begin{minipage}[t]{0.14\linewidth} 
            \subfigure[{\scriptsize MathStral-7B:InfoEX}]{
              \includegraphics[width=1.in]{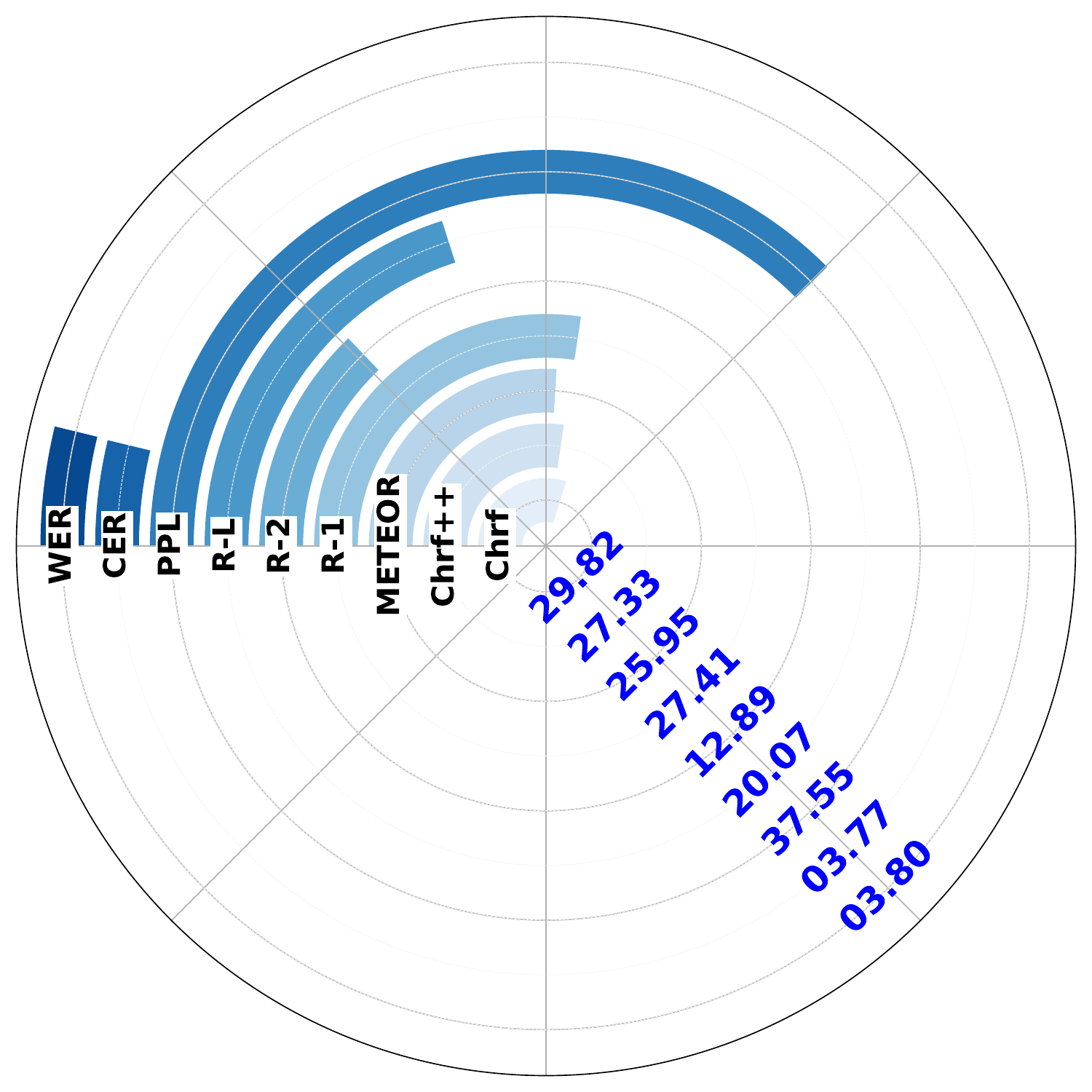}}
            \end{minipage}\hspace{0.025\linewidth}%
            \begin{minipage}[t]{0.14\linewidth} 
              \subfigure[{\scriptsize MathStral-7B:LogcRS}]{
                \includegraphics[width=1.in]{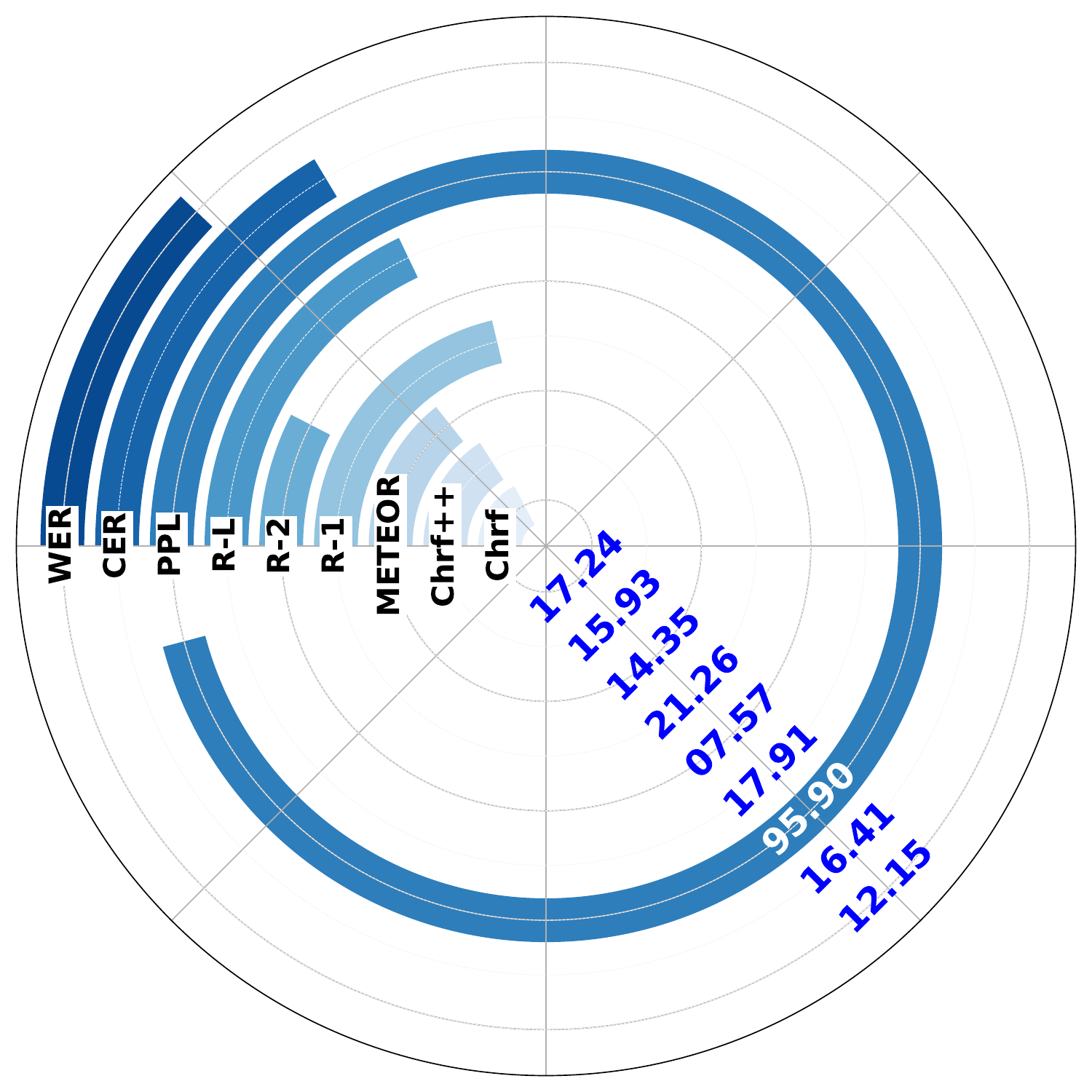}}
              \end{minipage}\hspace{0.025\linewidth}%
              \begin{minipage}[t]{0.14\linewidth} 
                \subfigure[{\scriptsize DS R1-7B:NumRS}]{
                  \includegraphics[width=1.in]{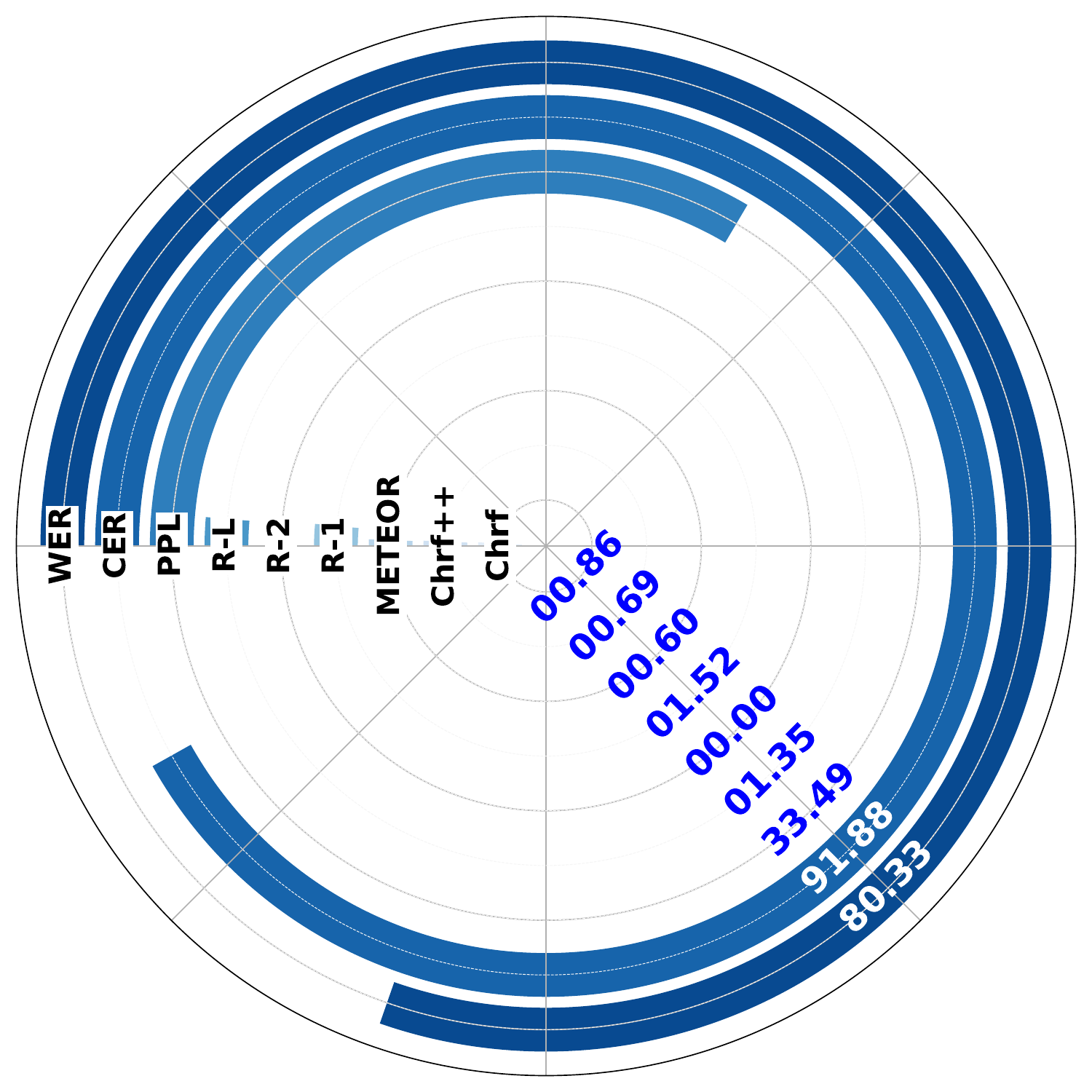}}
                \end{minipage}\hspace{0.025\linewidth}%
                \begin{minipage}[t]{0.14\linewidth} 
                  \subfigure[{\scriptsize DS R1-7B:InfoEX}]{
                    \includegraphics[width=1.in]{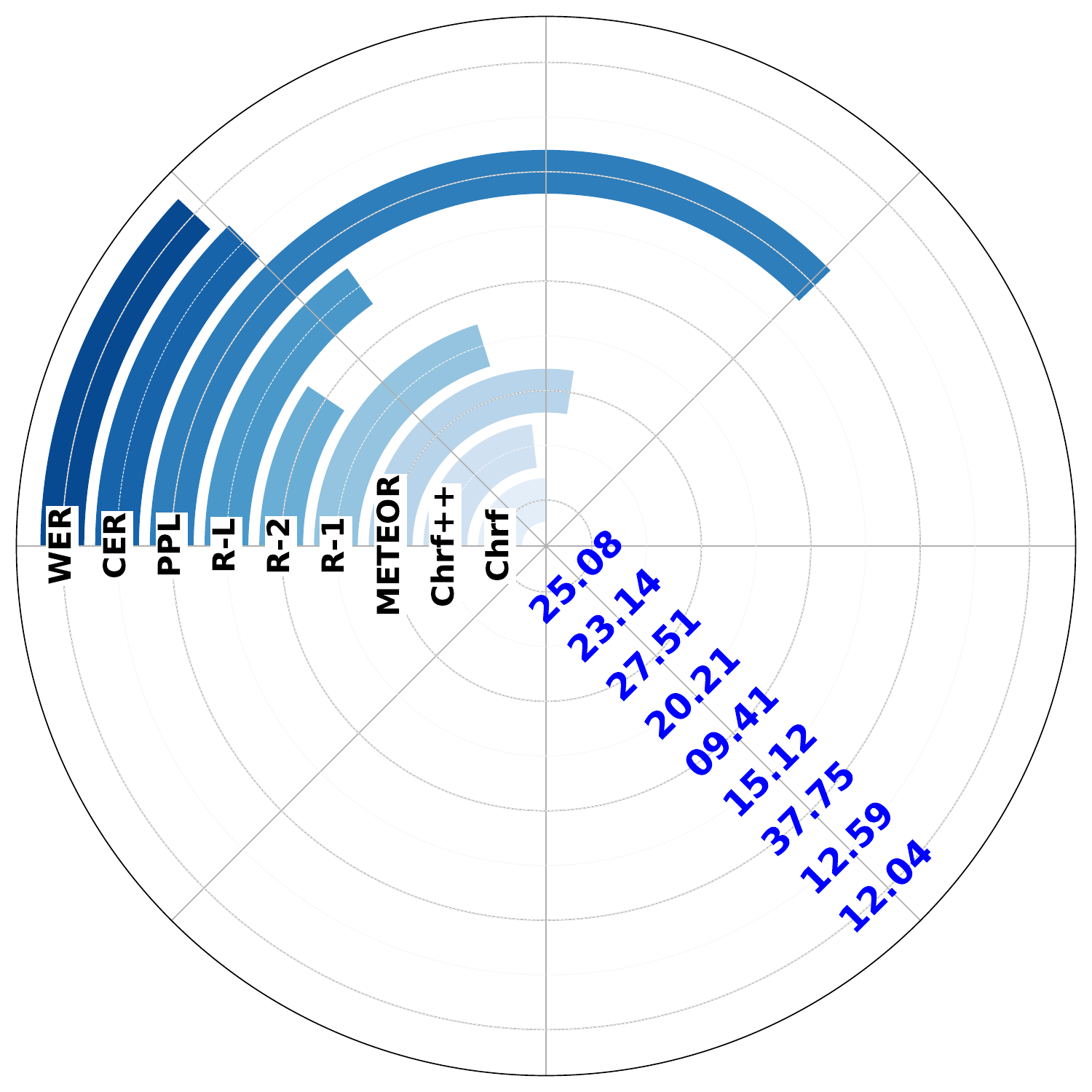}}
                  \end{minipage}\hspace{0.025\linewidth}%
         \begin{minipage}[t]{0.14\linewidth} 
                \subfigure[{\scriptsize DS R1-7B:LogcRS}]{
                  \includegraphics[width=1.in]{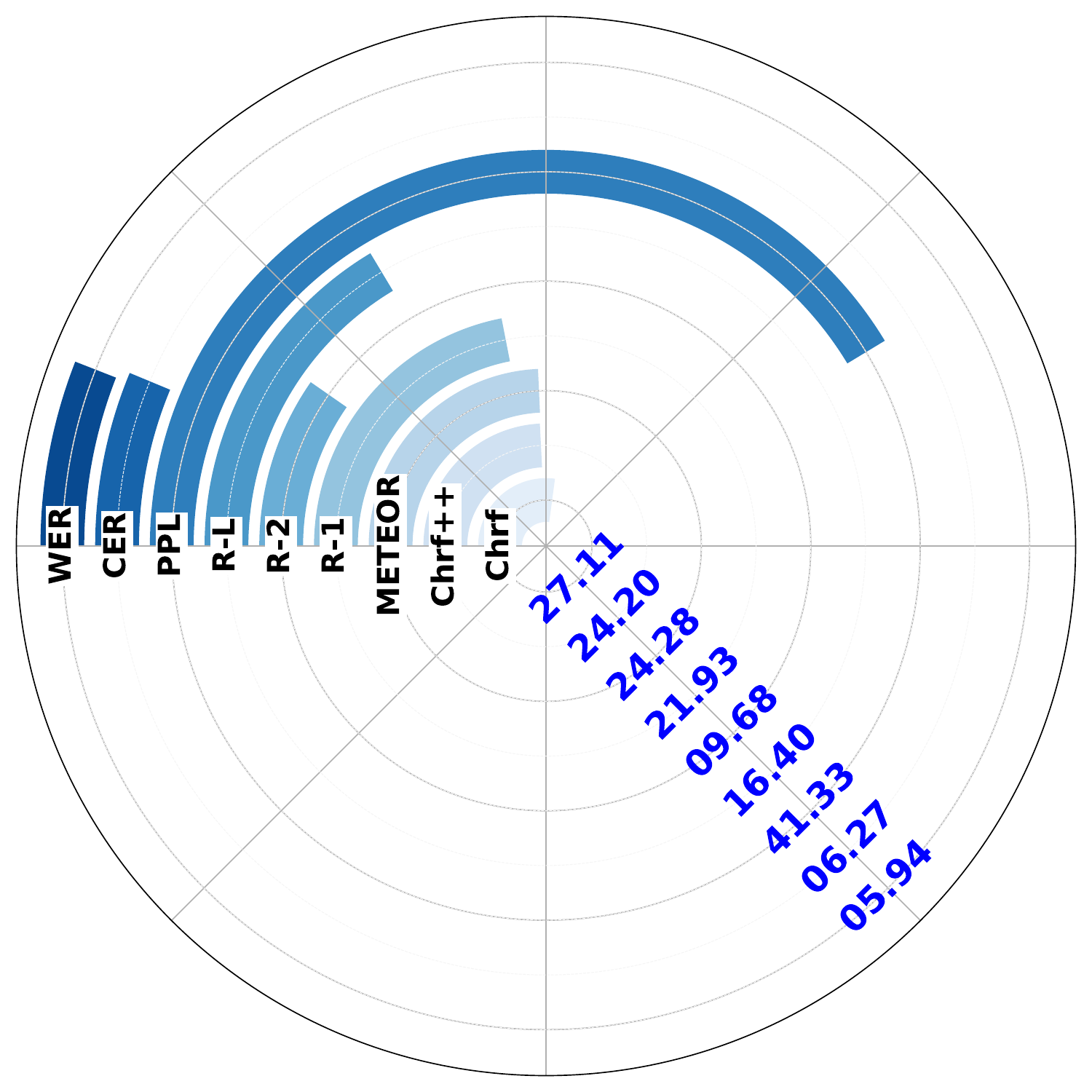}}
                \end{minipage}\hspace{0.025\linewidth}%
    \caption{Circular bar chart illustrations of question-answering performance on the test set, delineating various query types. The chart encompasses Domain-related queries (DR-Q), Metric-based queries (MB-Q), and Novel-based queries (NB-Q). Lower values of PPL, CER, and WER indicate better performance.}
    \label{graphmodels_B}
\end{figure*}

\subsection{Main Results of Downstream RAG Generation}
\label{Main_Results_upstream}
Tables~\ref{LLM_TEST} and ~\ref{LLM_validation} illustrate the performance of various large language models (LLMs) when integrated with FedE4RAG for downstream retrieval-augmented generation (RAG) tasks. Notably, GPT-4o Mini excels in both test and validation environments, attaining superior scores across the majority of evaluated metrics. This suggests that the integration of GPT-4o Mini with FedE4RAG yields exceptional results in generating high-quality outputs for these tasks.

Crafted with a multilingual focus, Llama3.1-8B reveals  slightly inferior performance in text generation  to other models. GPT-4o Mini, inheriting the robust language understanding and generation capabilities of the GPT series, accurately interprets user queries and provides reasonable responses. It excels in semantic understanding, analyzing complex sentence structures and semantic relationships, and outperforms others in information retrieval tasks, as evidenced by its superior performance on N-gram overlap metrics. Additionally, its favorable performance on edit distance metrics (CER and WER) highlights its advantage in text generation accuracy.
A testament to the GPT series' prowess in language understanding and generation, GPT-4o Mini shines brightly.  It masterfully decodes user queries and churns out sensible responses. 
The model adeptly handles complex multi-step logical reasoning tasks, such as mathematical proofs and scientific computations. However, its specialized focus may limit its broader applications compared to GPT-4o Mini or Llama3.1-8B.

In our experimental analysis, we observed that Perplexity (PPL) serves as a pivotal indicator of a language model's predictive accuracy. This metric essentially quantifies the model's uncertainty when processing text sequences, with a lower PPL value denoting a higher degree of confidence and precision in the model's predictions. Essentially, a lower PPL score suggests that the model is better at anticipating the next word in a sequence, which is crucial for generating coherent and contextually relevant text. In the provided test data, DeepSeek R1-7B achieves the lowest PPL value of 36.19 among all models, suggesting potential superior performance in text generation tasks. This low PPL value implies higher certainty in text processing, enabling more accurate word predictions and resulting in more fluent and precise text generation. Moreover, lower perplexity indicates better model performance and more accurate sequence prediction. Therefore, DeepSeek R1-7B's characteristics may make it suitable for tasks requiring high text output accuracy. Although DeepSeek R1-7B and Llama3.1-8B show competence in certain areas, they do not surpass GPT-4o Mini overall. This suggests that the selection of the most appropriate LLM for integration with RAG systems should be based on specific task requirements.

\subsection{Downstream Reasoning Analysis}
\label{Downstream_Reasoning}
For domain-specific queries (subfigures a, d, g and g of Figure~\ref{graphmodels_A}), several LLMs exhibit superior performance in terms of edit distance metrics, evidenced by low Word Error Rates (WER) and Character Error Rates (CER). For instance, the best-performing model, GPT-4o mini, achieves a CER of 1.32 and a WER of 1.46, indicating that the model's outputs closely align with the correct answers. In contrast, for metric-based queries (subfigures b, e, h and k of Figure~\ref{graphmodels_A}), these metrics deteriorate, as demonstrated by the higher error rates of models such as Llama3.1-8B and DeepSeek R1-7B in subfigures b and k of Figure~\ref{graphmodels_A}. This suggests that there is room for improvement in these models' data reasoning capabilities. Conversely, GPT-4o and the data-reasoning-specialized MathStral-7B model show a slight advantage in these metrics.
For novel-based queries  (subfigures c, f, i and l of Figure~\ref{graphmodels_A}), both GPT-4o Mini and MathStral-7B exhibit weaknesses, as indicated by their elevated Perplexity (PPL) scores. Specifically, GPT-4o Mini records PPL values of 108.29 and 113.66 in subfigures f and i of Figure~\ref{graphmodels_A}, respectively. Higher PPL values signify a greater discrepancy between the model's predictions and the actual text distribution, reflecting a lower grasp of the text. GPT-4o Mini demonstrates consistent performance across these three query types, even surpassing models specifically trained for reasoning, such as MathStral-7B and DeepSeek R1-7B.

Upon analyzing Figure~\ref{graphmodels_B}, we observe that several LLM models exhibit poor WER and CER performance on numerical reasoning tasks. This may be attributed to the nature of the answers, which are primarily numerical. These tasks demand not only computational skills from the models but also precision in character representation, thereby increasing the difficulty of inference. In contrast, for information extraction tasks, the models demonstrate stable performance, as evidenced by subfigures b, e, h, and k of Figure~\ref{graphmodels_B}. Regarding logical reasoning, DeepSeek shows a clear advantage, as indicated by subfigures f and h of Figure~\ref{graphmodels_B}.

\section{Conclusion and Future Work}\label{sec:conclu}
 Our framework addresses the challenges of privacy preservation in private RAG systems by leveraging federated learning and knowledge distillation. 
 Through extensive experiments on legal and financial datasets, we demonstrated that FedE4RAG can effectively improve the performance of localized RAG retrievers while maintaining data privacy.

 The key contributions of this work are as follows: We proposed the FedE4RAG framework, which enables collaborative training of client RAG retrieval models in a privacy-preserving manner using federated learning.
 We introduced Federated knowledge distillation to facilitate the transfer of local knowledge onto a global scale, enhancing the generalization of local RAG retrievers.
 We implemented homomorphic encryption to safeguard model parameters and mitigate data leakage concerns during the federated learning process. 
 We conducted comprehensive experiments on two real-world datasets, validating the effectiveness of FedE4RAG in improving localized RAG performance while preserving data privacy.

 For future work, we plan to explore the following directions: 
 Scalability and Efficiency: We aim to optimize the FedE4RAG framework for larger-scale deployments and improve its computational efficiency, particularly in terms of communication costs and training time.
 Advanced Privacy-Preserving Techniques: We plan to investigate more advanced privacy-preserving techniques, such as differential privacy, to further enhance the privacy guarantees provided by our framework.
 Broader Domain Applications: We intend to apply and evaluate FedE4RAG in other domains beyond legal and financial, such as healthcare and education, to assess its generalizability and adaptability.
 Robustness and Security Analysis: We analyze the framework's resilience to attacks like model inversion and membership inference, implementing countermeasures as needed.

\ifCLASSOPTIONcompsoc
  \section*{Acknowledgments}
\else
  \section*{Acknowledgment}
\fi
This work is supported by Zhongguancun Laboratory. We sincerely thank the Beijing Advanced Innovation Center for Future Blockchain and Privacy Computing for their invaluable support and guidance throughout the course of this research. Their practical insights and suggestions have been instrumental in shaping the direction of this work and enhancing its applicability to real-world scenarios. We are also grateful to all the individuals and institutions that have contributed to this project in various ways.




\ifCLASSOPTIONcaptionsoff
  \newpage
\fi




\normalem

{\footnotesize 
\bibliographystyle{ieeetr}
\bibliography{IEEEabrv}
%



%

\vspace{-0.4in}
\begin{IEEEbiography}[{\includegraphics[width=1in,height=1.28in]{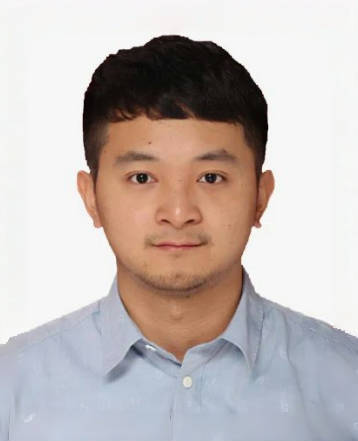}}]{Qianren Mao} is currently an Associate Researcher at the Zhongguancun Laboratory, Beijing, China. He has earned his Ph.D. from the Beihang University. His primary research areas include large language model (LLM) reasoning, knowledge graph reasoning, and data security for LLM. He has published research papers in top-tier journals and conferences, including \textit{IEEE/ACM Transactions on Audio, Speech and Language Processing}, ICML, KDD, SIGIR, IJCAI, EMNLP, NAACL, ICDM, and COLING, etc. 
\end{IEEEbiography}

\vspace{-0.4in}
\begin{IEEEbiography}[{\includegraphics[width=1in,height=1.27in]{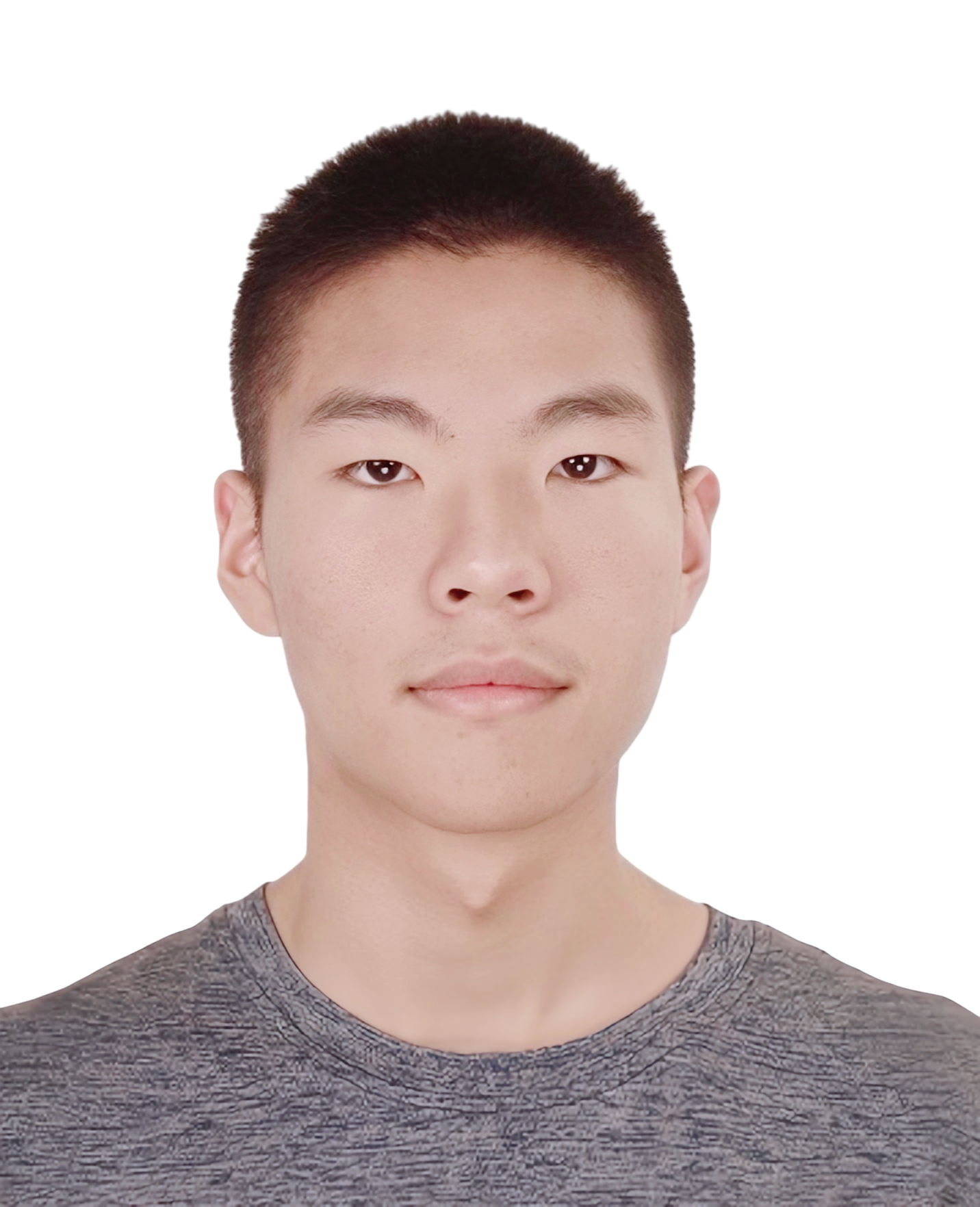}}]{Qili Zhang} is currently pursuing a Ph.D. degree at the National Superior College for Engineers, Beihang University. He has received the BS degree from School of Computer Science \& Engineering at Beihang University. His research interests focus on data security for LLMs, Agents and RAG systems.
\end{IEEEbiography}

\vspace{-0.4in}
\begin{IEEEbiography}[{\includegraphics[width=1in,height=1.27in]{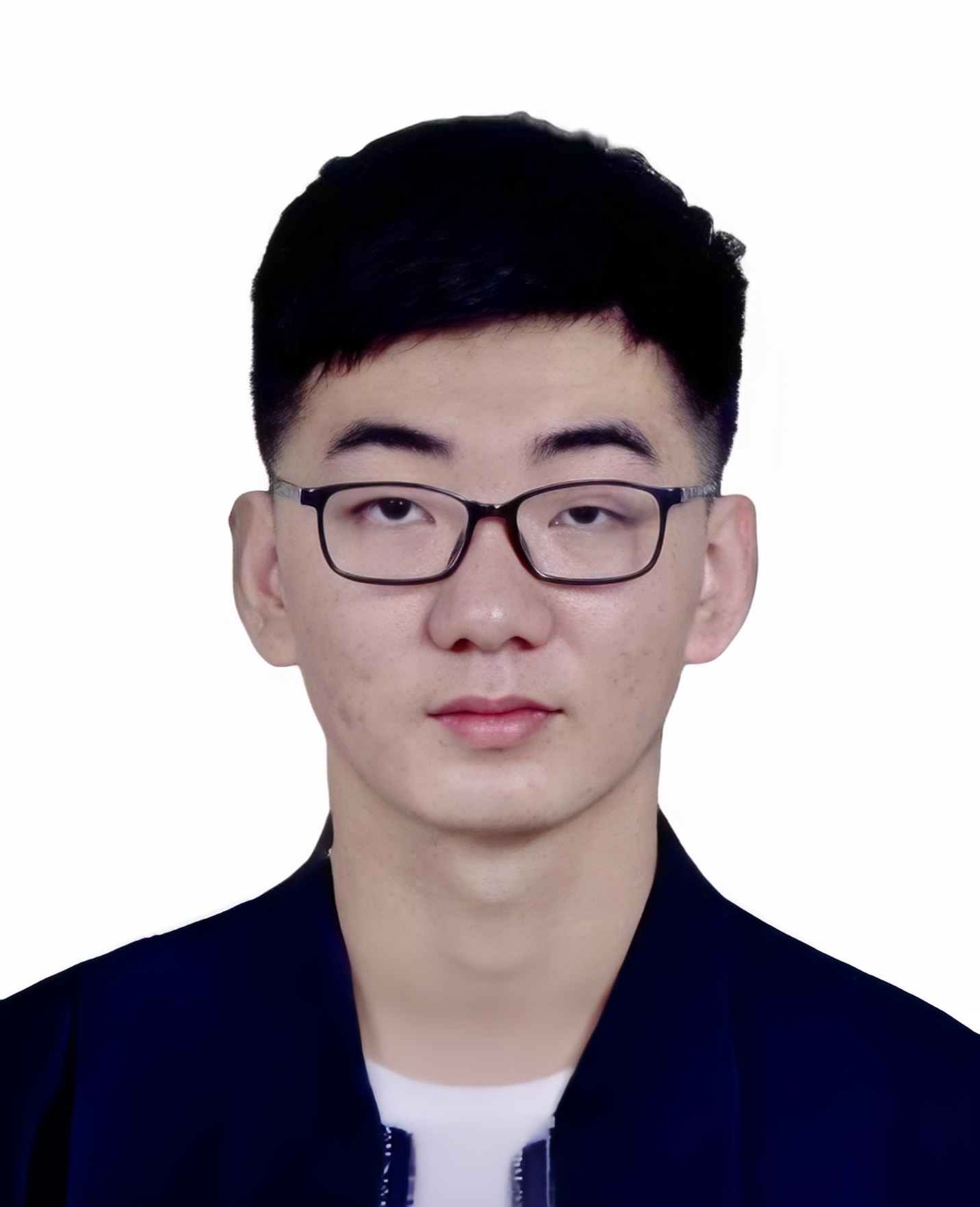}}]{Hanwen Hao} is currently pursuing the ME degree at the School of Computer Science \& Engineering, Beihang University. He has received the BE degree from the School of Software Engineering at Beihang University. His research interests include model security and RAG (Retrieval-Augmented Generation), focusing on improving the robustness of AI systems and the reliability of generative content.
\end{IEEEbiography}

\vspace{-0.4in}
\begin{IEEEbiography}[{\includegraphics[width=1in,height=1.26in]{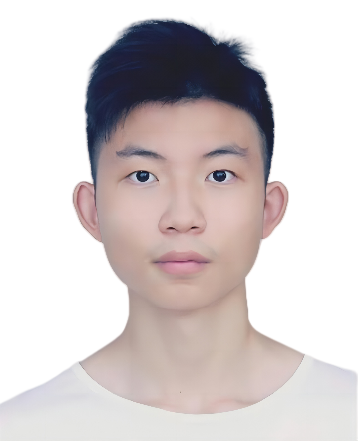}}]{Zhentao Han} is currently working toward the MS degree with the Database Systems Research Group in the School of Information at Renmin University of China. He has received the BS degree from the College of Computer Science and Technology in Beihang University. His research interests include AI for database systems and multimodal data processing.
\end{IEEEbiography}

\vspace{-0.4in}
\begin{IEEEbiography}[{\includegraphics[width=1in,height=1.28in]{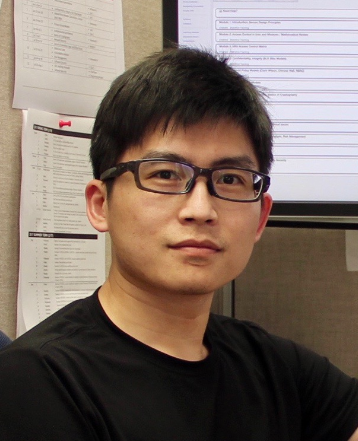}}]{Runhua Xu} (Member, IEEE) is currently a professor at the School of Computer Science and Engineering, Beihang University. Previously, he was a Research Staff Member at IBM Research - Almaden Lab. He has earned his Ph.D. from the University of Pittsburgh. His research focuses on enhancing privacy and trustworthiness in various computing domains, specializing in AI security and privacy solutions. He has published research papers in leading journals and conferences, such as \textit{IEEE Transactions on Dependable and Secure Computing}, \textit{IEEE Transactions on Information Forensics and Security}, \textit{ACM Transactions on Internet Technology}, NeurIPS, SRDS, ICDCS and ACM CCS.
\end{IEEEbiography}

\vspace{-0.4in}
\begin{IEEEbiography}[{\includegraphics[width=1in,height=1.26in]{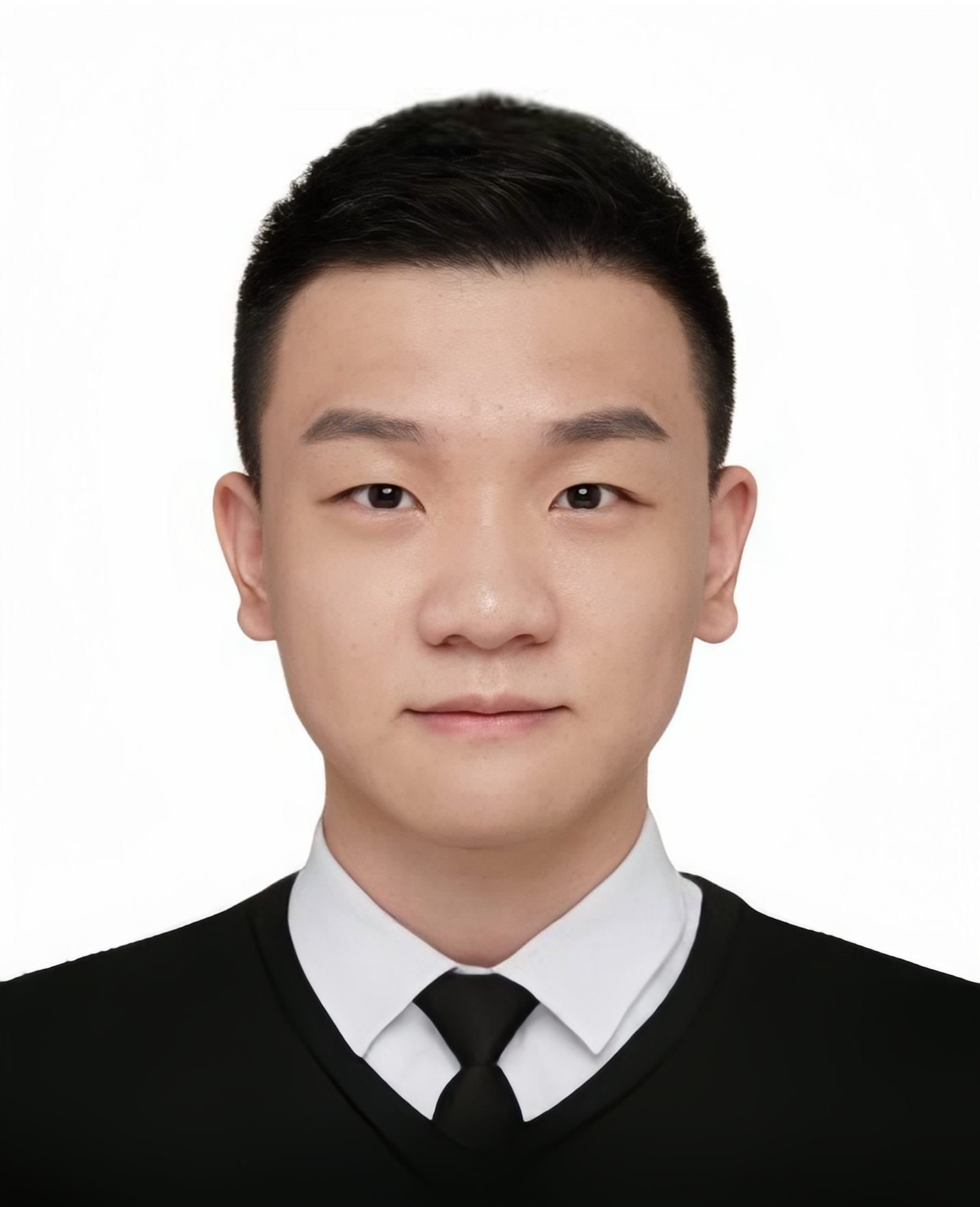}}]{Weifeng Jiang} is currently pursuing a Ph.D. degree in the College of Computing and Data Science at Nanyang Technological University, Singapore. He has earned a Master of Science degree in Artificial Intelligence from Nanyang Technological University in 2024, and a bachelor's degree from Beihang University, China, in 2022. His research interests focus on Trustworthy AI and model interpretability. He has published research papers in top-tier conferences, including ICML, AAAI, EMNLP, NAACL and COLING, etc. 
\end{IEEEbiography}


\vspace{-0.4in}
\begin{IEEEbiography}[{\includegraphics[width=1in,height=1.26in]{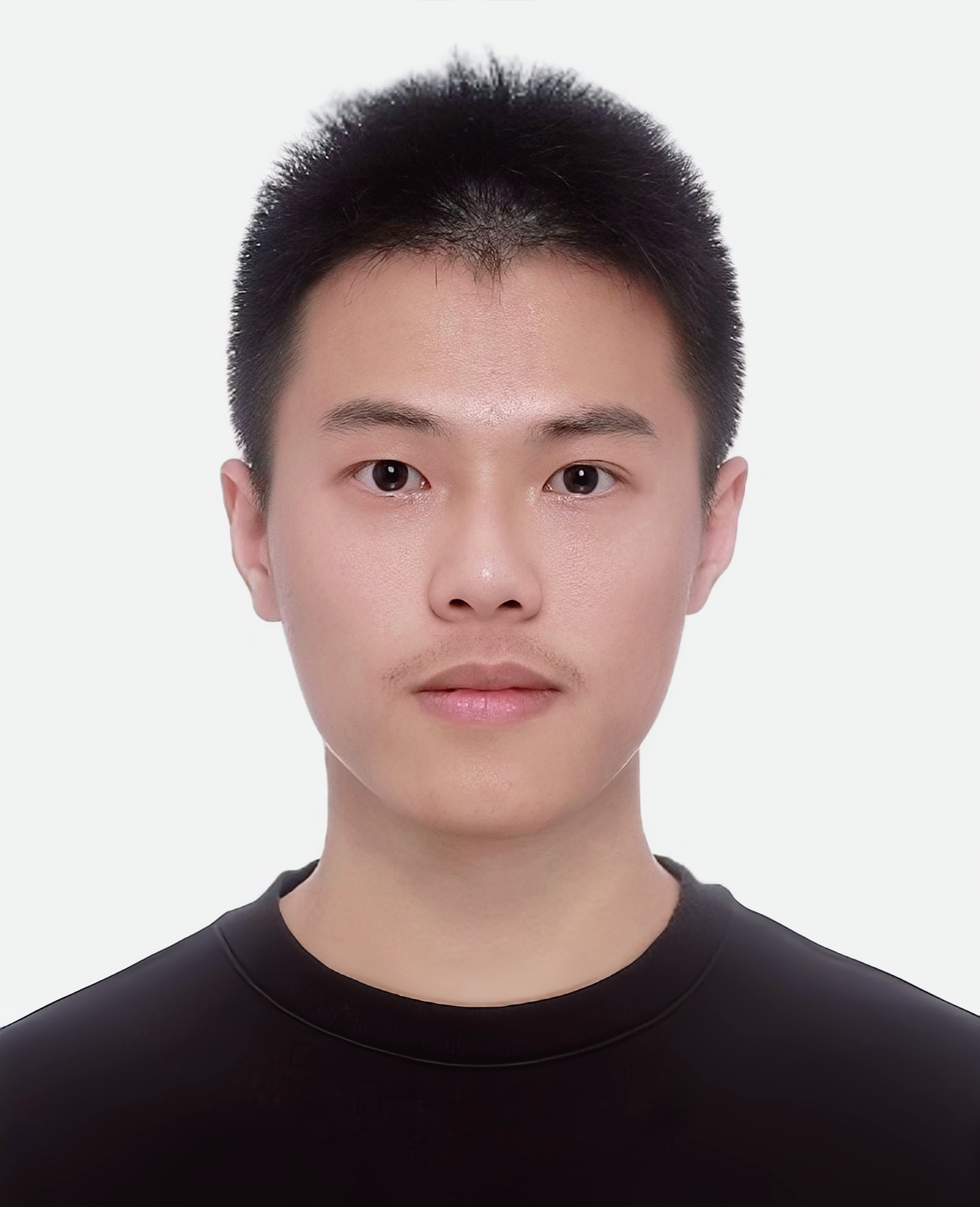}}]{Qi Hu} is currently a Ph.D. candidate at the School of Computer Science and Engineering at Hong Kong University of Science and Technology (HKUST) and he has received the BS degree from University of Science and Technology of China (USTC). His research interests include graph learning and privacy preservation. 
\end{IEEEbiography}

\vspace{-0.4in}
\begin{IEEEbiography}[{\includegraphics[width=1in,height=1.26in]{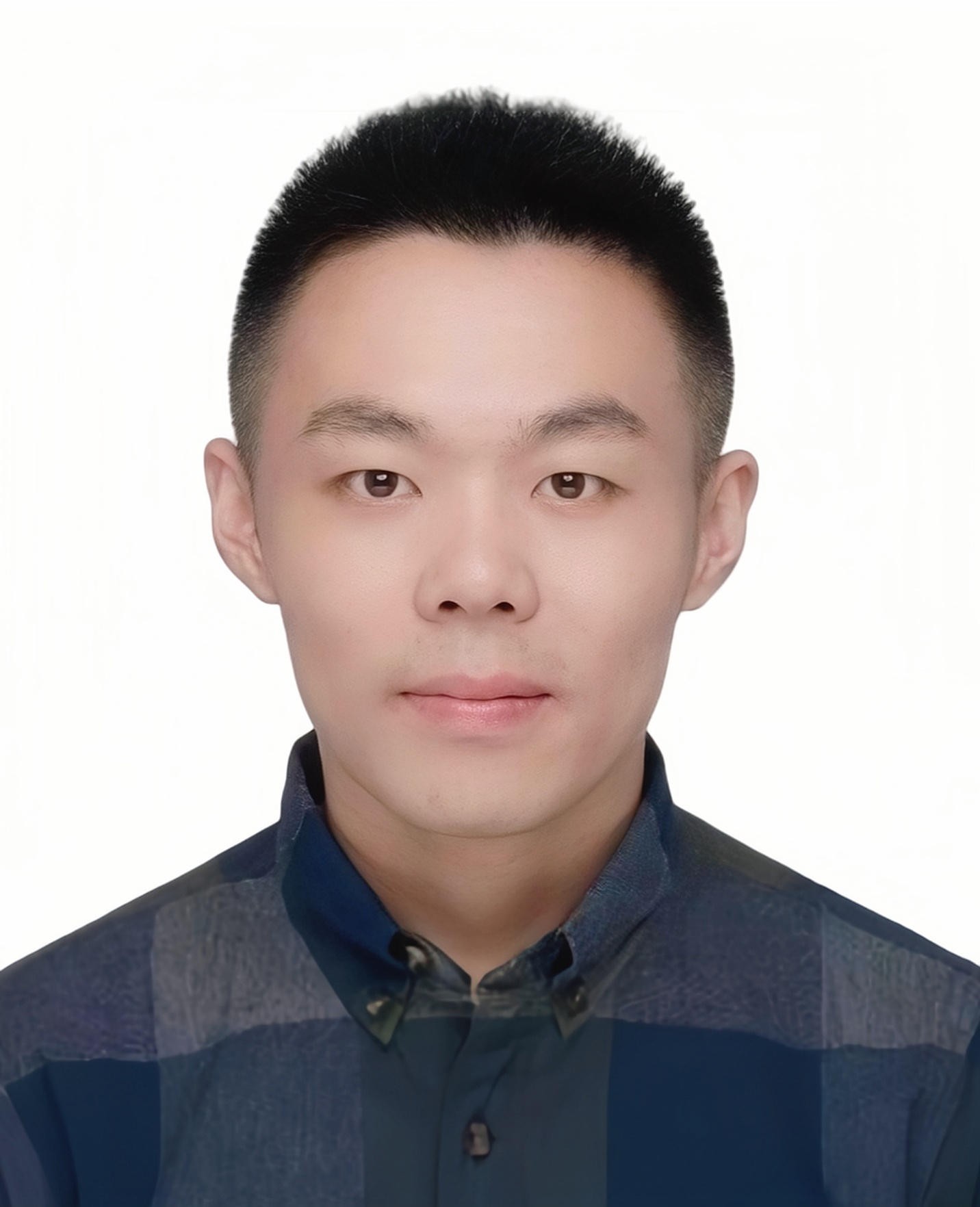}}]{Zhijun Chen} has received his Ph.D. from the School of Computer Science at Beihang University and is preparing to begin postdoctoral research. 
His primary research interests include ensemble algorithms for large language models, weak supervision learning, and its security. He has published research papers in top-tier conferences, including KDD, ICDE, AAAI, and IJCAI, with his KDD 2023 paper recommended as a Best Paper Candidate.
\end{IEEEbiography}

\vspace{-0.4in}
\begin{IEEEbiography}[{\includegraphics[width=1in,height=1.28in]{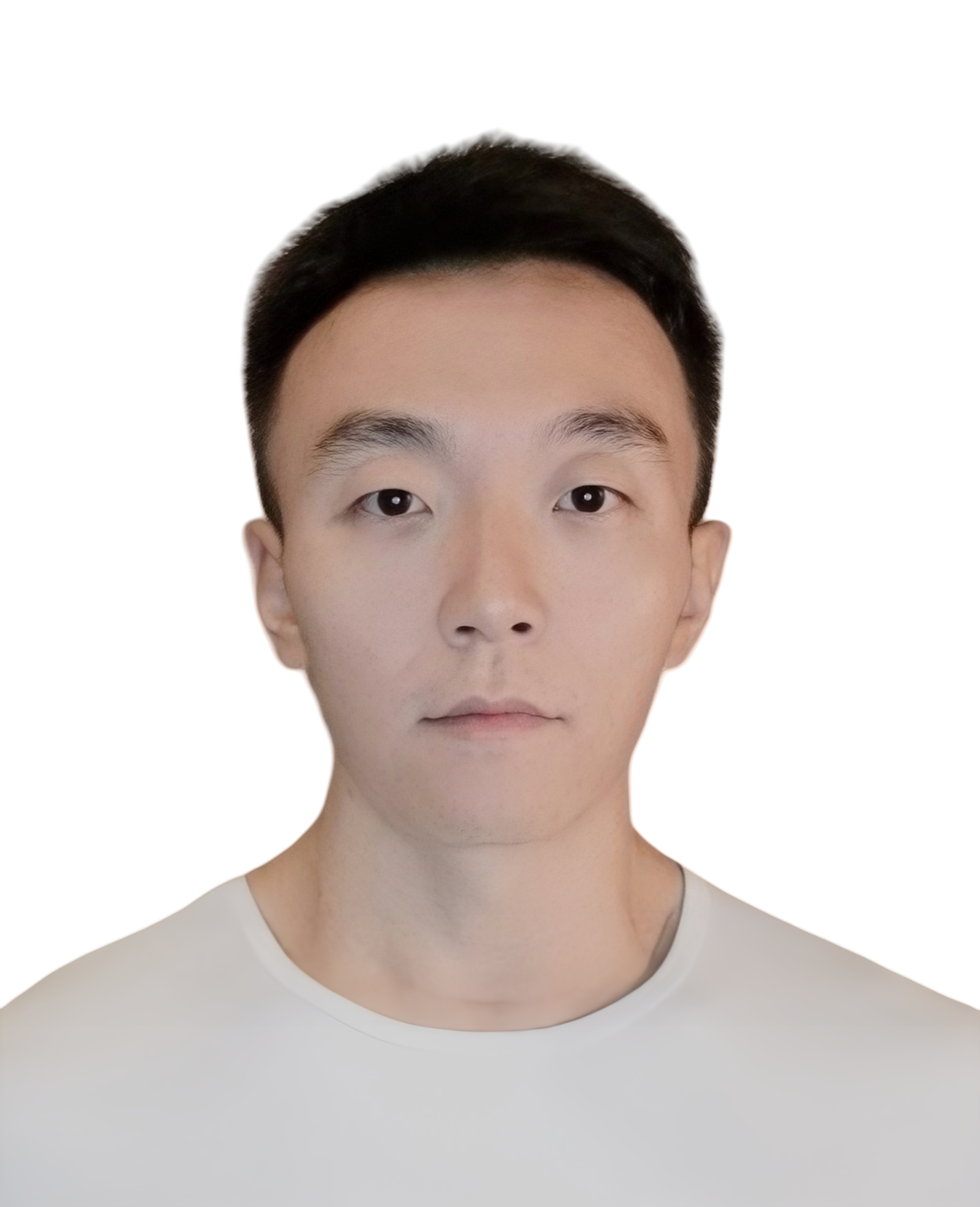}}]{Tyler Zhou} serves as architect of Beijing Academy of Blockchain and Edge Computing, BAEC. Chief Architect of Privecy Computing Product in BAEC.  He has nearly twenty years experience in software development and complex system architecture design. Have worked for many large manufacturers in the IT industry, such as HP, IBM and Huawei. Expert in Communications Technology, Electronic Commerce, Cloud, Block Chain and Privacy Computing area. 
\end{IEEEbiography}

\vspace{-0.4in}
\begin{IEEEbiography}[{\includegraphics[width=1in,height=1.28in]{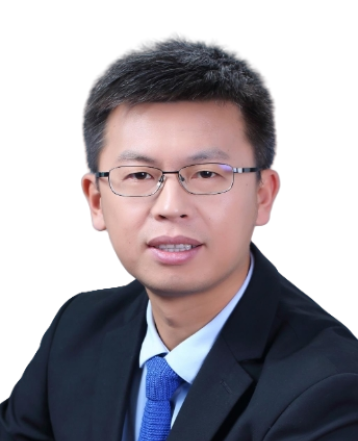}}]{Bo Li} is currently an associate professor at the School of Computer Science and Engineering, Beihang University, and a senior researcher at the Beijing Advanced Innovation Center for Big Data and Brain-inspired Intelligence. His research interests primarily focus on cybersecurity and data security. He has published research papers in top-tier journals and conferences, including \textit{IEEE Transactions on Information Forensics and Security, IEEE Transactions on Knowledge and Data Engineering, IEEE Transactions on Dependable and Secure Computing, and Science China}, AAAI and IJCAI, etc. 
\end{IEEEbiography}

\vspace{-0.4in}
\begin{IEEEbiography}[{\includegraphics[width=1in,height=1.28in]{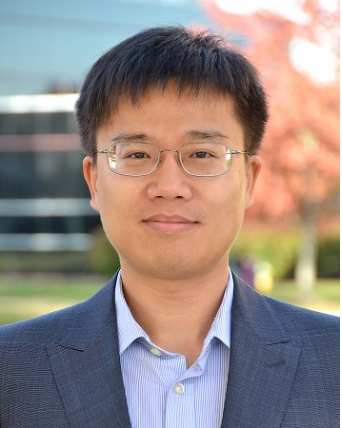}}]{Yangqiu Song} is now an associate professor at Department of CSE at HKUST with a joint appointment at Department of Math and Division of Emerging Interdisciplinary Areas, and an associate director of HKUST-WeBank Joint Lab. His research interests focus on leveraging (commonsense) knowledge and social norms to improve privacy, security, explainability, and fairness issues in machine learning models for NLP. He has published research papers in top-tier journals and conferences, including \textit{Artificial Intelligence, IEEE Transactions on Knowledge and Data Engineering}, ACL, KDD, WWW, NeuralPS, etc. 
\end{IEEEbiography}

\vspace{-0.4in}
\begin{IEEEbiography}[{\includegraphics[width=1in,height=1.28in]{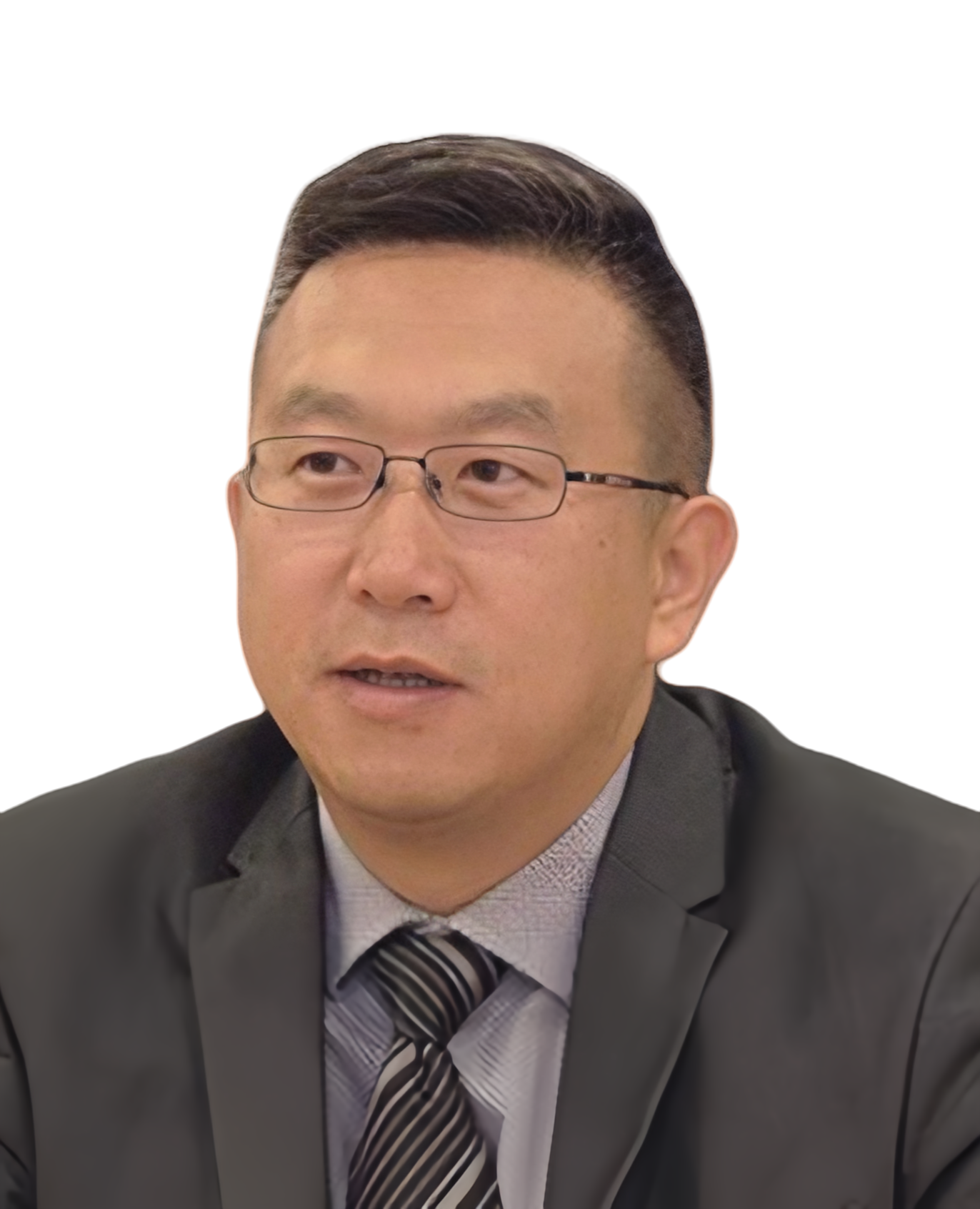}}]{Jin Dong} is the General Director of Beijing Academy of Blockchain and Edge Computing. He is also the General Director of Beijing Advanced Innovation Center for Future Blockchain and Privacy Computing. The team he led developed 'ChainMaker', the first hardware-software integrated blockchain system around the globe. He has been long dedicated in the research areas such as blockchain, artificial intelligence and low-power chip design.
\end{IEEEbiography}

\vspace{-0.4in}
\begin{IEEEbiography}[{\includegraphics[width=1in,height=1.28in]{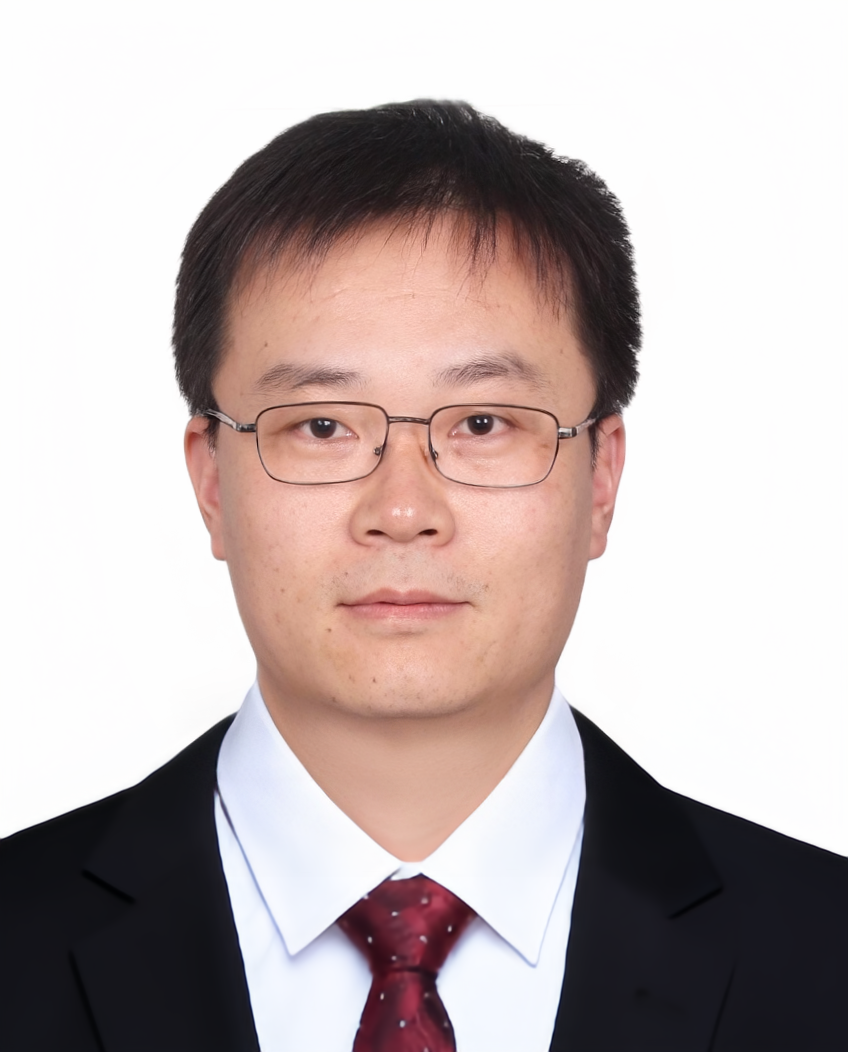}}]{Jianxin Li} (Senior Member, IEEE) is currently a professor with the School of Computer Science and Engineering, and Beijing Advanced Innovation Center for Big Data and Brain Computing in Beihang University. His current research interests include social networks, machine learning, Big Data, and trustworthy computing. He has published research papers in top-tier journals and conferences, including \textit{IEEE Transactions on Knowledge and Data Engineering, IEEE Transactions on Dependable and Secure Computing, Journal of Artificial Intelligence Research, ACM Transactions on Information Systems, ACM Transactions on Knowledge Discovery from Data}, ICML, KDD, AAAI, and WWW, etc. 
\end{IEEEbiography}

\vspace{-0.4in}
\begin{IEEEbiography}[{\includegraphics[width=1in,height=1.28in]{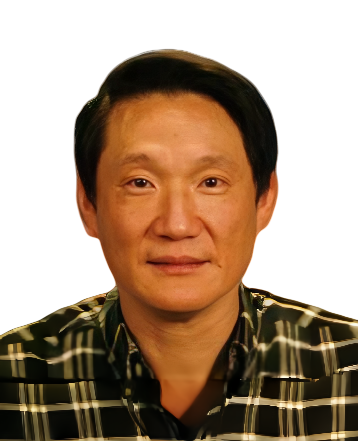}}]{Philip S. Yu} (Life Fellow, IEEE) is a Distinguished Professor and the Wexler Chair in Information Technology at the Department of Computer Science, University of Illinois at Chicago. Before joining UIC, he was with IBM Thomas J. Watson Research Center, where he was manager of the Software Tools and Techniques department. His main research interests include big data, data mining (especially on graph/network mining), privacy preserving data publishing, data stream, database systems, and Internet applications and technologies. He has published distinguished works extensively in refereed journals and conferences, amassing over 222,198 citations and achieving an H-index of 204 as of April 2025. He is a Fellow of the ACM and IEEE. He was the Editor-in-Chief of \textit{ACM Transactions on Knowledge Discovery from Data} (2011-2017) and \textit{IEEE Transactions on Knowledge and Data Engineering} (2001-2004). He is on the steering committee of \textit{ACM Conference on Information and Knowledge Management} and was a steering committee member of the \textit{IEEE Conference on Data Mining} and the \textit{IEEE Conference on Data Engineering}. 
\end{IEEEbiography}

}
\end{document}